\documentclass{article} % For LaTeX2e
\usepackage[preprint]{./neurips2026_style/neurips_2026}

\usepackage{amsmath,amsfonts,bm}

\def\eqref#1{equation~\ref{#1}}
\def\1{\bm{1}}

\def\vf{{\bm{f}}}

\def\vh{{\bm{h}}}

\def\vp{{\bm{p}}}

\def\vv{{\bm{v}}}
\def\vw{{\bm{w}}}
\def\vx{{\bm{x}}}
\def\vy{{\bm{y}}}
\def\vz{{\bm{z}}}

\def\mA{{\bm{A}}}

\def\mC{{\bm{C}}}

\def\mE{{\bm{E}}}
\def\mF{{\bm{F}}}
\def\mG{{\bm{G}}}
\def\mH{{\bm{H}}}
\def\mI{{\bm{I}}}
\def\mJ{{\bm{J}}}

\def\mQ{{\bm{Q}}}

\def\mU{{\bm{U}}}
\def\mV{{\bm{V}}}
\def\mW{{\bm{W}}}
\def\mX{{\bm{X}}}

\DeclareMathAlphabet{\mathsfit}{\encodingdefault}{\sfdefault}{m}{sl}
\SetMathAlphabet{\mathsfit}{bold}{\encodingdefault}{\sfdefault}{bx}{n}

\def\tC{{\tens{C}}}

\def\tG{{\tens{G}}}

\usepackage{xcolor}
\usepackage{hyperref}
\usepackage{url}
\usepackage{mathtools}
\usepackage{amssymb}
\usepackage{bm}
\usepackage{subcaption}
\usepackage{booktabs}
\usepackage{stackengine}

\newcommand{\T}{\mathsf{T}}
\newcommand{\N}{\mathcal{N}}
\newcommand{\D}{\mathcal{D}}
\newcommand{\dmod}{d_{\mathrm{model}}}
\newcommand{\tr}{\mathrm{Tr}}
\newcommand{\sm}{\mathrm{softmax}}
\newcommand{\diag}{\mathrm{diag}}

\renewcommand{\eqref}[1]{(\ref{#1})}

\title{Phase Transitions in Attention:\\A Bayesian Theory of Copy Head Emergence}
\author{%
\begin{tabular}{@{}c@{}}
{
Itay Lavie\textsuperscript{1,2}\thanks{Equal contribution.}~~\thanks{Correspondence to:
\texttt{itaylavie@g.harvard.edu}.} \quad
Kirsten Fischer\textsuperscript{3, 4}\footnotemark[1] \quad
Andrey Lekov\textsuperscript{5} \quad
Frederic Van Maele\textsuperscript{5}
}
\\[0.25em]
{
Zohar Ringel\textsuperscript{1} \quad
Moritz Helias\textsuperscript{3,6}
}
\\[0.7em]
{\normalfont\small\textit{
\textsuperscript{1}Racah Institute of Physics,
Hebrew University of Jerusalem,
Jerusalem 91904, Israel}
}
\\
{\normalfont\small\textit{
\textsuperscript{2}John A. Paulson School of Engineering and Applied Sciences, Harvard University,}
}\\
{\normalfont\small\textit{
Cambridge, MA 02138, USA}
}
\\
{\normalfont\small\textit{
\textsuperscript{3}Institute for Advanced Simulation (IAS-6),
Computational and Systems Neuroscience,}
}
\\
{\normalfont\small\textit{
J{\"u}lich Research Center,
J{\"u}lich 52428, Germany}
}
\\
{\normalfont\small\textit{
\textsuperscript{4}Institute of AI for Health, Helmholtz Munich, Munich, Germany}}
\\
{\normalfont\small\textit{
\textsuperscript{5}RWTH Aachen University, Aachen, Germany}}
\\
{\normalfont\small\textit{
\textsuperscript{6}Department of Physics, Faculty 1, RWTH Aachen University, Aachen, Germany}}
\end{tabular}%
}
\begin{document}

\maketitle

\begin{abstract}
Attention is the key mechanism underlying in-context learning in transformers, and attention patterns have been observed empirically to emerge abruptly during training.
We present a Bayesian theory of feature learning in attention; we then focus on how the copy subcircuit in the first layer of an induction head is learned by analyzing a single-layer softmax attention network trained on a copy task. We derive a closed-form posterior over the attention matrix and reduce it to a low-dimensional order parameter space. This reduction reveals a phase transition in the amount of training data, which we verify using both Bayesian sampling and standard training with Adam. We contrast our results with linear attention and find that softmax attention exhibits a \emph{first-order phase transition} while in linear attention an initial \emph{second-order phase transition} is followed by a smooth, continuous evolution toward the structured attention pattern (\emph{crossover}). Our work provides a first-principles theoretical account of the abrupt emergence of the copy subcircuit, reminiscent of the one observed in training large language models.
\end{abstract}

\section{Introduction}
In-context learning (ICL) has become a cornerstone of modern large language models (LLMs) and coding agents~\cite{brownLanguageModelsAre2020}, enabling models to solve and learn tasks at inference time without explicit training. Yet, while loss scaling with model size and data is well understood empirically~\cite{kaplanScalingLawsNeural2020a,hoffmannTrainingComputeOptimalLarge2022}, the ability to predict the development of capabilities such as ICL lags far behind~\cite{lourieScalingLawsAre2025}.

Whether such capabilities emerge abruptly or gradually has sparked much debate in recent years, with some works claiming some capabilities are emergent and evolve abruptly~\cite{weiEmergentAbilitiesLarge2022a,lourieScalingLawsAre2025} and others arguing that these results may be artifacts of the evaluation metric~\cite{schaefferAreEmergentAbilities2023,bertiEmergentAbilitiesLarge2025}. The former possibility has far-reaching consequences: if capabilities are acquired without prior signal, models may move between different risk classes~\cite{AnthropicsResponsibleScaling} abruptly without giving the opportunity to prepare appropriately. This possibility has motivated a line of work seeking progress measures that track capability acquisition continuously~\cite{barakHiddenProgressDeep2023,clauwInformationTheoreticProgressMeasures2024,nandaProgressMeasuresGrokking2023,Reddy24_iclr}. 

ICL capabilities have been extensively studied in two tractable synthetic settings: in-context regression, where, given input-output pairs of sequences in context, the model has to predict the output for a new input~\cite{gargWhatCanTransformers2022,akyurekWhatLearningAlgorithm2022,vonoswaldTransformersLearnInContext2023,luAsymptoticTheoryIncontext2024b} 
\begin{equation}
[x_1,\,f(x_1),\,x_2,\,f(x_2)\,...\,x_L]\to f(x_L),
\end{equation}
and in-context bigrams 
\begin{equation}
    [...\,A,\,B...A]\to B,
\end{equation}
where the model learns context-dependent token statistics via induction heads~\cite{olssonIncontextLearningInduction2022}. Remarkably, the latter was observed to be essential for a variety of ICL capabilities in LLMs~\cite{olssonIncontextLearningInduction2022,cabannesIterationHeadMechanistic2024,vonoswaldUncoveringMesaoptimizationAlgorithms2023} and their formation has been observed to exhibit an abrupt evolution, often termed \emph{emergent} and associated with a \emph{phase transition}~\cite{olssonIncontextLearningInduction2022,Reddy24_iclr,edelmanEvolutionStatisticalInduction2024,wangHowTransformersGet2025,Minegishi25BeyondInductionHeads,aoyamaPredictingEmergenceInduction2026}.

\begin{figure}[t]
    \centering
    \includegraphics[width=0.47\textwidth]{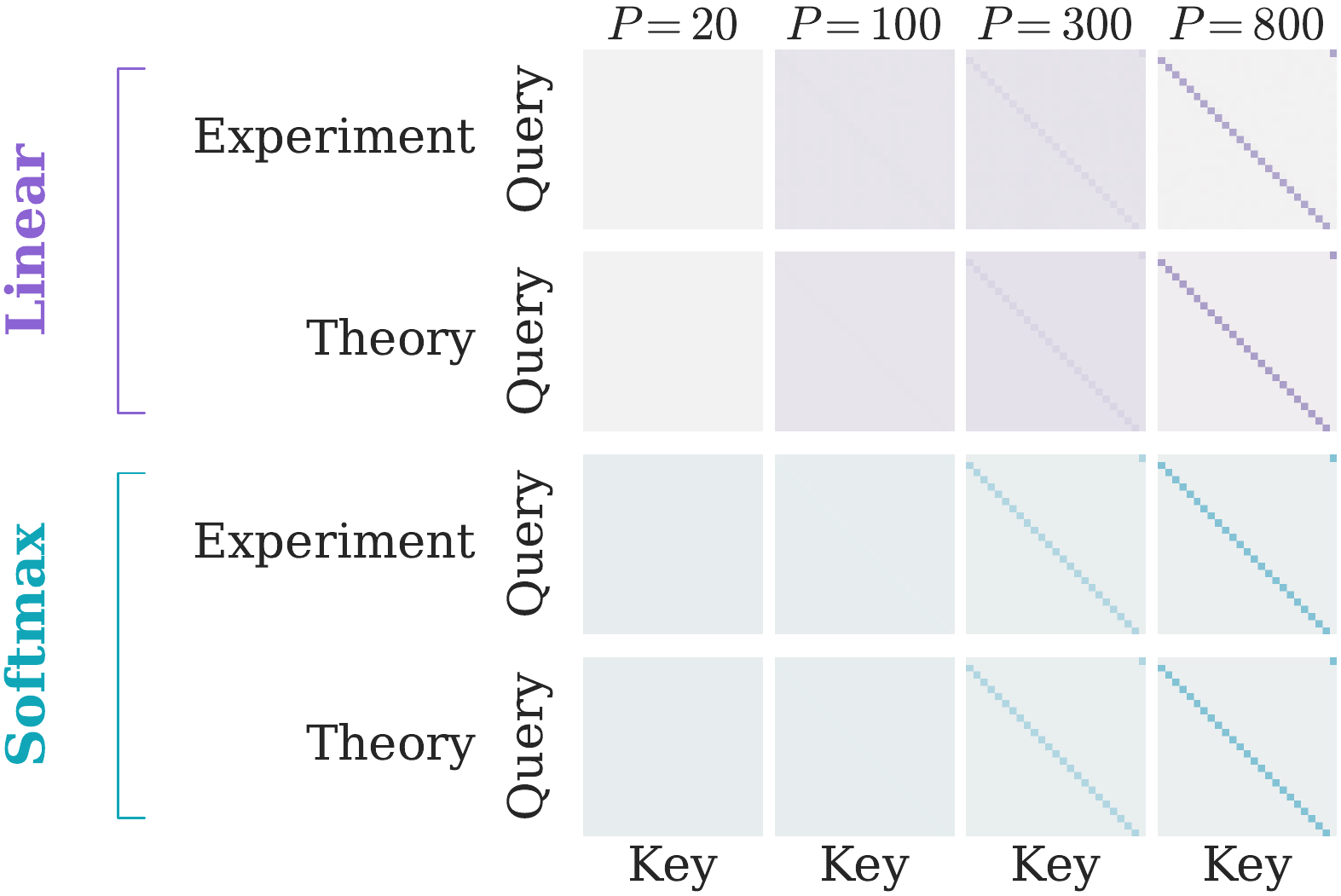}
    \includegraphics[width=0.49\textwidth]{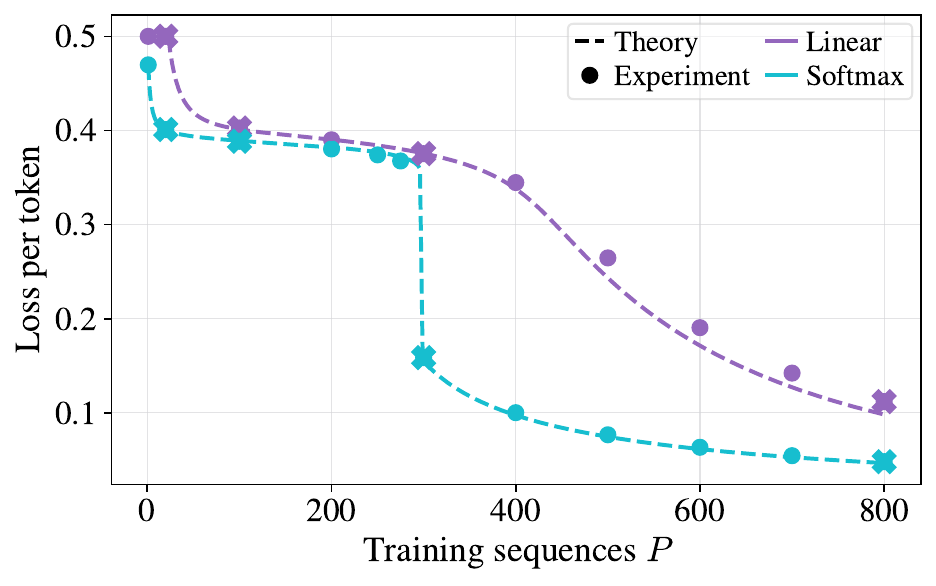}
    \caption{Copy head formation unfolds in multiple stages, with qualitatively distinct behavior for linear and softmax attention. \textbf{Left:} Attention patterns for linear (purple) and softmax (blue) attention from the trained model (top per color) and theory (bottom per color) for different amounts of training data $P$. (i) $P=20$: the attention layer does not contribute in the linear case and softmax attention remains at its uniform pooling default; (ii) $P=100$: after the second-order phase transition for linear attention, all tokens are attended to equally; (iii) $P=300$: the copy pattern begins to appear; (iv) $P=800$: it sharpens into the correct shift-by-one pattern. \textbf{Right:} Loss per token as a function of training set size $P$ for linear (purple) and softmax (blue) attention. Softmax attention exhibits a sharp, discontinuous loss drop near $P \approx 300$ corresponding to a first-order phase transition. Linear attention shows a fast continuous decrease at small $P$ (second-order transition) followed by a plateau and a slower decrease (gradual crossover). Crosses mark points on the left.}
    \label{fig:teaser_figure}
\end{figure}

In this work, we present a Bayesian theory of feature learning in a single-head attention network and analyze how the first layer of an induction head, namely the copying subcircuit,
$
    [...\,A,\,B]\to A
$
is learned.
Marginalizing over the network weights yields a posterior over attention patterns whose dominant behavior is captured by two scalar order parameters measuring the competition between uniform and copy attention. Softmax and linear attention undergo qualitatively different transitions (first-order in the former, second-order followed by a crossover in the latter) each producing distinct loss signatures (see Fig.~\ref{fig:teaser_figure}).

\textbf{Our main contributions are:}
\begin{itemize}
    \item We develop a novel analytical approach to Bayesian feature learning for a single attention head: We derive a closed-form posterior over the attention matrix and reduce it to a low-dimensional order parameter space, revealing how competing attention patterns drive a phase transition in the amount of training data.
    \item We find that the attention activation determines the order of the transition and thus the character of the capability emergence: softmax exhibits an abrupt first-order phase transition while linear attention shows a continuous second-order transition followed by a gradual crossover, providing a first-principles account of the abrupt emergence of the copy subcircuit in softmax transformers and predicting less sharp behavior in linear transformers.
    \item The order parameters map directly onto measurable loss signatures, connecting the theoretical transition types to empirically observable training metrics including the unigram-to-bigram learning progression reported in prior work~\cite{edelmanEvolutionStatisticalInduction2024}.
\end{itemize}

\subsection{Related Work} \label{subsec:related_works}
\paragraph{Theoretical limits of transformers.}
A first line of work studies transformers in the lazy limit, where no feature learning occurs: \cite{hronInfiniteAttentionNNGP2020} extended the NNGP correspondence to attention, showing single-head attention departs from Gaussianity, with $\mu$P scaling or many heads restoring it. \cite{lavieUnderstandingInductiveBias2024} characterized the spectral bias in the lazy limit through symmetry. \cite{cowsikGeometricDynamicsSignal2024,nociShapedTransformerAttention2023} studied signal propagation in transformers and \cite{bordelonInfiniteLimitsMultihead2024a} the diversity of attention heads at infinite width and heads.

\paragraph{Bayesian feature learning in attention.}
Bayesian treatments of feature learning have studied Gaussian kernels that adapt to data~\cite{Seroussi23Separation,fischer24a_pmlr,rubin25a_pmlr} and mean-field theories of strong feature learning, including grokking phase transitions~\cite{rubinGrokkingFirstOrder2023}. Closer in spirit is \cite{tiberiDissectingInterplayAttention2024} which focused on the interplay of multiple heads while freezing the attention patterns themselves. In contrast, our analysis goes beyond Gaussian features: we find a non-Gaussian distribution for attention activations and treat the layer coupling exactly, taking the attention pattern of an individual head as the primary degree of freedom rather than freezing it.

\paragraph{Training dynamics of transformers on synthetic tasks.}
\cite{edelmanEvolutionStatisticalInduction2024} showed that two-layer transformers trained on Markov sequences undergo three learning stages: (i) no in-context information used; (ii) in-context unigram statistics used; (iii) in-context bi-gram statistics used.  Our theoretical results echo these empirical findings: (i) the attention is ineffective; (ii) we see a pooling like behavior where all the context is attended equally, matching unigram statistics; (iii) only the specific position is attended to, matching the bigram statistics stage (see Fig~\ref{fig:teaser_figure}). \cite{Singh24RequirementsInductionHead} found similar stagewise dynamics (no context; in-context unigrams; in-context bigrams) in two-layer transformers on Omniglot. 
\cite{nichaniHowTransformersLearn2024,Chen24_neurips} proved that transformers learn the causal Markov chain structure, with \cite{nichaniHowTransformersLearn2024} identifying learning the first attention layer of an induction head, i.e. the copy subcircuit, as the main challenge. \cite{biettiBirthTransformerMemory2023} studied the tradeoff between in-context and in-weights learning through associative memory. 

\paragraph{Phase transitions and emergence of copy heads.}
\cite{olssonIncontextLearningInduction2022} coined the term \emph{induction head} and provided extensive empirical evidence for their abrupt emergence and causal role in ICL in LLMs across scales, with \cite{Reddy24_iclr} analyzing this in a minimal setting.
Recently, \cite{musatEmergenceInductionHeads2026} gave sharp asymptotic timescales for the appearance of an induction head in a two-layer simplified linear transformer, finding two stages with tight timescales $\Theta(L)$ and $\Theta(L^2)$, corresponding to the same stages we find here. Such timescale characterization, however, cannot distinguish a true phase transition from a crossover, as both would have tight asymptotic timescales.

\section{A Minimal Model for the Emergence of Copy Heads}
\label{sec:model}

Motivated by the emergence of induction heads in transformers~\cite{olssonIncontextLearningInduction2022}, we focus on the copy or shift-by-one subcircuit that constitutes their first layer. To this end, we study the following copy task: inputs are sequences of i.i.d. one-hot encoded tokens $x_{j}^{a}\in\{0,1\}^{L\times V}$, where $1\le a\le L$ with $L$ being the sequence length, and $V$ the vocabulary size. The target labels are $y_{i}^{a}(x)=x_{i}^{a-1}$ with periodic boundary condition.

We consider a simplified model of a single-layer attention network:
\begin{equation}
    h_{i}^{a} = \sum_j W_{ij}^{\mathrm{emb}}\,x_{j}^{a}+p_{i}^{a};
    \quad G_{H}^{ab} = \phi\!\left(\frac{\vh^{a\top}\,\mW_{H}^{G}\,\vh^{b}}{\sqrt{\dmod}}\right)_{\!b};
    \quad f_{i}^{a}(\vx) = \sum_{j,H}\,W_{ij,H}^{O}\,\sum_{b=1}^{L}G_{H}^{ab}\,h_{j}^{b},
\end{equation}
where $\phi$ denotes the attention activation: $\phi = \mathrm{softmax}$ for softmax attention, and $\phi(\cdot)_b = \frac{1}{L}(\cdot)_b$ for linear attention with a normalization factor $1/L$. The matrix $\mW^{\mathrm{emb}}$ embeds the tokens into the model space of dimension $\dmod$ and $p_{i}^{a}$ is the positional encoding. We fuse the standard matrices $\mW^{Q\top} \mW^K$ into a single matrix $\mW^G$ as commonly done in theoretical works~\cite{biettiBirthTransformerMemory2023,edelmanEvolutionStatisticalInduction2024,nichaniHowTransformersLearn2024,musatEmergenceInductionHeads2026}; in Appendix~\ref{app:attention_prior_KQ} we show that as long as $\dmod \gg L$ the resulting prior over attention patterns when parameterizing with two separate matrices $W^K$, $W^Q$ is the same, even though the prior over the product of weights $W^{K^\top}W^Q$ is different (see Appendix~\ref{app:dist_matrix_product}).  We do not include causal masking as it is not needed for the copying task.

Linear attention has become a widely adopted setting for theoretical studies of transformers~\cite{Tian23_neurips, Zhang25TrainingDynamicsLinearAttention} as it retains the token-mixing structure of softmax attention and reproduces many aspects of its optimization landscape~\cite{Ahn24LinearAttention, vonoswaldTransformersLearnInContext2023}. In this work, it serves as a controlled reference point: by studying both softmax and linear attention on the same task, we can isolate the effect of the attention activation on the phase transition order.

The full setup can be seen as a teacher-student problem with a teacher attention pattern given by ${G^*_{ab} = \delta^{a-1,b}}$ for both softmax and linear attention. The copy operation is the minimal circuit that is necessary for induction heads and other ICL capabilities~\cite{vonoswaldUncoveringMesaoptimizationAlgorithms2023}, and our task isolates it by construction. While induction heads empirically emerge under unsupervised next-token prediction \cite{olssonIncontextLearningInduction2022}, we adopt a supervised setting to enable a tractable Bayesian analysis \cite{fischer24a_pmlr, Ringel25_review}, consistent with theoretical work using synthetic tasks to study transformers \cite{vonoswaldTransformersLearnInContext2023, Tian23_neurips, Zhang25TrainingDynamicsLinearAttention}. As we show below even this minimal subcircuit exhibits a rich phenomenology.

\textbf{Scaling.}
We take a parametrization similar to $\mu$P scaling~\cite{Yang21_Neurips}: we downscale the attention weight scale $g_W= \bar{g}_W/\chi$, the output weight scale $g_O = \bar{g}_O/\chi$, and the observation noise $\sigma^2\to\sigma^2/\chi$ for large $\chi\gg1$. We also adopt the equivalent-kernel (EK) approximation, replacing the finite-dataset likelihood with the population likelihood, valid when $\sigma^2 \gg \sum_{a=1}^L (\vy^a-\vf^a(\vx))^2$~\cite{silverman1984_ek, Ringel25_review}. We first take $\dmod\to\infty$, then take $L,\sigma^2\to\infty$ jointly with $\tilde \sigma^2 \coloneqq \sigma^2/L$ to maintain EK validity. Applying the $\chi$ transformation, we have $\bar \sigma^2 \coloneqq \sigma^2/(L\chi)$ in total. Finally, we keep $\chi$ large while remaining agnostic to its scaling with $L$. As shown below, the scale of $P$ w.r.t. $L$ varies across the problem; identifying the sample complexity of each transition is one of our main theoretical findings.

\section{Bayesian Theory of Copy Head Emergence as a Phase Transition}
\label{sec:theory}
We develop a Bayesian framework over the attention matrix $\mG$:
marginalizing the network weights yields a closed-form posterior whose mode or maximum a-posteriori estimate
(MAP) is the attention pattern most favored by data and prior at a given
training-set size $P$, separately from whether an optimizer reaches it. The
negative log-posterior $\mathcal{S}(\mG)$, which we refer to as the
\emph{action}, plays the role of an effective loss landscape on $\mG$, which accounts for the number of microscopic weight configurations that produce the same attention pattern. We
develop the framework for multi-layer attention in App.~\ref{app:multi-layer} and specialize here to a single-layer network.

The copy task's inherent permutation symmetry singles out two irreducible representations of the symmetric group, corresponding to two structurally distinct attention patterns: attending uniformly versus attending to the token that predicts the next one. This structure reduces the posterior over the attention matrix $\mG$ to two corresponding scalar order parameters $\hat{c}_1$ and $\hat{c}_G$ from first principles; the order parameters are low-dimensional collective coordinates on which the posterior concentrates in the large-$\chi L$ limit, so that the dominant attention pattern is determined by the minima of an effective log-likelihood over $(\hat c_1,\hat c_G)$. These order parameters exhibit phase transitions as a function of the training data amount $P$, ultimately driving the emergence of a copy subcircuit; see App.~\ref{app:landau_primer} for a primer on order parameters and phase transitions. We first derive the theory for linear attention and characterize its phenomenology, then obtain the results for softmax attention by taking the pushforward of the resulting posterior under the softmax map.

\subsection{Network Prior as a Gaussian Process over the Attention Pattern}
Marginalizing the embedding $\mW^{\mathrm{emb}}$ and positional encoding $\vp^a$ (full derivation in App.~\ref{app:network_prior}) shows that the token embeddings $\vh^{a}_{i,\alpha}$ are i.i.d. Gaussian in the model dimension $i$ with zero mean, conditioned on their covariance (kernel). By a large-deviation argument, its empirical covariance concentrates in the $\dmod\to \infty$ limit
\begin{equation}
  Q^{ab}_{\alpha\beta}\coloneqq\frac{1}{\dmod}\sum_{i=1}^{\dmod} h^a_{i\alpha}h^b_{i\beta} \to
  C^{(xx)}_{ab,\alpha\beta}
  = g_{\mathrm{emb}}\sum_{j}x^{a}_{j\alpha}\,x^{b}_{j\beta}+g_p\,\delta_{ab}.
  \label{eq:Cxx}
\end{equation}
Integrating out $\mW^G$ gives again a conditional Gaussian prior on the attention
scores $\mG^{ab}_\alpha$ with covariance
\begin{equation}
  C^{(GG)}_{ab,cd,\alpha\beta}
  = \frac{g_K g_Q}{L^2 \dmod^2}\bigl(\vh^a_\alpha\cdot\vh^c_\beta\bigr)
    \bigl(\vh^b_\alpha\cdot\vh^d_\beta\bigr)  \;\rightarrow\; \frac{g_K g_Q}{L^2}\, C^{(xx)}_{ac,\alpha\beta}\,C^{(xx)}_{bd,\alpha\beta},
\end{equation}
where we again invoke the concentration of $\mQ$ to $\mC^{(xx)}$.
The result is a Gaussian prior on
$\mG$
\begin{equation}
  p(\mG\mid\mX)
  \propto
  \exp\!\Biggl(
    -\frac{1}{2}\sum_{ab,cd,\alpha\beta}
    G^{ab}_\alpha
    \Bigl[\frac{g_W}{L^2}C^{(xx)}_{ac,\alpha\beta}C^{(xx)}_{bd,\alpha\beta}\Bigr]^{-1}
    G^{cd}_\beta
  \Biggr),
  \qquad g_W\coloneqq g_K g_Q.
  \label{eq:prior_G}
\end{equation}
 
Finally, integrating out $\mW^O$
the conditional output distribution yields
\begin{equation}
  p(\mF\mid\mG)
  \propto
  \int\!\D\tilde{\vf}\;
  \exp\!\Bigl(
    -\tilde{f}^a_{i\alpha}f^a_{i\alpha}
    +\tfrac{g_O}{2}\,\tilde{f}^a_{i\alpha}\tilde{f}^c_{i\beta}\,
      G^{ab}_\alpha G^{cd}_\beta\,C^{(xx)}_{bd,\alpha\beta}
  \Bigr),
  \label{eq:F_given_G}
\end{equation}
with $\tilde{\vf}$ being the conjugate Fourier variable to $\vf$ and using the notation ${\mF=(\vf^a_{\alpha})_{\alpha=1,\ldots,P;\,a=1,\dots,L}}$. The latter expression \eqref{eq:F_given_G} shows that the distribution of $\vf$ factorizes over $i$.
 
\subsection{Network Posterior over the Attention Patterns}
\label{sec:posterior}

Assuming Gaussian noise on the outputs $\vy_{\alpha}=\vf_{\alpha}+\bm{\xi}$ with $\bm{\xi}\sim\N(0,\sigma^2\mI)$, the posterior then is
\begin{align}
    p(\mG \mid \{\vx_\alpha, \vy_\alpha\}_{\alpha})
  &\propto p(\mG \mid \mX) \int \D \vf  \,
     \exp \Big(-\frac{1}{2\sigma^2} \sum_{\alpha}
     \Vert \vy_{\alpha} - \vf_\alpha \Vert_2^2 \Big) \, p(\mF \mid \mG),
\end{align}
where we dropped the normalization constant (full derivation in App.~\ref{app:network_posterior}). Inserting the results for the network prior and integrating out the network outputs $\vf$ and their conjugate variables $\tilde{\vf}$, we get
\begin{equation}
  p(\mG\mid\{\vx_\alpha,\vy_\alpha\})
  \propto
  \exp\bigl(-\mathcal{S}(\mG)\bigr),
  \label{eq:posterior}
\end{equation}
with the negative log-posterior or \emph{action} being
\begin{equation}
\begin{aligned}
  \mathcal{S}(\mG)
  &=\frac{1}{2}\,\vy^{\top}\bigl[g_O\,\mG^{\top}\mC^{(xx)}\mG+\sigma^2\mI\bigr]^{-1}\vy
    +\frac{1}{2}\ln\det\bigl[g_O\,\mG^{\top}\mC^{(xx)}\mG+\sigma^2\mI\bigr]\\
  &+\frac{1}{2}\,\mG
    \bigl[\tfrac{g_W}{L^2}\mC^{(xx)}\mC^{(xx)}\bigr]^{-1} \mG
    +\frac{1}{2} \ln \det \Big[\tfrac{g_W}{L^2} \mC^{(xx)} \mC^{(xx)}\Big].
  \end{aligned}
  \label{eq:action}
\end{equation}
From this, we observe a competition between the network prior (last two terms) and the conditioning on the training labels (first two terms).
 
\subsection{Reduction of the Posterior to Interpretable Order Parameters}
Based on the permutation equivariance of the problem, we make the following representation theory-inspired ansatz for the attention matrix
\begin{equation}
  \mG = \hat c_1\,\frac{\bm{1}_L}{L} + \hat c_G\,\frac{\mG^*}{\sqrt{L}},
  \label{eq:ansatz}
\end{equation}
where $\bm{1}_L$ is the $L\times L$ all-ones matrix and $G^{*}_{ab} = \delta^{a-1,b}$ is the \emph{copy} or \emph{shift-by-one} operator. Each term is weighted by a coefficient
$(\hat c_1,\hat c_G)$, which we identify as the order parameters of the problem: $\hat c_G\gg\hat c_1$ implements previous token copying, while $\hat c_G\ll\hat c_1$ implements uniform pooling over the whole context, seeing only the single token empirical frequencies in context (in-context unigrams). The normalization factors give both modes unit Frobenius norm.

\paragraph{NNGP Prior Term.}
We evaluate $\mathcal{S}_{\mathrm{NNGP}} = \frac{L^2}{2g_W}\sum_{ab,cd,\alpha\beta}
G^{ab}_\alpha\,\bigl[C^{(xx)}C^{(xx)}
\bigr]_{_{\alpha\beta}^{ac;bd}}^{-1}G^{cd}_\beta$ by computing the summed pseudo-inverse
$\Sigma_{(ab),(cd)}=\sum_{\alpha\beta}(\mathcal{M}^{\dagger})_{ab\alpha,cd\beta}$
of $\mathcal{M}_{ab\alpha,cd\beta}=C^{(xx)}_{ac,\alpha\beta}C^{(xx)}_{bd,\alpha\beta}$ (full derivation in App.~\ref{app:nngp_projection}).
For one-hot encoded inputs, $C^{(\alpha\beta)}\otimes C^{(\alpha\beta)}$ admits a stable Kronecker--Gram
factorization for $P\gtrsim\lceil(L+V-1)^2/L^2\rceil$, which is typically small. Symmetry under permutation of token positions constrains $\Sigma$ to a three-parameter form, whose coefficients we determine numerically.
Contracting the attention patterns with the summed pseudo-inverse gives
\begin{equation}
    \small
    \mathcal{S}_{\mathrm{ NNGP}}(\hat{c}_1, \hat{c}_G)
    =\frac{\chi L}{2\bar{g}_{W}}\Bigl[\frac{1}{L}\frac{V^{2}}{(1+V/L)^{2}}\hat{c}_{1}^{2}+\frac{2}{L^{3/2}}\frac{V^{2}}{\left(1+V/L\right)^{2}}\hat{c}_{1}\hat{c}_{G}+L\frac{\left(1+V/L\right)^{2}-L^{-1}-2VL^{-2}}{\left(1+V/L\right)^{2}}\hat{c}_{G}^{2}\Bigr].
    \label{eq:NNGP_result}
\end{equation}

\paragraph{Data Term}
We evaluate $\mathcal{S}_{\mathrm{data}} = \frac{1}{2}\vy^\top[g_O\mG^\top\mC^{(xx)}\mG+\sigma^2\mI]^{-1}\vy$
by diagonalizing the operator $\mG^\top\mC^{(xx)}\mG$ (full derivation in App.~\ref{app:data_term_projection}). We choose a Fourier basis
adapted to the copy task:
\begin{equation}
\begin{gathered}
\begin{aligned}
\bigl[v_{0,0}\bigr]^a_{\alpha}
&= \frac{1}{\sqrt{PL}},
\qquad
\bigl[v_{0,k\ge 1}\bigr]^a_{\alpha}
&= \frac{1}{\sqrt{PL^2}}\sum_b \sum_j e^{i\frac{2\pi}{V}kj}\,x^b_{j,\alpha},
\end{aligned}
\\[4pt]
\begin{aligned}
\bigl[v_{1,k\ge 1}\bigr]^a_{\alpha}
&= \frac{1}{\sqrt{PL(1-L^{-1})}}\sum_j e^{i\frac{2\pi}{V}kj}
   \left(x^{a-1}_{j,\alpha}-\frac{1}{L}\sum_b x^b_{j,\alpha}\right).
\end{aligned}
\end{gathered}
\end{equation}
where $k=1,\dots,V-1$ and we note $\vv_{1,0}=0$.
The mode $\vv_{0,0}$ is uniform in both position and data;
the modes $\vv_{0,k\geq 1}$ are uniform in position but carry Fourier
structure over the vocabulary; and the modes $\vv_{1,k\geq 1}$ capture
position-to-position correlations by isolating the deviation of token
$a{-}1$ from the positional mean.
We denote the data-dependent kernel without the noise terms $\sigma^2 \mI$ as 
$\mathcal{K}^{ac}_{\alpha\beta}\coloneqq
g_O\sum_{bd}G^{ab}_\alpha G_\beta^{cd}C^{(xx)}_{bd,\alpha\beta}$.
 
The projection of the kernel $\mathcal{K}$ onto the uniform mode $\vv_{0,0}$ reduces to a scalar $\mathcal{K}_0$ given below.
For each $k \geq 1$, the kernel acts on the two-dimensional space
$(\vv_{0,k}, \vv_{1,k})$ as a $2\times 2$ matrix. Computing the action of
each of the four operator pairs
$(\bm{1}_L,\bm{1}_L)$, $(\bm{1}_L,\mG^*)$, $(\mG^*,\bm{1}_L)$,
$(\mG^*,\mG^*)$ on both modes and assembling with weights
$\hat{c}_1^2$, $\hat{c}_1\hat{c}_G$, $\hat{c}_G\hat{c}_1$, $\hat{c}_G^2$
respectively yields
\begin{equation}
\begin{aligned}
    \mathcal{K}_0 &= g_O P\!\left(\frac{L}{V}+1\right)
    \!\left(\hat{c}_1 + \frac{\hat{c}_G}{\sqrt{L}}\right)^{\!2};\quad
    &\mathcal{K}_k
    &= \frac{g_O P}{V}\,\bm{u}\bm{u}^\top;
    &\bm{u} &=
    \begin{pmatrix}
        \hat{c}_1 + \hat{c}_G/\sqrt{L} \\[3pt]
        \hat{c}_G\sqrt{(L-1)/L}
    \end{pmatrix}.
\end{aligned}
    \label{eq:u}
\end{equation}
By inverting with the Sherman--Morrison formula and summing over all Fourier modes, we get
\begin{equation}
\small
\mathcal{S}_{\mathrm{data}}\left(\hat{c}_{1},\hat{c}_{G}\right)=\frac{\chi L}{2}\left[\frac{1}{L}\frac{1}{\bar{g}_{O}\left(\hat{c}_{1}+\hat{c}_{G}/\sqrt{L}\right)^{2}\left(1+\frac{V}{L}\right)+\frac{\bar{\sigma}}{P}^{2}V}-\frac{P}{L\bar{\sigma}^{2}}\frac{V-1}{V}\left(\frac{\bar{g}_{O}\left(\hat{c}_{G}+\hat{c}_{1}/\sqrt{L}\right)^{2}}{\bar{g}_{O}\|\bm{u}\|^{2}+\frac{\bar{\sigma}^{2}LV}{P}}-1\right)\right].
  \label{eq:Sdata}
\end{equation}

 Together, Eqs.~\eqref{eq:Sdata} and \eqref{eq:NNGP_result} reduce the high-dimensional inference problem to a two-dimensional posterior without information loss. Next, we study the behavior of the action in terms of the order parameters for both linear and softmax attention. In addition to the order parameters, we obtain the network loss from the action directly, as detailed in App.~\ref{app:action_to_loss}.
\begin{figure}[t]
    \centering
   \includegraphics[width=0.24\linewidth]{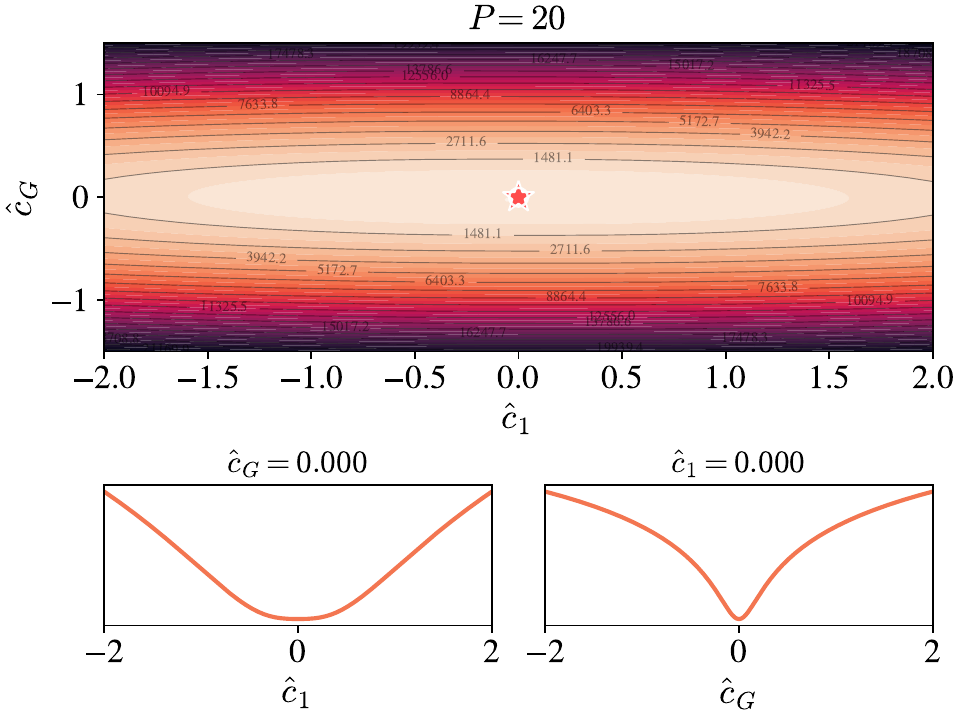}
   \includegraphics[width=0.24\linewidth]{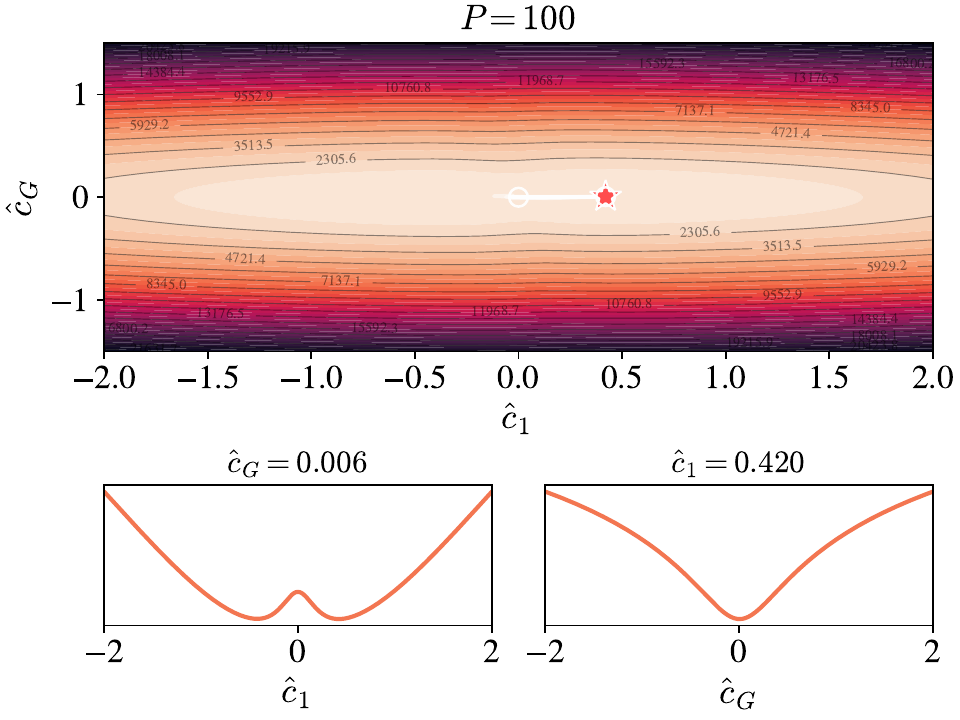}
   \includegraphics[width=0.24\linewidth]{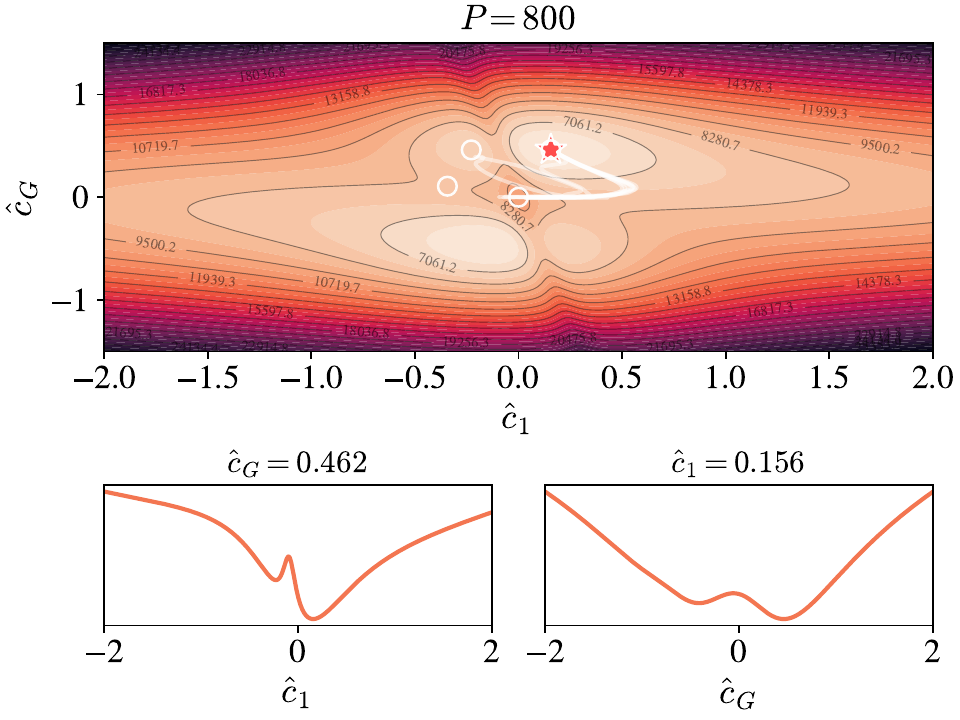}
    \includegraphics[width=0.24\linewidth]{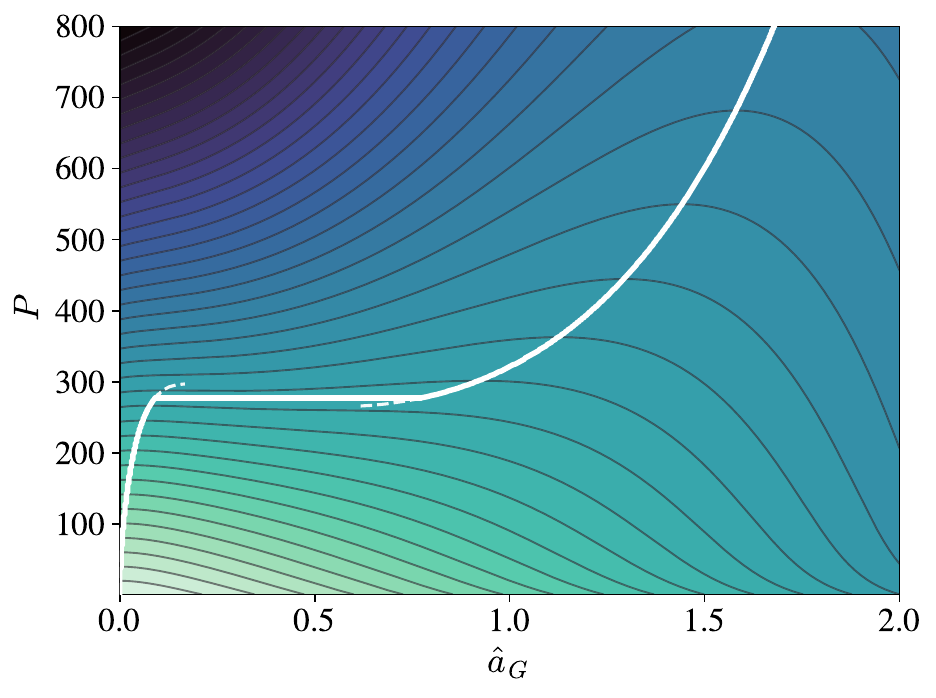}

    \caption{Different topologies of actions $\mathcal{S}^{\rm lin}(\hat{c}_1, \hat{c}_G)$ and $\mathcal{S}^{\rm softmax}( \hat{a}_G;P)$. \textbf{Linear attention (left):} Contour plots of action at increasing dataset size $P$, corresponding to the attention stages (i), (ii), (iv) in Fig~\ref{fig:teaser_figure}. From left to right: (i) the action has a single minimum at the origin - the model does not use the context; (ii) after the second-order phase transition, a minimum at nonzero $\hat{c}_1$ emerges, corresponding to uniform pooling attention where the model attends to the entire context equally; 
    (iv) finally, the landscape reorganizes and large $\hat{c}_G$ is preferred.
    The minimum deepens and splits, and two possible solutions are formed, but one is preferred and leads to lower loss values. Bottom row: one-dimensional sections through the global minimum (marked with a star in the contour plot). The plot reveals a clear Landau-type phase transition from (i) to (ii). Curves show the training trajectories projected onto $(\hat{c}_1, \hat{c}_G)$; these do not follow the gradient of $\mathcal{S}$ as optimization occurs in weight space.
    \textbf{Softmax attention (right):} Contour plot of the action $\mathcal{S}^{\rm softmax}(\hat{a}_G)$ at varying values of $P$. The white solid line indicates the location of the global minimum and dashed lines indicate local minima. At $P\approx 300$ both minima become equally deep, the global minimum jumps from the local minimum at smaller $\hat{a}_G$ to the one at larger $\hat{a}_G$, leading to a first-order phase transition.
    }
    \label{fig:mesh1}
\end{figure}

\paragraph{Linear attention action, order parameters, and sample complexities.}
Pulling out a common scaling factor of $\chi L$, we see that in the limit $L,\chi\gg1$ the action simplifies to $\mathcal{S}= \left(\mathcal{S}_{\rm NNGP}+\mathcal{S}_{\rm data} \right)+O(\chi^{-1})$ where the log-determinant term (see App.~\ref{app:fluct_term}) is subleading and can be dropped. We note that the leading terms are not strictly organized by powers of $L$ or $P$ as $\hat{c}_1,\hat{c}_G$ naturally have different $L$ and $P$ scales, as we show below. Crucially, the $\chi L$ scaling in front of $\mathcal{S}$ means  $(\hat c_1,\hat c_G)$ concentrate on deterministic values determined by a saddle point approximation. 

Notably, this action has a $\mathbb{Z}_2$ symmetry $(\hat c_1, \hat c_G)\to (-\hat c_1, -\hat c_G)$, which governs its second-order phase transition. Interestingly, this action has a larger approximate $\mathbb{Z}_2 \times \mathbb{Z}_2$ symmetry that allows sign flips of $\hat c_1, \hat c_2$ independently. The presence of this approximate symmetry leads to the second plateau and the subsequent crossover; we formalize this in a field-theoretic toy model in App.~\ref{app:field_theory_toy}.

Expanding the action around the origin gives two sample-complexity scales. The phase transition occurs when the $\mathbb{Z}_2$ symmetry is first broken, where the curvature in the $\hat c_1$ direction changes sign
\begin{equation}
P_{\hat{c}_{1}}^{*}=\frac{L^{2}V^{2} \bar{\sigma}^{2}}{L+V }\sqrt{\frac{1}{ \bar{g}_{O}\bar{g}_{W}\left(L+1\right)\left(L+V-1\right)}}\sim V^{2}.
\label{eq:P_c1}
\end{equation}
The crossover is estimated from the point where the curvature along the $\hat c_G$ direction changes sign
\begin{equation}
P_{\hat{c}_{G}}^{*}=\frac{L^{2}\bar{\sigma}^{2}}{L+V}\sqrt{\frac{V\left(L(L-1)+2(L-1)V+V^{2}\right)}{(L+1)\bar{g}_{O}\bar{g}_{W}}}\sim L^{3/2}V^{1/2}.
\label{eq:P_cross_main}
\end{equation}
Fig.~\ref{fig:mesh1} (left panels) illustrates the topology of the action $\mathcal{S}(\hat{c}_1, \hat{c}_G)$ for different $P$ values. At $P<P^*_{\hat c_1}$, the action has a single minimum at the origin. At $P=P_{\hat c_1}\approx23$, a new minimum at nonzero $\hat{c}_1$ emerges continuously and the sections reveal the characteristic flat-then-tilted shape of a Landau-type second-order phase transition (see App.~\ref{app:landau_primer}); enabling the model to exploit in-context unigram statistics. Beyond this, as $P$ increases toward $P_{\hat c_G}\approx322$, the saddle at the origin becomes a maximum, driving $\hat{c}_G$ away from zero and the crossover from context pooling to copying. Critically, this reorganization is not a second phase transition: the symmetry is already broken at $P_{\hat{c}_1}$.
\begin{figure}[t]
    \centering
    \includegraphics[width=0.35\textwidth]{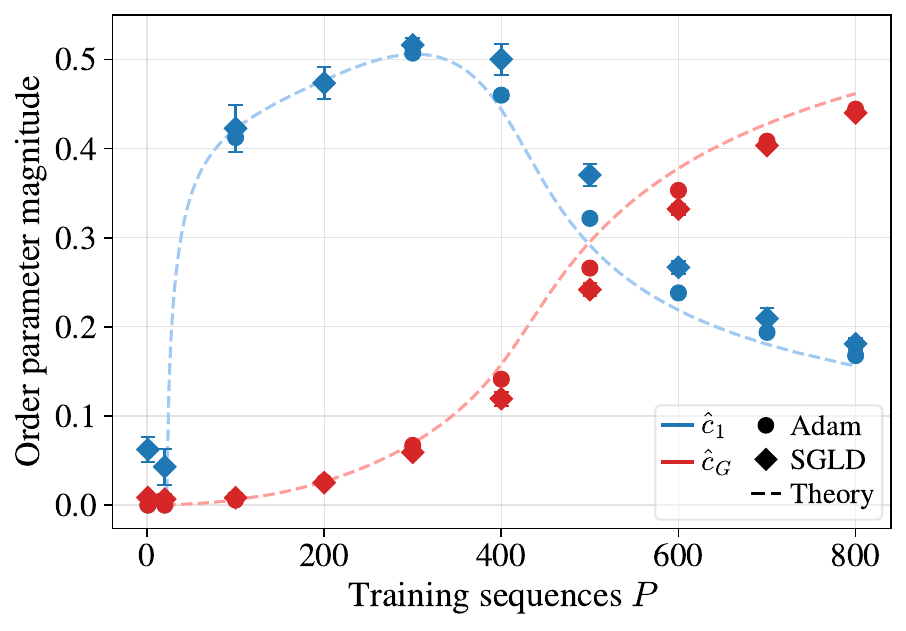}
    \includegraphics[width=0.64\textwidth]{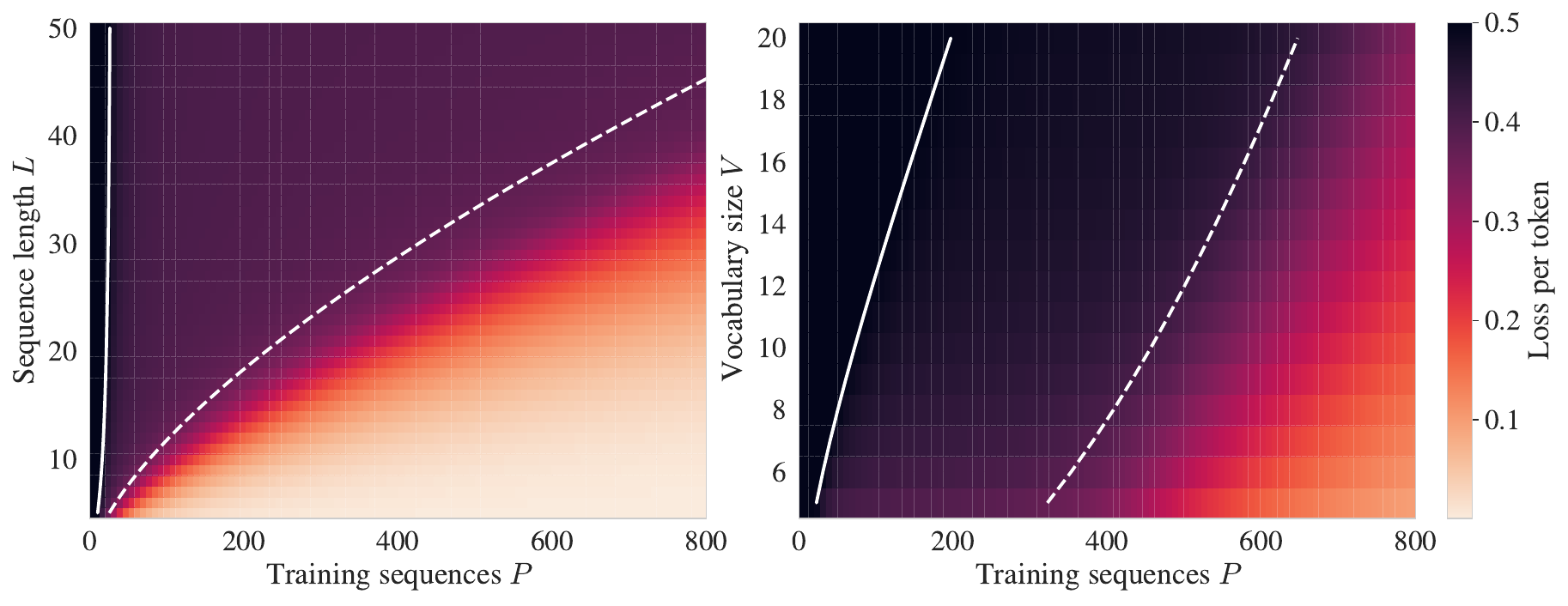}
    \caption{\emph{Linear attention} exhibits a second-order phase transition and subsequent gradual crossover. \textbf{Left:} Scalar order parameters $\hat{c}_1,\hat{c}_G$ measured from models trained with Adam (circles) and SGLD (diamonds) compared to the theory predictions (dashed) as a function of the amount of training data or sample complexity $P$. At the phase transition, $\hat{c}_1$ rises sharply from zero, while $\hat{c}_G$ remains near zero; beyond $P\approx 400$, $\hat{c}_1$ declines while $\hat{c}_G$ rises, marking the crossover from uniform pooling to copy attention; theory tracks the empirically observed behavior closely.
    \textbf{Center and right:}
    Phase diagram of copy subcircuit emergence. We show the loss per token as a function of training sequences $P$ vs.\ sequence length $L$ for $V=5$ fixed (center), and vs.\ vocabulary size $V$ for fixed $L=25$ (right).
    The two phases identified from the order parameters $\hat{c}_1, \hat{c}_G$ manifest as distinct loss plateaus separated by two transition regions; the phase boundaries are predicted by our theory as a function of $P^*(L,V)$ in Eqs.~\eqref{eq:P_c1} (solid white) and \eqref{eq:P_cross_main} (dashed white).}
    \label{fig:linear_OPs}
\end{figure}

\paragraph{Softmax attention action, order parameter, and sample complexity.}
We now take the pushforward of the posterior under the softmax map. We use $\hat{a}_1,\,\hat{a}_G$ for the order parameters past the softmax and define them as projections onto our base function; mirroring $\hat{c}_1,\,\hat{c}_G$. First, the row-wise normalization of softmax eliminates the $\hat a_1$ order parameter, as
\begin{equation}
    \hat a_1 = 1-\frac{\hat a_{G}}{\sqrt{L}}.
\label{eq:a1_to_ag}
\end{equation}
Furthermore, the softmax activation is invariant under overall shifts of the inputs, which makes $\hat{a}_1,\,\hat{a}_G$ independent of $\hat c_1$; therefore it will be fixed solely by the weight decay / prior term and $\hat c_1=0$ at the saddle point.
Next, since the data term is insensitive to how the attention pattern is parameterized, it requires no adaptation and we can simply plug in $(\hat{a}_1,\,\hat{a}_G)=(\hat{c}_1,\,\hat{c}_G)$. 
Finally an explicit calculation of the mapping between $\hat{c}_G$ and $\hat{a}_G$ gives (full derivation in App.~\ref{app:softmax_map})
\begin{equation}
    \hat c_{G}=\frac{1}{\sqrt{L}}\log\left[\frac{1+\frac{L-1}{\sqrt{L}} \hat a_{G}}{1-\frac{1}{\sqrt{L}} \hat a_{G}}\right].
\label{eq:cg_to_ag}
\end{equation}
Substituting Eq.~\eqref{eq:a1_to_ag} and $\hat{c}_1=0$ into the prior term Eq.~\eqref{eq:NNGP_result}, and replacing $(\hat{c}_1,\hat{c}_G)\to(\hat{a}_1,\hat{a}_G)$ in the data term Eq.~\eqref{eq:Sdata} (which depends only on the post-softmax attention matrix), gives the action for softmax attention. Dropping constants and subleading terms we find
\begin{equation}
    S_{\rm softmax}(\hat a_G)=\frac{\chi L}{2}\left(\frac{1}{\bar{g}_{W}}\log^{2}\left(\frac{1+\sqrt{L}\hat a_G}{1-\hat a_G/\sqrt{L}}\right)-\frac{P}{\bar{\sigma}^{2}L}\frac{V-1}{V}\frac{\left(\hat a_G+\frac{1}{\sqrt{L}}\right)^{2}\bar{g}_{O}}{\left(\hat a_G^{2}+1\right)\bar{g}_{O}+\frac{\bar{\sigma}^{2}L}{P}V}+O\left(\frac{1}{L}\right)\right).
\end{equation}
The self-consistent equation $\partial_{\hat a_G} S\stackrel{!}{=} 0$ has to be solved numerically for $\hat{a}_G^*$ to obtain the MAP values. Investigating the topology of the action in Fig.~\ref{fig:mesh1} (right), we find two competing minima, with the global minimum jumping from one to the other at a critical amount of training data $P^*_{\hat{a}_G}$, indicating a first-order phase transition. We can get an analytical approximation for the sample complexity $P^*_{\hat{a}_G}$ by solving $S(0)=S(\hat a)$ for $P$ and plugging in $\hat a=O(1)$ at large $L$
\begin{equation}
    P^*_{\hat{a}_G} \approx \frac{1}{4\bar{g}_{W}}L\frac{V}{V-1}\bar{\sigma}^{2}\ln^{2}(L)\left(\sqrt{1+4\frac{V-1}{\ln^{2}(L)}\frac{\bar{g}_{W}}{\bar{g}_{O}}}+1\right) \sim L \ln^2 (L),
\label{eq:P_a}
\end{equation}
constituting an improvement of $O(\sqrt{LV})$ in the sample complexity compared to linear attention.

\section{Experiments}
We validate our theoretical predictions against networks trained with Adam~\cite{Kingma15Adam} and sampled with SGLD~\cite{wellingBayesianLearningStochastic2011, Naveh21_064301} in Fig.~\ref{fig:teaser_figure}. Theory and experiment agree quantitatively across all stages. The two attention mechanisms differ qualitatively: linear attention exhibits a second-order phase transition at $P\approx100$, where $\hat{c}_1$ departs from zero and the model begins pooling uniformly over the context, followed by a crossover as $\hat{c}_G$ grows and the copy pattern gradually sharpens. Softmax attention instead undergoes a first-order transition at $P\approx300$, with a discontinuous loss drop from the initial uniform pooling solution directly to a copy solution. In both cases, the pooling stage likely corresponds to learning the first-order statistics of the data while the second stage reflects the landscape topology change as discussed in Fig.~\ref{fig:mesh1}: the uniform pooling minimum destabilizes and the copy minimum deepens.

Remarkably, while our predictions are for the MAP values, we find excellent agreement with models trained using Adam: minimizing the action \eqref{eq:action} at large $\chi$ corresponds to training with gradient flow with weight decay (see App. \ref{app:langevin} for derivation and refined statement). Since gradient descent may converge to local minima, we ran an ensemble of models per $P$ value and report results from runs that reached the global minimum (see App.~\ref{appendix:traj_filtering}; for softmax all runs converge to the global minimum); SGLD instead targets the Bayesian posterior as its stationary distribution and, given sufficient mixing time, would sample basins by their posterior weights without requiring such explicit filtering. 

Next, we examine the two transitions in detail through the order parameters $\hat{c}_1, \hat{c}_G$. For linear attention, Fig.\ref{fig:linear_OPs} (left panel) shows that both Adam and SGLD results track the theoretical predictions closely across different amounts of training data $P$. The transient deviation of SGLD near the crossover is attributable to large fluctuations as the uniform pooling minimum destabilizes and the copy minimum emerges (Fig.~\ref{fig:mesh1}), causing the posterior to broaden and the network to transiently explore directions outside the two-dimensional order parameter subspace; we quantify this coverage in App.~\ref{app:attn_coverage}. The three-phase structure is confirmed in the $(P,L)$ and $(P,V)$ phase diagrams (Fig.~\ref{fig:linear_OPs} center and right): the order parameter transitions manifest as distinct loss plateaus, with the two boundaries scaling with $L$ and $V$ as predicted by Eqs.~\eqref{eq:P_c1},\eqref{eq:P_cross_main}. The first boundary marks the onset of structured attention, the second the emergence of the copy pattern.

\begin{figure}
    \centering
    \includegraphics[width=0.35\textwidth]{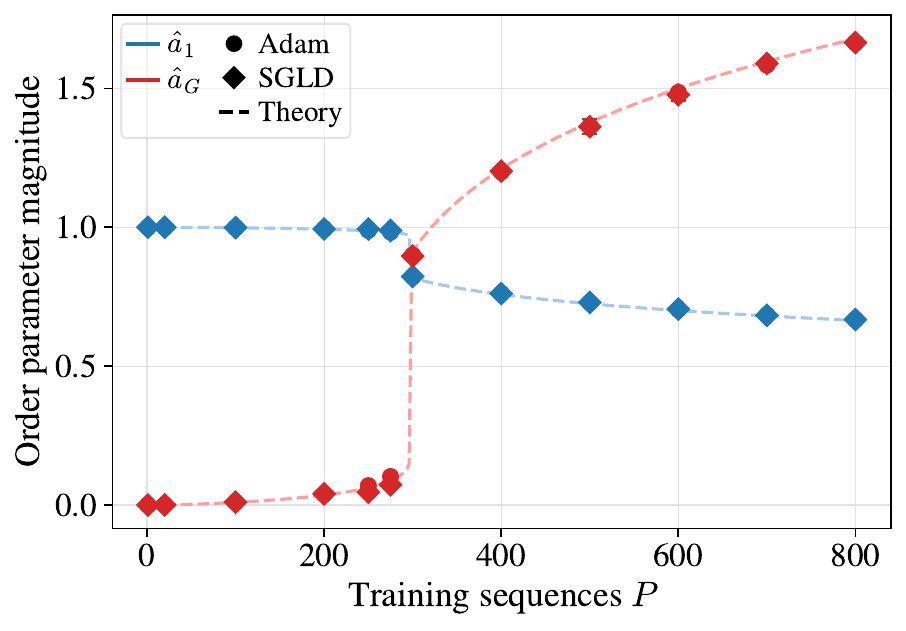}
    \includegraphics[width=0.64\linewidth]{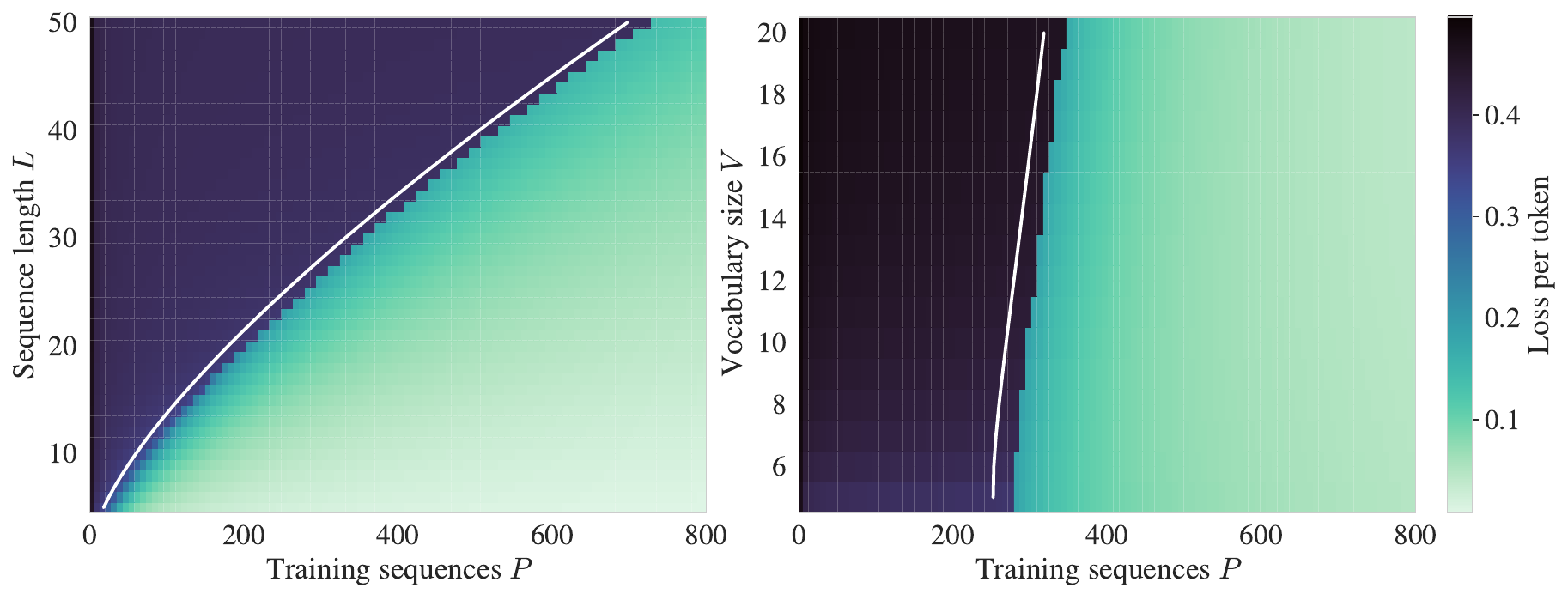}
    \caption{\emph{Softmax attention} exhibits a first-order phase transition. \textbf{Left:} Scalar order parameters $\hat{a}_1,\hat{a}_G$ measured from models trained with Adam (circles) and SGLD (diamonds) compared to the theory predictions (dashed lines) as a function of the amount of training data or sample complexity $P$. At the phase transition the value of $\hat{a}_G$ jumps marking a first-order phase transition. Note that the dots corresponding to SGLD are often covered by the diamonds corresponding to Adam due to their precise agreement.
    \textbf{Center and right:}
    Phase diagram of copy subcircuit emergence. We show the loss per token as a function of training sequences $P$ vs.\ sequence length $L$ for $V=5$ fixed (center), and vs.\ vocabulary size $V$ for fixed $L=25$ (right). The transition from a phase of uniform attention to a phase of copy attention manifests as a sharp drop in the loss; the phase boundary $P^*(L,V)$ (white) is predicted by our theory in Eq.~\eqref{eq:P_a}.}
    \label{fig:softmax_OPs}
\end{figure}

Softmax attention in Fig.~\ref{fig:softmax_OPs} presents a strikingly different picture: rather than the three-phase structure of linear attention, $\hat{a}_G$ jumps discontinuously at a single phase boundary and the loss drops abruptly, with no initial learning phase. The boundary is predicted by Eq.~\eqref{eq:P_a} across the full $(P,L)$ and $(P,V)$ planes, with the theory tracking the empirical transition remarkably well. The sharp separation between the two phases is a direct signature of the first-order phase transition.

\section{Discussion}
\label{sec:discussion}
We presented a novel Bayesian theory of feature learning in a single attention head, obtaining a first-principles theory explaining how the copy subcircuit emerges as a phase transition in the amount of training data. By reducing the posterior over attention patterns to a low-dimensional order parameter space, we obtain quantitative predictions for the phase boundary and the attention pattern evolution.

A key finding is that the transition order depends on the attention mechanism. For linear attention, an initial \emph{second-order phase transition} to uniform pooling attention is followed by a smooth, continuous crossover toward the copy attention; this is not a second phase transition, since the symmetry is already broken at the first transition and no additional singularity can occur. For softmax attention, we instead find a \emph{first-order phase transition}: a discontinuous onset of structured attention with no precursor signal. That the same Bayesian framework yields qualitatively different transition orders for the two mechanisms is a demonstration of its discriminative power: it resolves mechanistic differences that coarser analyses would miss. Together, these results provide a fine-grained and quantitative account of copy subcircuit emergence.

\paragraph{Implications to predictability of emergent phenomena.}
Our results show that the transition order has direct consequences for whether capability emergence can be anticipated. The first-order phase transition of softmax attention is discontinuous: the system provides no precursor signal before the phase boundary is crossed, making the onset of copying behavior intrinsically unpredictable from observable quantities. In contrast, the second-order phase transition of linear attention is preceded by diverging fluctuations~\cite{kardarStatisticalPhysicsFields2007} (we are aware of one work that utilizes these~\cite{hennickDensityMatricesPhase2026}), and the subsequent gradual crossover allows capabilities to be tracked before they are fully expressed. 
This makes understanding which type of transition governs a given capability in a given architecture crucial for designing a monitoring strategy, and suggests fundamental limits to what can be monitored.

\paragraph{Limitations and Outlook.}
To reduce the posterior to a low-dimensional order parameter space, we make several assumptions:
While our Bayesian framework admits a multi-layer generalization outlined in App. \ref{app:multi-layer}, we focus here on the single-head, single-layer setting and the supervised copy task to isolate the copy subcircuit, which is our main subject of interest. The most natural next step is a two-layer transformer with causal masking to study how the copy head feeds into the induction head and, in particular, whether that changes the transition character. The phase boundary $P^*(L, V)$ derived here could further serve as a basis for scaling laws predicting copy subcircuit emergence as a function of model and data scale. Extending from i.i.d.\ uniform to structured or natural-language token distributions will be interesting to investigate qualitative changes in the transition and shifts in the sample complexity $P^*$. Finally, our derivation relies on the equivalent-kernel approximation and the $d_\text{model}\to\infty$ limit, suppressing finite-size corrections of the data distribution and of the Gaussian prior on the attention pattern, respectively. These approximations are accurate in the strong regularization regime and for large model dimensions, respectively, but a systematic treatment of small regularization or finite $P$ and finite $d_\text{model}$ corrections similar to \cite{fischer24a_pmlr,rubin25a_pmlr} will be an interesting direction for future work.

\subsubsection*{Acknowledgments}
We thank Noa Rubin for the insightful discussion and comments on the manuscript. I.L. acknowledges the Helmholtz Information \& Data Science Academy (HIDA) for providing
financial support that enabled a short-term research stay at IAS-6, during which this project was initiated. Z.R. and I.L. acknowledge support from ISF
Grant 2250/19. The authors gratefully acknowledge the computing time granted by the JARA Vergabegremium and provided on the JARA Partition part of the supercomputer JURECA at Forschungszentrum Jülich (computation grant JINB33).

\subsubsection*{Societal Impact Statement}
This paper works towards understanding the emergence of copy heads in transformers, thus aiming to advance mechanistic explainability of transformers. While the latter surely has societal impacts, these will be much further down the line.
\clearpage

\bibliographystyle{unsrt}
\bibliography{references}

\newpage

\appendix
\part*{Appendix}
\section{A short introduction to Landau Theory and Phase Transitions}
\label{app:landau_primer}

This appendix gives a self-contained statistical physics background for some of the ideas used in the text. We focus on Landau theory, order parameters, symmetries, first- and second-order phase transitions, and crossovers. Phase transitions and statistical mechanics have a long history in machine learning, beginning with work on the phase transition in the capacity of the perceptron~\cite{gardnerSpaceInteractionsNeural1988,gardnerOptimalStorageProperties1988}. This appendix, however, does not aim to cover that rich history, some recent reviews of the field include~\cite{heliasStatisticalFieldTheory2020,bahriStatisticalMechanicsDeep2020a,cuiHighdimensionalLearningNarrow2025,Ringel25_review}. We discuss the most directly related work in Section~\ref{subsec:related_works}.

\subsection{Phases, Control Parameters, and Large-System Limits}

A \emph{phase} is a macroscopic regime characterized by qualitatively stable behavior. In this paper, the phases are regimes of the learned attention pattern: for example, a regime where the attention mechanism does not use the context, a regime where it pools uniformly over all positions, and a regime where it implements the copy pattern.

A \emph{control parameter} is a quantity whose variation can drive the system between phases. In our setting the central control parameter is the amount of training data $P$, while other quantities such as sequence length $L$, vocabulary size $V$, and noise scale $\sigma^2$ also affect the phase boundaries.

A true phase transition is strictly defined only in a large-system limit. For finite systems, observables are typically smooth functions of the control parameter. Sharp nonanalytic behavior, which is required for a phase transition, appears only after taking an asymptotic limit. In our theory, the analogous large parameter is the overall scale multiplying the reduced action, in particular the $\chi L$ scaling appearing in the posterior. At finite $\chi$ and finite $L$, transitions are softened; in the large-$\chi L$ limit, the posterior concentrates around the minima of the action and sharp transitions can appear.

\subsection{Macroscopic Actions and Concentration}

The central mathematical structure behind Landau theory of phase transitions, which we be explained below, is a probability distribution of the form
\begin{equation}
    p(m)
    \propto
    \exp[-\Lambda\, s(m)],
    \label{eq:large_deviation_form_primer}
\end{equation}
where $m$ is a low-dimensional macroscopic variable, $s(m)$ is an action density, and $\Lambda$ is a large coefficient; naturally this is intimately related to the theory of large deviations~\cite{Touchette09}. In our setting, $\Lambda$ is represented by the large factor $\chi L$ multiplying the reduced action.

The normalization of this distribution is
\begin{equation}
    Z
    =
    \int dm\,\exp[-\Lambda s(m)].
\end{equation}
The associated free energy is
\begin{equation}
    \mathcal F
    =
    -\log Z.
\end{equation}
When $\Lambda$ is large, the integral is dominated by the minimum of the action density:
\begin{equation}
    m^\star=\arg\min_m s(m).
\end{equation}
The saddle-point approximation then gives
\begin{equation}
    \mathcal F
    =
    -\log Z
    \simeq
    \Lambda s(m^\star)
    +
    \text{subleading fluctuation terms}.
\end{equation}
Thus the phase selected by the posterior is determined by the minimum of the reduced action, while the free energy is obtained by evaluating the corresponding marginal integral.

The large coefficient in Eq.~\eqref{eq:large_deviation_form_primer} has two roles. First, it justifies a saddle-point approximation. If $m^\star$ is a global minimum, then expanding around it gives
\begin{equation}
    s(m)
    =
    s(m^\star)
    +
    \frac{1}{2}s''(m^\star)(m-m^\star)^2
    +
    \cdots.
\end{equation}
Substituting this into Eq.~\eqref{eq:large_deviation_form_primer} gives
\begin{equation}
    p(m)
    \approx
    \exp[-\Lambda s(m^\star)]
    \exp\!\left[
        -\frac{\Lambda}{2}s''(m^\star)(m-m^\star)^2
    \right].
\end{equation}
Thus typical fluctuations around the minimum scale as
\begin{equation}
    |m-m^\star|
    \sim
    \frac{1}{\sqrt{\Lambda}}.
\end{equation}
As $\Lambda\to\infty$, the distribution collapses onto the minimizing value of the macroscopic variable.

Second, Eq.~\eqref{eq:large_deviation_form_primer} is a sign that a successful macroscopic description has emerged. The original system may have many microscopic degrees of freedom, such as network weights and hidden activations. After marginalizing or averaging over them, the dominant behavior is governed by a small number of collective variables and a scalar action. In this paper, the collective variables are the attention order parameters, and the large factor $\chi L$ makes their saddle-point values sharply defined.

In our Bayesian setting, the posterior over attention patterns has the form
\begin{equation}
  p(\mG\mid \text{data})
  \propto
  \exp[-\mathcal{S}(\mG)].
\end{equation}
After reducing to the relevant order-parameter space, this becomes
\begin{equation}
  p(\hat c_1,\hat c_G\mid \text{data})
  \propto
  \exp[-\mathcal{S}(\hat c_1,\hat c_G)].
\end{equation}
In the scaling regime used in the main text,
\begin{equation}
    \mathcal S(\hat c_1,\hat c_G)
    =
    \chi L\, S(\hat c_1,\hat c_G)
    +
    \text{subleading terms}.
\end{equation}
Therefore the dominant learned attention pattern is determined, to leading order, by the minima of the reduced action $S$.

\subsection{Order Parameters}

An \emph{order parameter} is a low-dimensional quantity that distinguishes phases. It is chosen to capture the collective degree of freedom that changes qualitatively across a transition.

In this paper, the order parameters are scalar projections of the attention matrix onto interpretable attention patterns. For linear attention, we have
\begin{equation}
  \mG
  =
  \hat c_1\,\frac{\bm{1}_L}{L}
  +
  \hat c_G\,\frac{\mG^*}{\sqrt L}.
\end{equation}
The coefficient $\hat c_1$ measures the strength of uniform pooling over the context, while $\hat c_G$ measures the strength of the copy or shift-by-one pattern. In the softmax case, row normalization removes one independent direction, leaving the copy order parameter $\hat a_G$.

The key point is that an order parameter is not merely a summary statistic. It is the macroscopic coordinate in which the posterior concentrates. In the present work, the order parameters distinguish whether the network ignores context, pools over context, or selectively attends to the previous token.

\subsection{Landau Polynomials, Minima, and Phase Transitions}

Landau theory studies the reduced action as a function of the order parameter. Near a transition, the action is often expanded as a polynomial. For a single scalar order parameter $m$, one writes
\begin{equation}
    S(m)
    =
    S_0
    +
    h m
    +
    \frac{r}{2}m^2
    +
    \frac{w}{3}m^3
    +
    \frac{u}{4}m^4
    +
    \cdots.
    \label{eq:general_landau_polynomial_primer}
\end{equation}
This polynomial is a local expansion of the macroscopic action after the microscopic degrees of freedom have been integrated out. The coefficients depend on the control parameter. In our setting, they depend on $P$ and on problem parameters such as $L$, $V$, and $\sigma^2$.

The phases are determined by the minima of this polynomial action. If the global minimum is at $m=0$, then the corresponding order parameter is absent. If the global minimum is at $m\neq0$, then the corresponding structure is present. A phase transition occurs when the identity or location of the global minimum changes nonanalytically in the large-$\chi L$ limit.

For example, suppose symmetry forbids odd powers of $m$, giving
\begin{equation}
  S(m)
  =
  \frac{r}{2}m^2
  +
  \frac{u}{4}m^4,
  \qquad u>0.
  \label{eq:landau_simple_primer}
\end{equation}
When $r>0$, the unique minimum is at
\begin{equation}
    m^\star=0.
\end{equation}
When $r<0$, the point $m=0$ becomes unstable and two new minima appear at
\begin{equation}
    m^\star
    =
    \pm\sqrt{\frac{-r}{u}}.
\end{equation}
Thus a change in the sign of a polynomial coefficient changes the structure of the minima, and hence changes the phase selected by the posterior.

In our Bayesian setting, the reduced action is the negative log-posterior restricted to the relevant order-parameter subspace. The learned attention pattern is determined by the minimum of $\mathcal{S}_{\rm linear}(\hat c_1,\hat c_G)$ for linear attention, or $\mathcal{S}_{\rm softmax}(\hat a_G)$ for softmax attention. The phase transition is therefore a change in the global minimum of the action as the amount of training data $P$ is varied.

The structure of the action reflects a competition between two effects. The prior term favors simple, small-norm attention patterns. The data term favors attention patterns that correctly predict the copy labels. A phase transition occurs when increasing the dataset size $P$ makes a structured attention pattern favorable enough to overcome the prior cost.

\subsection{Symmetry and Symmetry Breaking}

Symmetry is central to the classification of phase transitions. A system has a symmetry if the action is unchanged under a transformation of the order parameter. For example, the polynomial in Eq.~\eqref{eq:landau_simple_primer} is invariant under
\begin{equation}
  m\mapsto -m.
\end{equation}
This is a $\mathbb{Z}_2$ symmetry. When $r>0$, the minimum is at $m=0$, which is invariant under this symmetry. When $r<0$, the minima are at nonzero values of $m$, and choosing one of them breaks the symmetry.

Symmetry also determines which polynomial terms are allowed. If the action is invariant under $m\mapsto -m$, then odd powers such as $m$ and $m^3$ are forbidden. If this symmetry is absent, odd powers may appear, and the transition structure can change.

In the linear-attention theory, the reduced action has an exact diagonal sign symmetry
\begin{equation}
  (\hat c_1,\hat c_G)
  \mapsto
  (-\hat c_1,-\hat c_G).
\end{equation}
The second-order transition corresponds to the spontaneous selection of a nonzero ordered direction in this order-parameter space. The action also has an approximate larger symmetry that nearly allows independent sign flips of $\hat c_1$ and $\hat c_G$. This approximate structure is responsible for the later confined reorganization toward the copy mode, but because the exact symmetry has already been broken, this later reorganization is a crossover rather than a second phase transition.

In the softmax-attention case, the softmax map changes the effective geometry of the order-parameter space. Row normalization removes the uniform attention direction as an independent degree of freedom, and the resulting one-dimensional action develops two competing minima. The transition occurs when the global minimum jumps from the weak-copy solution to the strong-copy solution.

\subsection{Second-Order Phase Transitions}

A \emph{second-order} or \emph{continuous} phase transition is one in which the order parameter turns on continuously at the critical point. In the Landau polynomial of Eq.~\eqref{eq:landau_simple_primer}, minimizing $S(m)$ gives
\begin{equation}
  m^\star = 0
  \quad \text{for} \quad r>0,
  \qquad
  |m^\star|
  =
  \sqrt{\frac{-r}{u}}
  \quad \text{for} \quad r<0.
\end{equation}
Thus $m^\star$ approaches zero continuously as $r\to 0^-$. The curvature at the origin,
\begin{equation}
  \frac{d^2S}{dm^2}\bigg|_{m=0}=r,
\end{equation}
vanishes at the transition. This vanishing curvature means that fluctuations in the order parameter become large near the critical point:
\begin{equation}
    \langle (m-m^\star)^2\rangle
    \sim
    \frac{1}{\Lambda\, S''(m^\star)}.
\end{equation}
When the curvature tends to zero, the distribution becomes broad in the ordering direction before the ordered solution fully develops.

For linear attention in the main text, the first transition is of this kind. As $P$ crosses the critical value $P^*_{\hat c_1}$, the minimum at the origin loses stability and the pooling order parameter $\hat c_1$ turns on continuously. The network begins to exploit in-context information, but it initially does so through uniform pooling rather than through a sharp copy pattern.

This distinction matters for predictability. A second-order transition has precursor signals: the action becomes increasingly flat along the unstable direction, and fluctuations in the corresponding order parameter grow. In principle, this can make the approach to the transition detectable before the ordered capability is fully expressed.

\subsection{First-Order Phase Transitions}

A \emph{first-order} or \emph{discontinuous} phase transition occurs when the global minimum of the action jumps discontinuously from one local minimum to another. Unlike a second-order transition, the original minimum need not lose local stability at the transition. Instead, two locally stable phases coexist at the critical point, and the transition occurs when their action values become equal.

A minimal Landau polynomial with two competing minima is
\begin{equation}
  S(m)
  =
  \frac{r}{2}m^2
  -
  \frac{w}{4}m^4
  +
  \frac{v}{6}m^6,
  \qquad
  w>0,\quad v>0.
  \label{eq:first_order_landau_primer}
\end{equation}
The sixth-order term stabilizes the polynomial, while the negative quartic term allows a nonzero local minimum to appear before the zero-order-parameter minimum becomes unstable. At the transition point, the two minima have equal action and the global minimizer jumps.

The exponential concentration makes this jump sharp. If two local minima are located at $m_1^\star$ and $m_2^\star$, then their posterior weights scale as
\begin{equation}
    \frac{p(m_1^\star)}{p(m_2^\star)}
    \approx
    \exp\!\left[
        -\Lambda
        \left(
            S(m_1^\star)-S(m_2^\star)
        \right)
    \right].
\end{equation}
For large $\Lambda$, even a small difference in action is exponentially amplified. Therefore, when the two action values cross, the dominant saddle switches abruptly.

The corresponding free energy is controlled by the lower of the two saddle values:
\begin{equation}
    \mathcal F
    =
    -\log Z
    \simeq
    \Lambda\min\{S(m_1^\star),S(m_2^\star)\}
    +
    \text{subleading terms}.
\end{equation}
At the crossing point, this saddle-point approximation becomes nonanalytic in the large-$\Lambda$ limit. This nonanalyticity is the phase transition. The polynomial action itself is smooth; the nonanalyticity arises from selecting the dominant minimum.

In the softmax-attention theory, this is precisely the qualitative structure. The action $\mathcal{S}_{\rm softmax}(\hat a_G)$ has two competing minima: one with weak copy structure and one with strong copy structure. As $P$ increases, the strong-copy minimum becomes deeper. At the critical sample size $P^*_{\hat a_G}$, the global minimizer changes from one local minimum to the other and the MAP value of $\hat a_G$ jumps discontinuously.

This jump in the order parameter produces an abrupt loss drop. It also means there need not be a smooth precursor in the order parameter itself. The weak-copy phase can remain locally stable until the moment the global optimum switches to the copy phase. This is the sense in which the softmax transition is more abrupt and harder to anticipate from smooth progress measures.

\subsection{Crossovers}

A \emph{crossover} is a rapid but smooth change in behavior that is not a true phase transition. In a crossover, no quantity becomes nonanalytic in the large-system limit, and no symmetry is broken. Nevertheless, the system may show a pronounced change over a narrow range of the control parameter.

This is the interpretation of the later stage of linear-attention learning in the main text. The first true transition turns on the pooling order parameter $\hat c_1$. At larger $P$, the copy direction $\hat c_G$ becomes increasingly favorable and the attention pattern reorganizes from uniform pooling toward copying. This reorganization can be sharp in finite experiments and can produce a visible loss decrease.

\subsection{Phase Diagrams}

A \emph{phase diagram} shows which phase is realized as a function of control parameters. In this paper, we plot quantities such as the loss as a function of $P$ and $L$, or as a function of $P$ and $V$.

The theoretical phase boundaries are obtained by analyzing the reduced action. For linear attention, the first boundary is where the curvature in the pooling direction changes sign, giving the second-order transition at $P^*_{\hat c_1}$. The second boundary marks the scale where the copy direction becomes favorable and the crossover begins. For softmax attention, the phase boundary is determined by equality of the two competing minima of $\mathcal{S}_{\rm softmax}(\hat a_G)$.

The phase diagrams in the main text should therefore be read as maps of attention behavior: below the boundary, the model is in an unstructured or weakly structured phase; above it, the learned attention pattern implements the copy subcircuit.

\subsection{Interpretation for Attention and Emergence}

The statistical-physics terminology used in this paper can be summarized as follows. The attention matrix is a high-dimensional object, but the task symmetry identifies a small number of collective directions that matter. These directions are the order parameters. The Bayesian posterior over attention patterns induces an effective action over these order parameters. Because the action is multiplied by a large factor, the posterior concentrates near the minima of this action. As the number of training examples increases, the action changes shape. A phase transition occurs when the dominant saddle, equivalently the global MAP solution in the concentrated limit, changes qualitatively.

For linear attention, the action first undergoes a second-order transition into a pooling phase. The copy component then grows through a crossover. Thus the emergence of copying is preceded by an intermediate, partially useful context-aggregation regime. For softmax attention, the nonlinearity changes the effective action so that two minima coexist. The global minimum jumps directly to the copy phase, yielding a first-order transition and an abrupt loss drop.

This distinction is the main conceptual role of Landau theory in the paper. It explains why two attention mechanisms can solve the same task but display qualitatively different emergence profiles: the mechanism determines the structure of the reduced action, whose saddle points determine the order of the transition.

\section{Deriving the network prior as a function of the attention} \label{app:network_prior}

We are ultimately interested in the network posterior conditioned on the training data. Our approach follows a recent line of studies describing the network in terms of the feature kernels~\cite{fischer24a_pmlr, tiberiDissectingInterplayAttention2024, rubin25a_pmlr, vanmeegen2025coding, bauer2026unifiedtheoryfeaturelearning}. We start by deriving the network prior given inputs $\mX=(\vx^a_{\alpha})_{\alpha=1,\ldots,P;\,a=1,\dots,L}$ and network outputs ${\mF=(\vf^a_{\alpha})_{\alpha=1,\ldots,P;\,a=1,\dots,L}}$:
\begin{equation}
    p(\mF\vert \mX)=\int\,\prod_{\alpha,a}\,p(\vf^a_{\alpha}\vert \vx^a_{\alpha},\bm{\theta})\,p(\bm{\theta})\,d\bm{\theta}.
\end{equation}
For fixed network parameters $\bm{\theta} \coloneqq \{W^\mathrm{emb}, W^O, W^K, W^Q\}$, the probability $p(\mF\vert \mX,\bm{\theta})$ is given by enforcing the network model
\begin{equation}
\begin{aligned}
    \vh^a_{\alpha} &= \mW^{\mathrm{emb}} \vx^a_{\alpha} + \vp^a,
    & \mG^{ab}_{\alpha} &= \frac{1}{L\sqrt{\dmod}} \vh^{a\T}_{\alpha} \mW^{G} \vh^b_{\alpha},
    &\vf^a_{\alpha} &= \mW^O \mG^{ab}_{\alpha} \vh^b_{\alpha},
\end{aligned}
\end{equation}
with Dirac $\delta$-distributions as
\begin{equation}
    \begin{aligned}
        p(\mF\vert \mX,\bm{\theta})	&=\prod_{a, b, \alpha}\int d \vh^a_{\alpha}\,\delta(-\vh^a_{\alpha} + \mW^{\mathrm{emb}} \vx^a_{\alpha} + \vp^a)\\
	&\qquad\times\delta(-\mG^{ab}_{\alpha} + \frac{1}{L\sqrt{\dmod}} \vh^{a\T}_{\alpha} \mW^{G} \vh^b_{\alpha})\\
	&\qquad\times\delta(-\vf^a_{\alpha} + \mW^O \mG^{ab}_{\alpha} \vh^b_{\alpha}).
    \end{aligned}
\end{equation}

We use the Fourier representation of the Dirac $\delta$-distribution
\begin{equation}
    \delta(\vz)=\int\,\exp\big(\tilde{\vz}^{\T}\vz\big)\,d\tilde{\vz}
\end{equation}
with $\tilde{\vz}$ the conjugate variable to $\vz$ and $\tilde{\vz}^{\T}\vz=\sum_{i=1}^{N}\tilde{z}_{i}z_{i}$, where we integrate along the imaginary axis $\int d\tilde{\vz}=\prod_{k}\int_{i\mathbb{R}}\frac{d\tilde{z}_{k}}{2\pi i}$ for notational convenience, and rewrite
\begin{equation}
    \begin{aligned}
        p(\mF\vert \mX,\bm{\theta})	&=\int \D \vh \int \D \tilde{\vh}\,\exp(-\sum_{a,\alpha} \tilde{\vh}^{a\T}_{\alpha} \vh^a_{\alpha} + \sum_{a,\alpha} \tilde{\vh}^{a\T}_{\alpha} \mW^{\mathrm{emb}} \vx^a_{\alpha} + \sum_{a,\alpha}\tilde{\vh}^{a\T}_{\alpha} \vp^a)\\
	&\quad\times \int \D \mG \int \D \tilde{\mG} \exp(-\sum_{a b,\alpha} \tilde{\mG}^{ab\T}_{\alpha} \mG^{ab}_{\alpha} + \frac{1}{L\sqrt{d_k}} \sum_{a b,\alpha} \tilde{\mG}^{ab\T}_{\alpha}\vh^{a\T}_{\alpha} \mW^{G} \vh^b_{\alpha})\\
	&\quad\times \int \D \tilde{\vf} \exp(-\sum_{a,\alpha}\tilde{\vf}^{a\T}_{\alpha} \vf^a_{\alpha} + \sum_{a b,\alpha} \tilde{\vf}^{a\T}_{\alpha}\mW^O \mG^{ab}_{\alpha} \vh^b_{\alpha}),
    \end{aligned}
\end{equation}
where we have used the shorthands $\D \vh = \prod_{a, \alpha} d \vh^a_{\alpha}$, $\D \mG = \prod_{ab, \alpha} d \mG^{ab}_{\alpha}$, and similarly for the conjugate fields.

\subsection*{Embedding Layer}

Next, we perform the marginalization over network parameters. In the following, we denote the expectation value over the statistics of network parameters $W^{\circ}$ as $\left\langle \dots\right\rangle _{W^{\circ}}$. For the embedding layer, we have the following
\begin{align}
    p(\mH\vert \mX)	&= \int \D \tilde{\vh} \,\,p(\vh^a_{\alpha}\vert \vx^a_{\alpha},\mW^{\mathrm{emb}}, \vp)\,p(\mW^{\mathrm{emb}}, \vp)\,d\mW^{\mathrm{emb}} d \vp\\
    &=\int \D \tilde{\vh}\,\langle\exp(-\sum_{a,\alpha,i} \tilde{h}^{a}_{i,\alpha} h^a_{i,\alpha} + \sum_{a,\alpha,ij} \tilde{h}^{a}_{i,\alpha} W^{\mathrm{emb}}_{ij} x^a_{j, \alpha} + \sum_{a,\alpha,i}\tilde{h}^{a}_{i, \alpha} p^a_i)\rangle_{W^{\mathrm{emb}}_{ij}, p_i}\\
    &=\int \D \tilde{\vh}\,\exp(-\sum_{a,\alpha,i} \tilde{h}^{a}_{i,\alpha} h^a_{i,\alpha}) \,\langle\exp(\sum_{a,\alpha,ij} \tilde{h}^{a}_{i,\alpha} W^{\mathrm{emb}}_{ij} x^a_{j, \alpha})\rangle_{W^{\mathrm{emb}}_{ij}} \,\langle \exp(\sum_{a,\alpha,i}\tilde{h}^{a}_{i, \alpha} p^a_i)\rangle_{p_i}\\
    &=\int \D \tilde{\vh}\,\exp(-\sum_{a,\alpha,i} \tilde{h}^{a}_{i,\alpha} h^a_{i,\alpha}) \exp\bigg(\frac{1}{2}\frac{g_{\mathrm{emb}}}{V}\sum_{a b, \alpha \beta, ij} \tilde{h}^{a}_{i,\alpha} \tilde{h}^{b}_{i,\beta} x^a_{j, \alpha} x^b_{j, \beta}\bigg)\\
    & \qquad \times \exp\bigg(\frac{1}{2}g_p\sum_{a b, \alpha \beta, i}\tilde{h}^{a}_{i, \alpha} \delta_{ab} \tilde{h}^{b}_{i, \beta}\bigg)\nonumber \\
    &=\int \D \tilde{\vh}\,\exp(-\sum_{a,\alpha,i} \tilde{h}^{a}_{i,\alpha} h^a_{i,\alpha}) \exp\bigg(\frac{1}{2}\sum_{a b, \alpha \beta, ij} \tilde{h}^{a}_{i,\alpha} \tilde{h}^{b}_{i,\beta} \bigg[ \frac{g_{\mathrm{emb}}}{V} x^a_{j, \alpha} x^b_{j, \beta} + g_p \delta_{a b}\bigg]\bigg).
\end{align}
Integrating the conjugate variables $\tilde{\vh}$, we find that the token embeddings $h^a_{i,\alpha}$ are i.i.d. Gaussian distributed in model space $i$ with zero mean and covariance matrix
\begin{equation}
    \mC^{(xx)}_{a b, \alpha \beta}= g_{\mathrm{emb}} \sum_j x^a_{j, \alpha} x^b_{j, \beta} + g_p \delta_{a b}.
\end{equation}

\subsection*{Attention Layer}
\label{app:attention_prior}
The attention mechanism is given by
\begin{equation}
    \mG^{ab}_{\alpha} = \frac{1}{L \sqrt{\dmod}} \vh^{a\T}_{\alpha} \mW^{G} \vh^b_{\alpha}\,.
\end{equation}
We now marginalize with respect to a single network parameter $\mW^{\mG}$, yielding
\begin{align}
        p(\mG\vert \mH)	&=  \int \D \tilde{\mG} \, \Big\langle\exp\Big(\sum_{a b,\alpha}-\tilde{G}^{ab}_{\alpha} G^{ab}_{\alpha} + \sum_{a b,\alpha, ij} \frac{1}{L\sqrt{\dmod}} \tilde{G}^{ab}_{\alpha} h^{a}_{i,\alpha} W^{\mG}_{ij} h^b_{j, \alpha}\Big)\Big\rangle_{W^{\mG}_{ij}}\\
        &=  \int \D \tilde{\mG} \, \exp\Big(-\sum_{a b,\alpha}\tilde{G}^{ab}_{\alpha} G^{ab}_{\alpha}\Big) \, \exp\Big(\frac{1}{2}\frac{\dmod \, g_K\,g_Q}{d_\mathrm{model}^2} \frac{1}{L^2 \dmod} \sum_{a b,  c d, \alpha \beta} \tilde{G}^{ab}_{\alpha} \tilde{G}^{cd}_{\beta} (\vh^{a}_{\alpha} \cdot \vh^{c}_{\beta}) (\vh^b_{\alpha} \cdot \vh^d_{\beta})\Big)\\
        &=\N\big(\mG \mid 0,  C^{(\mG \mG)}_{ab, cd, \alpha \beta}\big) \label{eq:prior_G}
\end{align}
with covariance given by
\begin{equation}
    C^{(\mG \mG)}_{ab, cd, \alpha \beta} = \frac{g_K\,g_Q}{L^2 d_\mathrm{model}^2} (\vh^{a}_{\alpha} \cdot \vh^{c}_{\beta}) (\vh^b_{\alpha} \cdot \vh^d_{\beta})
\end{equation}
To decouple the appearing terms $(\vh^{a}_{\alpha} \cdot \vh^{c}_{\beta})$ in the covariance, we introduce an auxiliary variable as
\begin{equation}
    \mQ^{ab}_{\alpha \beta} \coloneqq \frac{1}{\dmod} \sum_i h^a_{i,\alpha} h^b_{i,\beta}. \label{eq:Q_def}
\end{equation}
To obtain the distribution of $\mQ$, we will make use of the G\"artner-Ellis theorem~\cite{Touchette09}. For this, we first compute its cumulant-generating function $\mathcal{W}_{\mQ}$ as
\begin{align}
    \mathcal{W}_{\mQ}(\tilde{\mQ}) &= \ln \int \D \vh \, \exp\Big(\tilde{Q}^{ab}_{\alpha \beta} \, \frac{1}{\dmod} \sum_i h^{a}_{i, \alpha} h^b_{i, \beta} \Big) \; \N(\vh \vert 0,\mC^{(xx)})\\
    &=\dmod \, \ln \int \D h \, \exp\Big(\tilde{Q}^{ab}_{\alpha \beta} \, \frac{1}{\dmod} h^{a}_{\alpha} h^b_{\beta} \Big) \; \N(h \vert 0,\mC^{(xx)})\\
    &=\dmod \, \ln \int \D h \, \exp\Big(-\frac{1}{2} h^{a}_{\alpha} \Big[ -\frac{2}{d_\mathrm{model}} \tilde{Q}^{ab}_{\alpha \beta} + \big(C^{(xx)}_{a b, \alpha \beta}\big)^{-1} \Big] h^b_{\beta}\Big)\\
    &\qquad\times \Big(\det \big(C^{(xx)}_{a b, \alpha \beta}\big) \Big)^{-1/2} \nonumber\\
    &=\dmod \, \ln \Big(\det \big(-\frac{2}{d_\mathrm{model}} \tilde{Q}^{ab}_{\alpha \beta} + \big(C^{(xx)}_{a b, \alpha \beta}\big)^{-1} \big)\Big)^{-1/2} \, \Big(\det \big(C^{(xx)}_{a b, \alpha \beta}\big) \Big)^{-1/2}\\
    &=-\frac{\dmod}{2} \ln\det\Big(\mI-\frac{2}{d_\mathrm{model}} \tilde{\mQ} \mC^{(xx)}\Big).
\end{align}
Since the cumulant-generating function has a scaling form $\mathcal{W}_{\mQ}(\tilde{\mQ})=\dmod \, \lambda(\tilde{\mQ}/\dmod)$ with $\lambda$ being a $\dmod$-independent function, we may obtain the distribution of the auxiliary variable $\mQ$ by a large deviation principle as
\begin{align}
    \ln p(\mQ \vert C^{(xx)}) &= \sup_{\tilde{\mQ}}\tr\big[\mQ^{\T}\tilde{\mQ}\big] - \mathcal{W}_{\mQ}(\tilde{\mQ})\\
    &\eqqcolon \Gamma(\mQ),
\end{align}
expressed in terms of the rate function $\Gamma$~\cite{Touchette09}.

We compute the rate function via the supremum condition
\begin{equation}
    0\stackrel{!}{=}Q_{\mu\nu}^{sz}-\frac{\dmod}{2}\Big[\mI-\frac{2}{d_\mathrm{model}} \tilde{\mQ} \mC^{(xx)}\Big]_{sa, \mu\alpha}^{-1}\,C_{a z, \alpha \nu}^{(xx)},
\end{equation}
yielding
\begin{equation}
    \tilde{\mQ}=\frac{\dmod}{2}\Big(\big[\mC^{(xx)}\big]^{-1}-\big[\mQ\big]^{-1}\Big).
\end{equation}
This gives for the rate function
\begin{align}
    \Gamma(\mQ) &=  \tr\Big[\mQ\frac{\dmod}{2}\Big(\big[\mC^{(xx)}\big]^{-1}-\big[\mQ\big]^{-1}\Big)\Big]\\
    &\quad+\frac{\dmod}{2}\ln\det\Big(I-\frac{2}{\dmod}\frac{\dmod}{2}\Big(\big[\mC^{(xx)}\big]^{-1}-\big[\mQ\big]^{-1}\Big)\mC^{\left(xx\right)}\Big)\nonumber\\
    &= \frac{\dmod}{2} \tr\Big(\mQ\big[\mC^{(xx)}\big]^{-1}-\mI\Big) + \frac{\dmod}{2} \ln\det \Big(\big[\mQ\big]^{-1}\mC^{\left(xx\right)}\Big).
\end{align}
For $\dmod\to\infty$, the first term dominates the rate function $\Gamma(\mQ)$ and $\mQ$ concentrates on $\mC^{(xx)}$, so we may replace $\mQ$ in \eqref{eq:prior_G} with $\mC^{(xx)}$. Thus, we obtain
\begin{align}
    p(\mG\vert \mX)	&= \int \D \tilde{\mG} \, \exp\Big(-\sum_{a b,\alpha}\tilde{G}^{ab}_{\alpha} G^{ab}_{\alpha}\Big) \, \exp\Big(\frac{1}{2} \, \frac{g_K\,g_Q}{L^2} \sum_{a b,  c d, \alpha \beta} \tilde{G}^{ab}_{\alpha} \tilde{G}^{cd}_{\beta} C^{(xx)}_{a c, \alpha \beta} C^{(xx)}_{bd, \alpha \beta} \Big)\\
    &= \exp\bigg(-\frac{1}{2} \sum_{a b,  c d, \alpha \beta} G^{ab}_{\alpha} \Big[\frac{g_K\,g_Q}{L^2} C^{(xx)}_{a c, \alpha \beta} C^{(xx)}_{bd, \alpha \beta}\Big]^{-1} G^{cd}_{\beta} - \ln \det \Big[\frac{g_K\,g_Q}{L^2} C^{(xx)}_{a c, \alpha \beta} C^{(xx)}_{bd, \alpha \beta}\Big]^{1/2}\bigg).\label{eq:prob_G_concentrated}
\end{align}
For notational brevity, we denote $g_W\coloneqq g_K g_Q$ in the following.

\subsection*{Attention Layer Parameterized with $W^K,W^Q$}
\label{app:attention_prior_KQ}
The attention mechanism is given by
\begin{equation}
    \mG^{ab}_{\alpha}
    =
    \frac{1}{L\sqrt{\dmod}}
    h^a_{i,\alpha}
    W^Q_{li}
    W^K_{lj}
    h^b_{j,\alpha}.
\end{equation}
We now marginalize with respect to the $W^K$ and $W^Q$,
\begin{align}
    p(\mG\vert \mH)
    &=
    \int \D \tilde{\mG}\,
    \Big\langle
    \exp\Big(
        -\sum_{ab,\alpha}
        \tilde{G}^{ab}_{\alpha}G^{ab}_{\alpha}
        +
        \frac{1}{L\sqrt{\dmod}}
        \sum_{ab,\alpha,ijl}
        \tilde{G}^{ab}_{\alpha}
        h^a_{i,\alpha}
        W^Q_{li}
        W^K_{lj}
        h^b_{j,\alpha}
    \Big)
    \Big\rangle_{\mW^Q,\mW^K}.
\end{align}
Using independence in the index $l$, this becomes
\begin{align}
    p(\mG\vert \mH)
    &=
    \int \D \tilde{\mG}\,
    \exp\Big(
        -\sum_{ab,\alpha}
        \tilde{G}^{ab}_{\alpha}G^{ab}_{\alpha}
        +
        \mathcal{W}_{\mG}(\tilde{\mG})
    \Big),
\end{align}
where
\begin{align}
    \mathcal{W}_{\mG}(\tilde{\mG})
    &=
    d_k
    \ln
    \Big\langle
    \Big\langle
    \exp\Big(
        \frac{1}{L\sqrt{\dmod}}
        \sum_{ab,\alpha,ij}
        \tilde{G}^{ab}_{\alpha}
        h^a_{i,\alpha}
        W^Q_{li}
        W^K_{lj}
        h^b_{j,\alpha}
    \Big)
    \Big\rangle_{\mW^K}
    \Big\rangle_{\mW^Q}.
\end{align}
Performing first the Gaussian integral over the row $W^K_{lj}$ gives
\begin{align}
    \mathcal{W}_{\mG}(\tilde{\mG})
    &=
    d_k
    \ln
    \Big\langle
    \exp\Big(
        \frac{g_K}{2L^2\dmod^2}
        \sum_{ab,cd,\alpha\beta,ijk}
        \tilde{G}^{ab}_{\alpha}
        \tilde{G}^{cd}_{\beta}
        h^a_{i,\alpha}
        W^Q_{li}
        h^b_{j,\alpha}
        h^d_{j,\beta}
        h^c_{k,\beta}
        W^Q_{lk}
    \Big)
    \Big\rangle_{\mW^Q}.
\end{align}
To decouple the appearing hidden-state contractions, we introduce an auxiliary variable as
\begin{equation}
    Q^{ab}_{\alpha\beta}
    \coloneqq
    \frac{1}{\dmod}
    \sum_i
    h^a_{i,\alpha}
    h^b_{i,\beta}.
    \label{eq:Q_def}
\end{equation}
Then
\begin{align}
    \mathcal{W}_{\mG}(\tilde{\mG})
    &=
    d_k
    \ln
    \Big\langle
    \exp\Big(
        \frac{g_K}{2L^2\dmod}
        \sum_{ab,cd,\alpha\beta,ik}
        \tilde{G}^{ab}_{\alpha}
        \tilde{G}^{cd}_{\beta}
        Q^{bd}_{\alpha\beta}
        h^a_{i,\alpha}
        W^Q_{li}
        h^c_{k,\beta}
        W^Q_{lk}
    \Big)
    \Big\rangle_{\mW^Q}.
\end{align}
The remaining Gaussian integral is over the feature-indexed row vector $W^Q_{li}$, and therefore gives a determinant over the feature indices $i,k$:
\begin{equation}
    \mathcal{W}_{\mG}(\tilde{\mG})
    =
    -\frac{d_k}{2}
    \ln\det_{ik}
    \left[
        \delta_{ik}
        -
        \frac{g_Qg_K}{L^2\dmod^2}
        \sum_{ab,cd,\alpha\beta}
        \tilde{G}^{ab}_{\alpha}
        Q^{bd}_{\alpha\beta}
        \tilde{G}^{cd}_{\beta}
        h^a_{i,\alpha}
        h^c_{k,\beta}
    \right].
    \label{eq:G_cgf_factorized_feature_det}
\end{equation}

We now rewrite this determinant in the token-sample space. Let the combined token-sample index be denoted by $(a\alpha)$, and define
\begin{equation}
    H_{i,a\alpha}
    \coloneqq
    h^a_{i,\alpha},
    \qquad
    \widehat{\tilde{G}}_{a\alpha,b\beta}
    \coloneqq
    \tilde{G}^{ab}_{\alpha}\delta_{\alpha\beta},
    \qquad
    Q_{a\alpha,b\beta}
    \coloneqq
    Q^{ab}_{\alpha\beta}.
\end{equation}
Then Eq.~\eqref{eq:G_cgf_factorized_feature_det} can be written in matrix form as
\begin{equation}
    \mathcal{W}_{\mG}(\tilde{\mG})
    =
    -\frac{d_k}{2}
    \ln\det_i
    \left[
        \mI
        -
        \frac{g_Qg_K}{L^2\dmod^2}
        \mH
        \widehat{\tilde{\mG}}
        \mQ
        \widehat{\tilde{\mG}}^\T
        \mH^\T
    \right].
\end{equation}
The matrix inside the feature-space determinant is symmetric. We factor it as
\begin{equation}
    \mH
    \widehat{\tilde{\mG}}
    \mQ
    \widehat{\tilde{\mG}}^\T
    \mH^\T
    =
    \left(
        \mH
        \widehat{\tilde{\mG}}
        \mQ^{1/2}
    \right)
    \left(
        \mH
        \widehat{\tilde{\mG}}
        \mQ^{1/2}
    \right)^\T .
\end{equation}
Using the Weinstein--Aronszajn identity,
\begin{equation}
    \det(\mI-\mA\mA^\T)
    =
    \det(\mI-\mA^\T\mA),
\end{equation}
together with
\begin{equation}
    \mH^\T\mH
    =
    \dmod\,\mQ,
\end{equation}
we obtain
\begin{align}
    \mathcal{W}_{\mG}(\tilde{\mG})
    &=
    -\frac{d_k}{2}
    \ln\det
    \left[
        \mI
        -
        \frac{g_Qg_K}{L^2\dmod}
        \mQ^{1/2}
        \widehat{\tilde{\mG}}^\T
        \mQ
        \widehat{\tilde{\mG}}
        \mQ^{1/2}
    \right].
    \label{eq:G_cgf_factorized_Q}
\end{align}

To obtain the distribution of $\mQ$, we will make use of the G\"artner-Ellis theorem~\cite{Touchette09}. For this, we first compute its cumulant-generating function $\mathcal{W}_{\mQ}$ as
\begin{align}
    \mathcal{W}_{\mQ}(\tilde{\mQ})
    &=
    \ln
    \int \D \vh\,
    \exp\Big(
        \tilde{Q}^{ab}_{\alpha\beta}
        \frac{1}{\dmod}
        \sum_i
        h^a_{i,\alpha}
        h^b_{i,\beta}
    \Big)
    \N(\vh\vert 0,\mC^{(xx)})
    \\
    &=
    \dmod
    \ln
    \int \D h\,
    \exp\Big(
        \tilde{Q}^{ab}_{\alpha\beta}
        \frac{1}{\dmod}
        h^a_{\alpha}
        h^b_{\beta}
    \Big)
    \N(h\vert 0,\mC^{(xx)})
    \\
    &=
    \dmod
    \ln
    \int \D h\,
    \exp\Big(
        -\frac{1}{2}
        h^a_{\alpha}
        \Big[
            -\frac{2}{\dmod}
            \tilde{Q}^{ab}_{\alpha\beta}
            +
            \big(C^{(xx)}_{ab,\alpha\beta}\big)^{-1}
        \Big]
        h^b_{\beta}
    \Big)
    \Big(
        \det \mC^{(xx)}
    \Big)^{-1/2}
    \\
    &=
    -\frac{\dmod}{2}
    \ln\det
    \Big(
        \mI
        -
        \frac{2}{\dmod}
        \tilde{\mQ}
        \mC^{(xx)}
    \Big).
\end{align}
Since the cumulant-generating function has a scaling form
\begin{equation}
    \lambda_\mQ = \frac{1}{\dmod}\mathcal{W}_{\mQ}(\dmod \tilde{\mQ})
    =
    -\frac{1}{2} \ln\det  \Big( \mI - 2 \tilde{\mQ}       \mC^{(xx)} \Big)
\end{equation}
we may obtain the distribution of the auxiliary variable $\mQ$ by a large deviation principle as
\begin{align}
    \ln p(\mQ\vert C^{(xx)})
    &=
    -\dmod \Gamma(\mQ),
    \\
    \Gamma(\mQ)
    &=
    \sup_{\tilde{\mQ}}
    \left\{
        \tr[\mQ^\T\tilde{\mQ}]
        -
        \lambda_{\mQ}(\tilde{\mQ})
    \right\}.
\end{align}
We compute the rate function $\Gamma$ by evaluating the supremum condition
\begin{equation}
    0
    =
    Q^{sz}_{\mu\nu}
    -
    \frac{\dmod}{2}
    \left[
        \mI
        -
        \frac{2}{\dmod}
        \tilde{\mQ}\mC^{(xx)}
    \right]^{-1}_{sa,\mu\alpha}
    \frac{2}{\dmod}
    C^{(xx)}_{az,\alpha\nu},
\end{equation}
yielding
\begin{equation}
    \tilde{\mQ}
    =
    \frac{\dmod}{2}
    \left(
        [\mC^{(xx)}]^{-1}
        -
        [\mQ]^{-1}
    \right).
\end{equation}
This gives for the rate function
\begin{align}
    \Gamma(\mQ)
    &=
    \frac{\dmod}{2}
    \tr\Big(
        \mQ[\mC^{(xx)}]^{-1}
        -
        \mI
    \Big)
    +
    \frac{\dmod}{2}
    \ln\det
    \Big(
        [\mQ]^{-1}
        \mC^{(xx)}
    \Big).
\end{align}
For $\dmod\to\infty$, $\mQ$ concentrates on $\mC^{(xx)}$, so we may replace $\mQ$ in Eq.~\eqref{eq:G_cgf_factorized_Q} with $\mC^{(xx)}$. Defining
\begin{equation}
    \mC_{a\alpha,b\beta}
    \coloneqq
    C^{(xx)}_{ab,\alpha\beta},
\end{equation}
we obtain
\begin{equation}
    \mathcal{W}_{\mG}(\tilde{\mG})
    =
    -\frac{d_k}{2}
    \ln\det
    \left[
        \mI
        -
        \frac{g_Qg_K}{L^2\dmod}
        \mC^{1/2}
        \widehat{\tilde{\mG}}^\T
        \mC
        \widehat{\tilde{\mG}}
        \mC^{1/2}
    \right].
    \label{eq:G_cgf_factorized_Cxx}
\end{equation}

We now evaluate the remaining integral over $\tilde{\mG}$ by the same large-deviation argument. Define
\begin{equation}
    \widehat{G}_{a\alpha,b\beta}
    \coloneqq
    G^{ab}_{\alpha}\delta_{\alpha\beta}.
\end{equation}
We define the "whiten" matrices
\begin{equation}
    \bar{\mG}
    \coloneqq
    \mC^{-1/2}
    \widehat{\mG}
    \mC^{-1/2},
    \qquad
    \bar{\tilde{\mG}}
    \coloneqq
    \mC^{1/2}
    \widehat{\tilde{\mG}}
    \mC^{1/2}.
\end{equation}
Then
\begin{equation}
    \sum_{ab,\alpha}
    \tilde{G}^{ab}_{\alpha}G^{ab}_{\alpha}
    =
    \tr[
        \bar{\tilde{\mG}}^\T
        \bar{\mG}
    ],
\end{equation}
and
\begin{equation}
    \mathcal{W}_{\mG}(\bar{\tilde{\mG}})
    =
    -\frac{d_k}{2}
    \ln\det
    \left[ \mI - \frac{g_Qg_K}{L^2\dmod} \bar{\tilde{\mG}}^\T    \bar{\tilde{\mG}}
    \right].
    \label{eq:G_cgf_factorized_whitened}
\end{equation}
Using $k=\frac{\dmod}{d_k}=O(1)$, where $k$ is often taken to be equal to the number of heads, so for our single-head attention we specialize to the case $k=1\Rightarrow \dmod =d_k\,\,$\footnote{the result is unchanged for any $k=O(1)$}. We can expand the cumulant-generating function in a power series using $\ln \det = \tr \ln$
\begin{equation}
    \frac{d_k}{2}
    \tr
    \left[ 
    \sum_{n=1}^\infty \frac{1}{n} \left( \frac{g_Qg_K}{L^2\dmod} \bar{\tilde{\mG}}^\T \bar{\tilde{\mG}} \right)^n
    \right]
    =
    \frac{d_k}{2} \left( \frac{g_Qg_K}{L^2\dmod} \tr\left[\bar{\tilde{\mG}}^\T \bar{\tilde{\mG}} \right] + O\left(\tr \left[\left( \frac{g_Qg_K}{L^2\dmod} \bar{\tilde{\mG}}^\T \bar{\tilde{\mG}} \right)^2 \right] \right) \right)
\end{equation}
So the expansion's control parameter is $\epsilon:=\frac{g_Qg_K}{L^2\dmod} \left\|\bar{\tilde{\mG}} \right\|_{op}^2$ and we may truncate the series at Gaussian order if it is small. We thus turn to estimate $\left\|\bar{\tilde{\mG}} \right\|_{op}^2$ self-consistently. At this Gaussian order we see 
\begin{equation}
    \bar{\mG} \sim \frac{g_Qg_K}{L^2} \quad \Rightarrow \quad \bar{\tilde{\mG}} \sim L \sqrt{\frac{1}{g_Qg_K}}.
\end{equation}
Since $\bar{\tilde{\mG}}$ is an $L \times L$ matrix it operator norm is of order
\begin{equation}
    \left\|\bar{\tilde{\mG}} \right\|_{op}^2 \sim \frac{L^3}{g_Qg_K}
\end{equation}
we find our expansion parameter $\epsilon$ is of order
\begin{equation}
    \epsilon \sim \frac{L}{\dmod}.
\end{equation}
We see that under our scaling, where $\dmod \gg L$ the expansion is consistent and we may approximate
\begin{equation}
    \mathcal{W}_{\mG}(\bar{\tilde{\mG}}) \approx \frac{d_k g_Qg_K}{2 L^2\dmod} \tr\left[\bar{\tilde{\mG}}^\T \bar{\tilde{\mG}} \right]
\end{equation}
which agrees with the central limit theorem intuition that when we sum a large number of modes $\dmod$ into a small matrix $L$ we should get Gaussian behavior. We conclude the prior distribution for the attention pattern $\mG$ is
\begin{align}
    p(\mG\vert \mX)
    &= \exp\bigg(-\frac{1}{2} \sum_{a b,  c d, \alpha \beta} G^{ab}_{\alpha} \Big[\frac{d_k g_K\,g_Q}{\dmod L^2} C^{(xx)}_{a c, \alpha \beta} C^{(xx)}_{bd, \alpha \beta}\Big]^{-1} G^{cd}_{\beta} - \ln \det \Big[\frac{d_k g_K\,g_Q}{\dmod L^2} C^{(xx)}_{a c, \alpha \beta} C^{(xx)}_{bd, \alpha \beta}\Big]^{1/2}\bigg).\label{eq:prob_G_concentrated}
\end{align}
For notational brevity, we denote $g_W\coloneqq g_K g_Q d_k/\dmod$ in the following.
\subsection*{Readout Layer}
Finally, we look at the network output and marginalize over the network parameters as
\begin{align}
    p(\mF\vert \mG)	&= \int \D \tilde{\vf} \, \bigg\langle\exp\bigg(-\sum_{a,\alpha, i}\tilde{f}^{a}_{i, \alpha} f^a_{i, \alpha} + \sum_{a b,\alpha, ij} \tilde{f}^{a}_{i, \alpha}W^O_{ij} G^{ab}_{\alpha} h^b_{j, \alpha}\bigg)\bigg\rangle_{W^O_{ij}}\\
    &= \int \D \tilde{\vf} \, \exp\bigg(-\sum_{a,\alpha, i}\tilde{f}^{a}_{i, \alpha} f^a_{i, \alpha}\bigg) \, \exp\bigg(\frac{1}{2} \frac{g_O}{\dmod}\sum_{a b c d,\alpha \beta, i} \tilde{f}^{a}_{i, \alpha} \tilde{f}^{c}_{i, \beta} \, G^{ab}_{\alpha} G^{cd}_{\beta} \, \sum_j h^b_{j, \alpha} h^d_{j, \beta}\bigg)\\
    &= \int \D \tilde{\vf} \, \exp\bigg(-\sum_{a,\alpha, i}\tilde{f}^{a}_{i, \alpha} f^a_{i, \alpha}\bigg) \, \exp\bigg(\frac{1}{2} g_O \sum_{a b c d,\alpha \beta, i} \tilde{f}^{a}_{i, \alpha} \tilde{f}^{c}_{i, \beta} \, G^{ab}_{\alpha} G^{cd}_{\beta} \, Q^{bd}_{\alpha\beta}\bigg),
\end{align}
where we have identified the last sum term as the auxiliary variable $\mQ$ defined in \eqref{eq:Q_def}. For $\dmod \gg 1$, $\mQ$ concentrates and we may replace by $\mC^{(xx)}$ as in \eqref{eq:prob_G_concentrated}, yielding
\begin{equation}
    p(\mF\vert \mG)	= \int \D \tilde{\vf} \, \exp\bigg(-\sum_{a,\alpha, i}\tilde{f}^{a}_{i, \alpha} f^a_{i, \alpha}\bigg) \, \exp\bigg(\frac{1}{2} g_O \sum_{a b c d,\alpha \beta, i} \tilde{f}^{a}_{i, \alpha} \tilde{f}^{c}_{i, \beta} \, G^{ab}_{\alpha} G^{cd}_{\beta} \, C^{(xx)}_{b d, \alpha\beta}\bigg).
\end{equation}

\section{Deriving the Network Posterior}
\label{app:network_posterior}
We are interested in the posterior attention matrix $\bar{\mG}$. For the posterior, we assume Gaussian noise on the network outputs $\vy_{\alpha}=\vf_{\alpha}+\bm{\xi}_\alpha$ with $\bm{\xi}_\alpha \stackrel{\text{i.i.d.}}{\sim}\N(0,\sigma^2\mI)$. Then, we obtain
\begin{align}
    p(\mG \vert \{\vx_\alpha, \vy_\alpha\}_{\alpha})
  &\propto p(\mG \vert \mX) \int \D \vf \, p(\{\vy_\alpha\} \vert \mF)\, p(\mF \vert \mG) \\
  &\propto p(\mG \vert \mX) \int \D \vf  \,
     \exp \Big(-\frac{1}{2\sigma^2} \sum_{\alpha}
     \Vert \vy_{\alpha} - \vf_\alpha \Vert_2^2 \Big) \, p(\mF \vert \mG)\\
    &= \int \D \vf \int \D \tilde{\vf} \, \exp \Big(-\frac{1}{2 \sigma^2} \sum_{\alpha} \Vert \vy_{\alpha}-\vf_\alpha \Vert_2^2 - \tilde{\vf}_{\alpha}^{\T} \vf_{\alpha}\Big)\\
    &\qquad\times\exp\Big(\frac{1}{2} g_O \sum_{a b c d,\alpha \beta, i} \tilde{f}^{a}_{i, \alpha} \tilde{f}^{c}_{i, \beta} \, G^{ab}_{\alpha} G^{cd}_{\beta} \, C^{(xx)}_{b d, \alpha\beta}\Big) \, p(\mG \vert \mX),
\end{align}
where we drop the normalization constant throughout since we are ultimately interested in the most likely values of the attention in the posterior.
Integrating out the network outputs $\vf$ in the first term yields
\begin{equation}
     \int \D \vf \, \exp \Big(-\frac{1}{2 \sigma^2} \sum_{\alpha} \Vert \vy_{\alpha}-\vf_\alpha \Vert_2^2 - \tilde{\vf}_{\alpha}^{\T} \vf_{\alpha}\Big) \propto \exp \Big(\sum_{\alpha} \frac{\sigma^2}{2} \tilde{\vf}^{\T}_{\alpha} \tilde{\vf}_{\alpha} - \tilde{\vf}^{\T}_{\alpha} \vy_{\alpha}\Big).
\end{equation}
Overall, this gives
\begin{align}
    p(\mG \vert \{\vx_\alpha, \vy_\alpha\}_{\alpha}) &\propto \int \D \tilde{\vf} \, \exp \Big(-\sum_{\alpha}\tilde{\vf}^{\T}_{\alpha} \vy_{\alpha}\Big)\\
    &\qquad\times\exp\bigg(\frac{1}{2} \sum_{a b c d,\alpha \beta, i} \tilde{f}^{a}_{i, \alpha} \, \Big[g_O G^{ab}_{\alpha} G^{cd}_{\beta} \, C^{(xx)}_{b d, \alpha\beta} + \sigma^2 \mI \Big] \tilde{f}^{c}_{i, \beta}\bigg) \, p(\mG \vert \mX)\\
    &= \exp\bigg(-\frac{1}{2} \sum_{a b c d,\alpha \beta, i} y^{a}_{i, \alpha} \, \Big[g_O G^{ab}_{\alpha} G^{cd}_{\beta} \, C^{(xx)}_{b d, \alpha\beta} + \sigma^2 \mI \Big]^{-1} y^{c}_{i, \beta} \nonumber\\
    &\qquad\qquad -\frac{1}{2} \ln\det \Big[g_O G^{ab}_{\alpha} G^{cd}_{\beta} \, C^{(xx)}_{b d, \alpha\beta} + \sigma^2 \mI \Big] \bigg) p(\mG \vert \mX).
\end{align}
Since we are interested in the dominant attention behavior of the posterior, we may focus on the negative exponent or action $\mathcal{S}(\mG)\coloneqq -\ln p(\mG \vert \{\vx_\alpha, \vy_\alpha\}_{\alpha})$. Inserting the result for $p(\mG \vert \mX)$ from \eqref{eq:prob_G_concentrated}, we have
\begin{equation}
    \label{eq:action_G}
    \begin{aligned}
        \mathcal{S}(\mG) &= \frac{1}{2} \sum_{a b c d,\alpha \beta, i} y^{a}_{i, \alpha} \, \Big[g_O G^{ab}_{\alpha} G^{cd}_{\beta} \, C^{(xx)}_{b d, \alpha\beta} + \sigma^2 \mI \Big]^{-1} y^{c}_{i, \beta} + \frac{1}{2} \ln\det \Big[g_O G^{ab}_{\alpha} G^{cd}_{\beta} \, C^{(xx)}_{b d, \alpha\beta} + \sigma^2 \mI \Big]\\
        &\quad+ \frac{1}{2} \sum_{a b,  c d, \alpha \beta} G^{ab}_{\alpha} \Big[\frac{g_W}{L^2} C^{(xx)}_{a c, \alpha \beta} C^{(xx)}_{bd, \alpha \beta}\Big]^{-1} G^{cd}_{\beta} + \frac{1}{2} \ln \det \Big[\frac{g_W}{L^2} C^{(xx)}_{a c, \alpha \beta} C^{(xx)}_{bd, \alpha \beta}\Big].
    \end{aligned}
\end{equation}
The first term in the log probability is the prior term, identical to the one found in NNGP theory~\cite{neal1996bayesian, lee2018_nngp}. The second term in the log probability is the likelihood term of the labels given the data.

We now look at the scaling of the different terms. We downscale the output weights, the attention weights and the noise by a factor $\chi \sim L$, similar to taking $\mu P$ scaling (so-called "rich regime")~\cite{Yang21_Neurips}, such that the transformation is 
\begin{equation}
\begin{aligned}
g_W &\to \frac{g_W}{\chi}, &\quad
g_O &\to \frac{g_O}{\chi}, &\quad
\sigma^2 &\to \frac{\sigma^2}{\chi}.
\end{aligned}    
\end{equation}
Intuitively, the scaling with $\chi$ suppresses fluctuations in directions that are irrelevant to performing the task. Consequently, we may neglect the term $\ln \det$, which is a direct result of the Gaussian fluctuations of $\tilde{\vf}$. This scaling also increases the overall scale of the action and thus the sharpness of the probability landscape in $\mG$.

Explicitly, we have with these scalings
\begin{equation}
    \begin{aligned}
        \mathcal{S}(\mG) &= \chi \frac{1}{2} \sum_{a b c d,\alpha \beta, i} y^{a}_{i, \alpha} \, \Big[g_O G^{ab}_{\alpha} G^{cd}_{\beta} \, C^{(xx)}_{b d, \alpha\beta} + \sigma^2 \mI \Big]^{-1} y^{c}_{i, \beta} + \frac{1}{2} \ln\det \Big[g_O G^{ab}_{\alpha} G^{cd}_{\beta} \, C^{(xx)}_{b d, \alpha\beta} + \sigma^2 \mI \Big]\\
        &\quad+ L^2 \chi \frac{1}{2} \sum_{a b,  c d, \alpha \beta} G^{ab}_{\alpha} \big[g_W C^{(xx)}_{a c, \alpha \beta} C^{(xx)}_{bd, \alpha \beta}\big]^{-1} G^{cd}_{\beta} + \frac{1}{2} \ln \det \big[g_W C^{(xx)}_{a c, \alpha \beta} C^{(xx)}_{bd, \alpha \beta}\big] + const.(\mG)
    \end{aligned}
\end{equation}

\section{Projecting the NNGP term onto the order parameters}
\label{app:nngp_projection}
We first look at the term
\begin{equation}\label{eq:target}
    \mathcal{S}_{\mathrm{NNGP}}(G)
    \;=\;
    \sum_{ab,\,cd,\,\alpha\beta}
    G^{ab}\;
    \bigl[\frac{g_W}{L^2}\mathcal{M}_{ab\alpha,\,cd\beta}\bigr]^{\dagger}\;
    G^{cd}\,,
\end{equation}
where $\mathcal{M}_{ab\alpha,\,cd\beta} = C^{(xx)}_{ac,\alpha\beta}\,C^{(xx)}_{bd,\alpha\beta}$
is the NNGP kernel built from the input kernel
$C^{(xx)}_{ac,\alpha\beta} = \sum_{j}x^{a}_{j\alpha}\,x^{c}_{j\beta}+\delta^{ac}$. We will project this operator onto the attention patterns $(\bm{1}_L, \mG)$ and in the end transform to the normalized attention patterns with coefficients $(\hat{c}_1, \hat{c}_G)$ as in \eqref{eq:ansatz} of the main text. We use the pseudo-inverse denoted by $\dagger$ here to account for zero eigenvalues of the matrix.

Since the attention patterns that we are interested in have no dependence on $\alpha$ or $\beta$, we can directly look at the summed pseudo-inverse
\begin{equation}
    \Sigma_{(ab),(cd)}
    \;=\;
    \sum_{\alpha,\beta=1}^{P}
    \bigl(\mathcal{M}^{\dagger}\bigr)_{ab\alpha,\,cd\beta}\,.
\end{equation}

Each data sample $\alpha$ is a sequence
$\sigma_\alpha\colon\{1,\dots,L\}\to\{1,\dots,V\}$, one-hot encoded as
$x^{a}_{j\alpha}=\delta_{j,\sigma_\alpha(a)}$.
The data overlap matrix $D^{(\alpha\beta)}_{ac}=\delta_{\sigma_\alpha(a),\sigma_\beta(c)}$
gives the input kernel as $C^{(\alpha\beta)}_{ac} = D^{(\alpha\beta)}_{ac}+\delta_{ac}$.
Introducing the embedding matrix $\Phi_\alpha = [\mathrm{Id}_L\mid X_\alpha]$, which extends each data vector $(X_\alpha)_{aj}=\delta_{j,\sigma_\alpha(a)}$ by the identity matrix $\mathrm{Id}_L$ to represent the positional encoding, we have $C^{(\alpha\beta)}=\Phi_\alpha\Phi_\beta^\mathsf{T}$,
and the $(\alpha,\beta)$-block of the forward operator takes the Kronecker form
\begin{equation}\label{eq:Sblock}
    \mathcal{S}_{\alpha\beta}
    \;=\;
    C^{(\alpha\beta)}\otimes C^{(\alpha\beta)}
    \;=\;
    (\Phi_\alpha\otimes\Phi_\alpha)\,(\Phi_\beta\otimes\Phi_\beta)^{\mathsf{T}}\,.
\end{equation}
Setting $\Psi_\alpha=\Phi_\alpha\otimes\Phi_\alpha$ and stacking over samples gives
$\mathcal{S}=\Psi\Psi^\mathsf{T}$.
The rank of $\mathcal{S}$ saturates at $(L+V-1)^2$ once $P\geq\lceil(L+V-1)^2/L^2\rceil$
(typically $P\geq2$--$4$), after which both $\mathcal{S}$ and $\Sigma$ are $P$-independent.

\paragraph{Symmetry constraints on $\Sigma$.}
For sequences drawn uniformly over the vocabulary, the data distribution is invariant under
permutations of token labels and of position indices that are applied simultaneously to rows and columns.
The summed pseudoinverse $\Sigma$ inherits these symmetries.
The most general $L^2\times L^2$ operator consistent with position-permutation symmetry and the exchange symmetry $(ab)\leftrightarrow(cd)$ is
\begin{equation}\label{eq:ansatz}
    \Sigma
    \;=\;
    \mathfrak{a}\,I_{L^2}
    \;+\;\mathfrak{b}\,\mJ_{L^2}
    \;+\;\mathfrak{c}\,
    \bigl(I_L\otimes\mJ_L\,+\,\mJ_L\otimes I_L\bigr),
\end{equation}
where $\mJ_L$ is the $L\times L$ all-ones matrix.
The three unknowns $(\mathfrak{a},\mathfrak{b},\mathfrak{c})$ are fixed by evaluating
$\mathcal{S}_{\mathrm{NNGP}}(G)=\operatorname{vec}(G)^\mathsf{T}\Sigma\,\operatorname{vec}(G)$
on three linearly independent test matrices and matching against the
brute-force pseudoinverse of the full $L^2P\times L^2P$ matrix,
giving $\mathfrak{a}=1$, $\mathfrak{b}={(L{+}V)^{-2}}$, $\mathfrak{c}=-(L{+}V)^{-1}$, and therefore
\begin{equation}\label{eq:Sigma_result}
    \Sigma_{(ab),(cd)}
    \;=\;
    \delta_{ac}\,\delta_{bd}
    \;-\;\frac{\delta_{ac}}{L{+}V}
    \;-\;\frac{\delta_{bd}}{L{+}V}
    \;+\;\frac{1}{(L{+}V)^2}
    \;=\;
    \Pi_{ac}\,\Pi_{bd}\,,
\end{equation}
with $\Pi=I_L-\frac{1}{L+V}\mJ_L$. We verified this result numerically for all $3\leq L\leq 11$, $2\leq V\leq\min(L,6)$, across $P\in\{1,\dots,25\}$ and multiple random seeds.

\noindent\emph{Remark.}
The summed pseudoinverse $\Sigma = \sum_{\alpha\beta}(\mathcal{M}^{\dagger})_{ab\alpha,cd\beta}$
is \emph{not} the inverse of the summed forward operator
$\bar{\mathcal{S}}^{-1} = \bigl(\sum_{\alpha\beta}\mathcal{S}_{\alpha\beta}\bigr)^{-1}$,
because pseudoinversion and summation over sample pairs $(\alpha,\beta)$ do
not commute.
The coefficients $(\mathfrak{a},\mathfrak{b},\mathfrak{c})$ are
determined by matching against the brute-force pseudoinverse of the full $L^2\!P\times L^2\!P$ matrix.
We were unable to derive the coefficients analytically and therefore resorted to numerics.

\subsection*{Determining the projections onto the attention patterns}
Using the expression for \eqref{eq:Sigma_result}, we compute $\mathcal{S}_{\mathrm{NNGP}}(G, G')=\sum_{abcd}G_{ab}\,\Sigma_{(ab),(cd)}\,G'_{cd}$ explicitly
\begin{align}
    \frac{g_W}{L^2}\mathcal{S}_{\mathrm{NNGP}}(G, G')
    &=\sum_{abcd}G_{ab}\Bigl[\delta_{ac}\delta_{bd}-\frac{\delta_{ac}}{L+V}-\frac{\delta_{bd}}{L+V}+\frac{1}{(L+V)^2}\Bigr]G'_{cd}\nonumber\\
    &=\sum_{ab}G_{ab} G'_{ab}
    -\frac{1}{L+V}\sum_{abd}G_{ab}\,G'_{ad}
    -\frac{1}{L+V}\sum_{abc}G_{ab}\,G'_{cb}\\
    &\qquad+\frac{1}{(L+V)^2}\Bigl(\sum_{ab}G_{ab}\Bigr) \Bigl(\sum_{ab}G'_{ab}\Bigr)\nonumber\\
    &=\sum_{ab} G_{ab} G'_{ab} -\frac{\sum_a r_a r'_a + \sum_b c_b c'_b}{L+V}+\frac{s s'}{(L+V)^2}\,,
\end{align}
where $r_a=\sum_{b}G_{ab}$ and $r'_a=\sum_{b}G'_{ab}$ are the row sums, $c_b=\sum_{a}G_{ab}$ and $c'_b=\sum_{a}G'_{ab}$ are the column sums,
and $s=\sum_{ab}G_{ab}$ and $s'=\sum_{ab}G'_{ab}$ are the total sums.

We use this expression to compute the projections onto the basis $\{\bm{1}_L,\;G^*\}$, yielding
\begin{equation}\label{eq:Mprime}
    \frac{1}{(L+V)^2}
    \begin{pmatrix}
    L^2V^2 & LV^2 \\
    LV^2 & L\bigl[(L+V)^2-(L+2V)\bigr]
    \end{pmatrix}.
\end{equation}

We project onto the attention patterns normalized with respect to the Frobenius norm $\frac{1}{L} \,\mathbf{1}_L, \frac{1}{\sqrt{L}}\,G^*$, so making the ansatz $G=\hat{c}_1 \frac{1}{L} \,\mathbf{1}_L+\hat{c}_G \frac{1}{\sqrt{L}}\,G^*$, we overall get
\begin{equation}\label{eq:QcJcS}
    \frac{1}{(L+V)^2}\Bigl[
    V^2\,\hat{c}_1^2
    \;+\;2\frac{V^2}{\sqrt{L}}\,\hat{c}_1\,\hat{c}_G
    \;+\;\bigl((L+V)^2-L-2V\bigr)\,\hat{c}_G^2
    \Bigr].
\end{equation}

\section{Projecting the Data Term onto the Order Parameters}
\label{app:data_term_projection}
We now look at the data term
\begin{equation}
    \vy^{\T} \Big[g_{O} \mG^{\T} \mC^{(xx)} \mG +\sigma^{2} \mI \Big]^{-1} \vy
\end{equation}
and make the ansatz $\mG=c_1 \bm{1}_L + c_G \mG^{\ast}$. In the end, we transform to the normalized operators basis with coefficients $(\hat{c}_1, \hat{c}_G)$.
Due to the contraction with the target values $\vy$, we will use its Fourier modes as a base. For the copy task, the targets are the inputs shifted by one, so we choose
\begin{align}
    \big[\vv_{0,0}\big]_{\alpha}^{a}&=\frac{1}{\sqrt{PL}}\frac{1}{L}\sum_{b}\sum_{j}x_{j,\alpha}^{b}=\frac{1}{\sqrt{PL}},\\
    \big[\vv_{0,k\geq1}\big]_{\alpha}^{a}&=\frac{1}{\sqrt{P L^2}}\sum_{b}\sum_{j}e^{i\frac{2\pi}{V}kj}x_{j,\alpha}^{b},\\
    \big[\vv_{1,k}\big]_{\alpha}^{a}&=\frac{1}{\sqrt{P L(1-L^{-1})}}\sum_{j}e^{i\frac{2\pi}{V}kj}\bigg(x_{j,\alpha}^{a-1}-\frac{1}{L}\sum_{b}x_{j,\alpha}^{b}\bigg).
\end{align}
Note that $\vv_{1, 0}=0$ and that the entries of $\big[\vv_{0,k}\big]^{a}$ are independent of $a$. The different normalization constants for $\big[\vv_{0,0}\big]^{a}_\alpha$ and $\big[\vv_{0,k\geq1}\big]^{a}_\alpha$ result from the application of the equivalent kernel approximation. For $k\geq 1$, we will use the Fourier orthogonality relation $\sum_j e^{i\frac{2\pi}{V}kj} = 0$ and the decomposition identity
\begin{equation}\label{eq:decomp}
    \sum_j e^{i\frac{2\pi}{V}kj}x_{j,\alpha}^{a-1} = \sqrt{PL(1-L^{-1})}\;[\vv_{1,k}]^a_\alpha + \sqrt{P}\;[\vv_{0,k}]^a_\alpha.
\end{equation}

Throughout, we will use the equivalent kernel approximation replacing $\sum_{\alpha} \circ \approx P \mE_x[\circ]$ and the one-hot expectations:
\begin{equation}
    \mE[x_{i,\beta}^a x_{j,\beta}^a] = \frac{\delta_{ij}}{V}, \qquad \mE[x_{i,\beta}^a x_{j,\beta}^b] = \frac{1}{V^2} \;\;(a\neq b), \qquad \mE[x_{j,\beta}^a] = \frac{1}{V}.
\end{equation}

\subsection*{Decomposition of labels}
We start by computing the spanning coefficients of $y$
\begin{align}
    \vy^{\T}_{i} \vv_{0, 0} &= \frac{1}{\sqrt{P L}} \sum_{\alpha, a}  x^{a-1}_{i \alpha} \approx \sqrt{P L} \frac{1}{V},\\
    \vy^{\T}_{i} \vv_{0, k\geq 1} &= \frac{1}{\sqrt{PL^2}}\sum_{\alpha,a} x_{i,\alpha}^{a-1}\sum_{j}e^{i\frac{2\pi}{V}kj}\sum_{b}x_{j,\alpha}^{b}\\
    &= \frac{1}{\sqrt{P}}\sum_{\alpha,a}\sum_{j}e^{i\frac{2\pi}{V}kj}\bigg(\frac{1}{L}\sum_{b}\delta^{a-1,b}x_{i,\alpha}^{a-1}x_{j,\alpha}^{a-1}+\big(1-\delta^{a-1,b}\big)x_{i,\alpha}^{a-1}x_{j,\alpha}^{b}\bigg)\\
    &\approx \frac{1}{\sqrt{P }}P\sum_{a}\sum_{j}e^{i\frac{2\pi}{V}kj} \frac{1}{L}\Big(\frac{\delta_{ij}}{V}+\left(L-1\right)\frac{1}{V^{2}}\Big) = \sqrt{P} e^{i\frac{2\pi}{V}ki}\frac{1}{V},\\
    \vy^{\T}_{i} \vv_{1, k\geq 1} &= \frac{1}{\sqrt{PL(1-L^{-1})}}\sum_{\alpha,a}\sum_{j}e^{i\frac{2\pi}{V}kj}x_{i,\alpha}^{a-1}\bigg(x_{j,\alpha}^{a-1}-\frac{1}{L}\sum_{b}x_{j,\alpha}^{b}\bigg)\\
    &= \frac{1}{\sqrt{PL(1-L^{-1})}}\sum_{\alpha,a}\sum_{j}e^{i\frac{2\pi}{V}kj}\bigg(x_{i,\alpha}^{a-1}x_{j,\alpha}^{a-1}\\
    &\qquad\qquad-\frac{1}{L}\sum_{b}\left(\delta^{a-1,b}x_{i,\alpha}^{a-1}x_{j,\alpha}^{a-1}+\left(1-\delta^{a-1,b}\right)x_{i,\alpha}^{a-1}x_{j,\alpha}^{b}\right)\bigg)\\
    &\approx \frac{1}{\sqrt{PL(1-L^{-1})}}P\sum_{a}\sum_{j}e^{i\frac{2\pi}{V}kj}\bigg(\frac{\delta_{ij}}{V}-\frac{1}{L}\bigg(\frac{\delta_{ij}}{V}+(L-1)\frac{1}{V^{2}}\bigg)\bigg)\\
    &=\sqrt{\frac{PL}{1-L^{-1}}}e^{i\frac{2\pi}{V}ki}\frac{1}{V}\bigg(1-\frac{1}{L}\bigg).
\end{align}

\subsection*{Decomposition of operator-matrix-contraction}
We look at the term $\mG^{\T} \mC^{(xx)} \mG$ and make the ansatz $\mG = c_1 \bm{1}_L + c_G \mG_{\ast}$. Inserting this ansatz, we have
\begin{equation}
    \sum_{bd}G^{ab}G^{cd}C^{(xx)}_{bd,\alpha\beta} = c_1^2\sum_{bd}C_{bd} + c_1c_G\sum_b C_{b,c-1} + c_Gc_1 \sum_d C_{a-1,d} + c_G^2 C_{a-1,c-1},
\end{equation}
corresponding to the four operator pairs $(\bm{1}_L,\bm{1}_L)$, $(\bm{1}_L,\mG^*)$, $(\mG^*,\bm{1}_L)$, $(\mG^*,\mG^*)$. We compute the action of each on the Fourier basis vectors $\vv_{0,0}$, $\vv_{0,k\geq1}$, $\vv_{1,k\geq1}$ from the right, i.e.\ we evaluate $\sum_{\beta,c}\sum_{bd}M_L^{ab}C^{(xx)}_{bd,\alpha\beta}M_R^{cd}[\vv]^c_\beta$.

\allowdisplaybreaks
We first look at the term corresponding to $(\bm{1}_L,\bm{1}_L)$, so the application of $\sum_{bd}C_{bd,\alpha\beta}$ to the Fourier basis.

For $\vv_{0,0}$, we have
\begin{align}
    \sum_{\beta, c} \sum_{bd} C^{(xx)}_{bd,\alpha \beta} \big[\vv_{0,0}\big]^c_{\beta} &= \sum_{\beta, c} \sum_{bd}\bigg(\sum_{i}x_{i,\alpha}^{b} x_{i,\beta}^{d}+\delta^{bd}\bigg) \frac{1}{\sqrt{P L}}\\
    &\approx \frac{P}{\sqrt{PL}}\sum_{c}\sum_{bd}\bigg(\sum_{i}x_{i,\alpha}^{b}\frac{1}{V}+\delta^{bd}\bigg)\\
    &= \frac{PL^{2}}{\sqrt{PL}}\bigg(\frac{L}{V}+1\bigg) = PL^{2}\bigg(\frac{L}{V}+1\bigg) \big[\vv_{0,0}\big]^a_{\alpha}.
\end{align}

For $\vv_{0,k\geq1}$, we get
\begin{align}
    \sum_{\beta, c} \sum_{bd}& C^{(xx)}_{bd,\alpha \beta} \big[\vv_{0,k\geq1}\big]^c_{\beta}\\
    &= \sum_{\beta,c}\sum_{bd}\bigg(\sum_{i}x_{i,\alpha}^{b}x_{i,\beta}^{d}+\delta^{bd}\bigg)\frac{1}{\sqrt{P L^2}}\sum_{a}\sum_{j}e^{i\frac{2\pi}{V}kj}x_{j,\beta}^{a} \\
    &= \frac{L}{\sqrt{PL^2}} \sum_{\beta}\bigg(\sum_{bd}\sum_{i}x_{i,\alpha}^{b}x_{i,\beta}^{d}\sum_{a}\sum_{j}e^{i\frac{2\pi}{V}kj}x_{j,\beta}^{a}+L\sum_{a}\sum_{j}e^{i\frac{2\pi}{V}kj}x_{j,\beta}^{a}\bigg)\\
    &=\frac{1}{\sqrt{P}}\sum_{\beta}\bigg(\sum_{bda}\sum_{ij}e^{i\frac{2\pi}{V}kj}x_{i,\alpha}^{b}\big(\delta^{ad}x_{i,\beta}^{a}x_{j,\beta}^{a}+\big(1-\delta^{ad}\big)x_{i,\beta}^{d}x_{j,\beta}^{a}\big)+ L \sum_{a}\sum_{j}e^{i\frac{2\pi}{V}kj}x_{j,\beta}^{a}\bigg)\\
    &\approx \frac{1}{\sqrt{P}}P\bigg(\sum_{b}\sum_{ij}e^{i\frac{2\pi}{V}kj}x_{i,\alpha}^{b}L\bigg(\frac{\delta_{ij}}{V}+\frac{L-1}{V^{2}}\bigg)+L\sum_{a}\sum_{j}e^{i\frac{2\pi}{V}kj}\frac{1}{V}\bigg)\\
    &=\sqrt{P} \bigg(\sum_{b}\sum_{j}e^{i\frac{2\pi}{V}kj}x_{j,\alpha}^{b}L\frac{1}{V}\bigg) = \frac{PL^{2}}{V} \big[\vv_{0,k\geq1}\big]^a_{\alpha},
\end{align}
where the $1/V^2$ and $1/V$ terms vanish by Fourier orthogonality for $k\geq 1$.

Finally, for $\vv_{1,k\geq1}$ it is
\begin{align}
    \sum_{\beta, c} \sum_{bd}& C^{(xx)}_{bd,\alpha \beta} \big[\vv_{1,k\geq1}\big]^c_{\beta}\\
    &= \sum_{\beta,c}\sum_{bd}\bigg(\sum_{i}x_{i,\alpha}^{b}x_{i,\beta}^{d}+\delta^{bd}\bigg)\frac{1}{\sqrt{PL (1-L^{-1})}}\sum_{j}e^{i\frac{2\pi}{V}kj}\bigg(x_{j,\beta}^{c-1}-\frac{1}{L}\sum_{a}x_{j,\beta}^{a}\bigg)\\
    &=\frac{1}{\sqrt{PL (1-L^{-1})}} \sum_{\beta,c}\sum_{j}e^{i\frac{2\pi}{V}kj}\bigg(\sum_{i}\bigg(\sum_{bd}x_{i,\alpha}^{b}x_{i,\beta}^{d}x_{j,\beta}^{c-1}-\frac{1}{L}\sum_{abd}x_{i,\alpha}^{b}x_{i,\beta}^{d}x_{j,\beta}^{a}\bigg)\nonumber\\
    &\qquad \qquad+L\bigg(x_{j,\beta}^{c-1}-\frac{1}{L}\sum_{a}x_{j,\beta}^{a}\bigg)\bigg)\\
    &\approx \frac{P}{\sqrt{PL (1-L^{-1})}} \sum_{c}\sum_{j}e^{i\frac{2\pi}{V}kj}\bigg(\sum_{i}\bigg(\sum_{bd}x_{i,\alpha}^{b}\bigg(\delta^{d,c-1}\frac{\delta_{ij}}{V}+\bigg(1-\delta^{d,c-1}\bigg)\frac{1}{V^{2}}\bigg)\nonumber\\
    &\qquad \qquad-\frac{1}{L}\sum_{abd}x_{i,\alpha}^{b}\left(\delta^{ad}\frac{\delta_{ij}}{V}+\left(1-\delta^{ad}\right)\frac{1}{V^{2}}\right)\bigg)+L\bigg(\frac{1}{V}-\frac{1}{L}\sum_{a}\frac{1}{V}\bigg)\bigg)\\
    &= \frac{P}{\sqrt{PL (1-L^{-1})}}\sum_{c}\sum_{ij}e^{i\frac{2\pi}{V}kj}\nonumber\\
    &\qquad\left(\sum_{b}x_{i,\alpha}^{b}\left(\frac{\delta_{ij}}{V}+\left(L-1\right)\frac{1}{V^{2}}\right)-\sum_{b}x_{i,\alpha}^{b}\left(\frac{\delta_{ij}}{V}+\left(L-1\right)\frac{1}{V^{2}}\right)\right)\\
    &=0.
\end{align}

Next, we consider the term from $(\bm{1}_L,\mG^*)$ corresponding to $\sum_b C_{b,c-1,\alpha\beta}$.

For $\vv_{0,0}$, we get
\begin{align}
    \sum_{\beta, c} \sum_{bd} C^{(xx)}_{bd,\alpha \beta} \delta^{c-1,d} \big[\vv_{0,0}\big]^c_{\beta} &= \sum_{\beta,c}\sum_{b}\bigg(\sum_{i}x_{i,\alpha}^{b}x_{i,\beta}^{c-1}+\delta^{b,c-1}\bigg)\frac{1}{\sqrt{PL}}\\
    &\approx \frac{P}{\sqrt{PL}}\sum_{bc}\bigg(\sum_{i}x_{i,\alpha}^{b}\frac{1}{V}+\delta^{b, c-1}\bigg)\\
    &= \frac{PL}{\sqrt{PL}}\Big(\frac{L}{V}+1\Big) = P L \Big(\frac{L}{V}+1\Big) \big[\vv_{0,0}\big]^a_{\alpha}.
\end{align}

For $\vv_{0,k\geq1}$, it is
\begin{align}
    \sum_{\beta, c} \sum_{bd}& C^{(xx)}_{bd,\alpha \beta} \delta^{c-1,d} \big[\vv_{0,k\geq1}\big]^c_{\beta}\\
    &= \sum_{\beta,c}\sum_{b}\bigg(\sum_{i}x_{i,\alpha}^{b}x_{i,\beta}^{c-1}+\delta^{b,c-1}\bigg)\frac{1}{\sqrt{P}}\frac{1}{L}\sum_{a}\sum_{j}e^{i\frac{2\pi}{V}kj}x_{j,\beta}^{a}\\
    &= \frac{1}{\sqrt{P}}\frac{1}{L}\sum_{\beta}\bigg(\sum_{bd}\sum_{i}x_{i,\alpha}^{b}x_{i,\beta}^{d}\sum_{a}\sum_{j}e^{i\frac{2\pi}{V}kj}x_{j,\beta}^{a} +L\sum_{a}\sum_{j}e^{i\frac{2\pi}{V}kj}x_{j,\beta}^{a}\bigg)\\
    &=\frac{1}{\sqrt{P}}\frac{1}{L}\sum_{\beta}\bigg(\sum_{bda}\sum_{ij}e^{i\frac{2\pi}{V}kj}x_{i,\alpha}^{b}\bigg(\delta^{ad}x_{i,\beta}^{a}x_{j,\beta}^{a}+\left(1-\delta^{ad}\right)x_{i,\beta}^{d}x_{j,\beta}^{a}\bigg)\nonumber\\
    &\qquad \qquad+L\sum_{a}\sum_{j}e^{i\frac{2\pi}{V}kj}x_{j,\beta}^{a}\bigg)\\
    &\approx \frac{1}{\sqrt{P}}\frac{1}{L} P\bigg(\sum_{b}\sum_{ij}e^{i\frac{2\pi}{V}kj}x_{i,\alpha}^{b}L\left(\frac{\delta_{ij}}{V}+\frac{L-1}{V^{2}}\right)+L\sum_{a}\sum_{j}e^{i\frac{2\pi}{V}kj}\frac{1}{V}\bigg)\\
    &= \frac{1}{\sqrt{P}}\frac{1}{L}LP\bigg(\sum_{b}\sum_{j}e^{i\frac{2\pi}{V}kj}x_{j,\alpha}^{b}\frac{1}{V}\bigg) = \frac{PL}{V} \big[\vv_{0,k\geq1}\big]^a_{\alpha}.
\end{align}

Applied to $\vv_{1,k\geq1}$, we have
\begin{align}
    \sum_{\beta, c} \sum_{bd}& C^{(xx)}_{bd,\alpha \beta} \delta^{c-1,d} \big[\vv_{1,k\geq1}\big]^c_{\beta} \\
    &=\sum_{\beta,c}\sum_{b}\bigg(\sum_{i}x_{i,\alpha}^{b}x_{i,\beta}^{c-1}+\delta^{b, c-1}\bigg)\frac{1}{\sqrt{PL(1-L^{-1})}}\sum_{j}e^{i\frac{2\pi}{V}kj}\bigg(x_{j,\beta}^{c-1}-\frac{1}{L}\sum_{a}x_{j,\beta}^{a}\bigg)\\
    &=\frac{1}{\sqrt{PL(1-L^{-1})}} \sum_{\beta,c}\sum_{j}e^{i\frac{2\pi}{V}kj}\bigg(\sum_{i}\bigg(\sum_{b}x_{i,\alpha}^{b}x_{i,\beta}^{c}x_{j,\beta}^{c}-\frac{1}{L}\sum_{ab}x_{i,\alpha}^{b}x_{i,\beta}^{c}x_{j,\beta}^{a}\bigg)\nonumber\\
    &\qquad\qquad+\bigg(x_{j,\beta}^{c}-\frac{1}{L}\sum_{a}x_{j,\beta}^{a}\bigg)\bigg),
\end{align}
where we relabelled $c-1\to c$. Applying the equivalent kernel:
\begin{align}
    &\approx \frac{P}{\sqrt{PL(1-L^{-1})}}\sum_{c}\sum_{j}e^{i\frac{2\pi}{V}kj}\bigg(\sum_{i}\bigg(\sum_{b}x_{i,\alpha}^{b}\frac{\delta_{ij}}{V}\nonumber\\
    &\qquad\qquad-\frac{1}{L}\sum_{ab}x_{i,\alpha}^{b}\left(\delta^{ac}\frac{\delta_{ij}}{V}+\Big(1-\delta^{ac}\right)\frac{1}{V^{2}}\Big)\bigg)+\left(\frac{1}{V}-\frac{1}{L}\sum_{a}\frac{1}{V}\right)\bigg)\\
    &=\frac{P}{\sqrt{PL(1-L^{-1})}}\sum_{c}\sum_{ij}e^{i\frac{2\pi}{V}kj}\left(\sum_{b}x_{i,\alpha}^{b}\frac{\delta_{ij}}{V}
    -\frac{1}{L}\sum_{b}x_{i,\alpha}^{b}\Big(\frac{\delta_{ij}}{V}+\left(L-1\right)\frac{1}{V^{2}}\Big)\right)\\
    &= \frac{P}{\sqrt{PL(1-L^{-1})}} \frac{L}{V}\Big(1-\frac{1}{L}\Big)\sum_{j}e^{i\frac{2\pi}{V}kj}\sum_{b}x_{j,\alpha}^{b}\\
    &= \frac{PL}{V}\sqrt{L\Big(1-\frac{1}{L}\Big)}\bigg(\frac{1}{\sqrt{P}}\frac{1}{L}\sum_{b}\sum_{j}e^{i\frac{2\pi}{V}kj}x_{j,\alpha}^{b}\bigg) = \frac{PL\sqrt{L-1}}{V} \big[\vv_{0,k\geq1}\big]^a_{\alpha}.
\end{align}

For the terms switched in order $(\mG^*,\bm{1}_L)$, we have $\delta^{a-1,b}\sum_d C_{bd,\alpha\beta}$.

For $\vv_{0,0}$, this yields
\begin{align}
    \sum_{\beta, c} \sum_{bd} \delta^{a-1,b} C^{(xx)}_{bd,\alpha \beta}  \big[\vv_{0,0}\big]^c_{\beta} &= \sum_{\beta,c}\sum_{d}\bigg(\sum_{i}x_{i,\alpha}^{a-1}x_{i,\beta}^{d}+\delta^{a-1,d}\bigg)\frac{1}{\sqrt{PL}}\\
    &\approx \frac{P}{\sqrt{PL}}\sum_{c}\sum_{d}\bigg(\sum_{i}x_{i,\alpha}^{a-1}\frac{1}{V}+\delta^{a-1,d}\bigg)\\
    &= \frac{PL}{\sqrt{PL}}\Big(\frac{L}{V}+1\Big) = PL\Big(\frac{L}{V}+1\Big) \big[\vv_{0,0}\big]^a_{\alpha}.
\end{align}

For $\vv_{0,k\geq1}$, it is
\begin{align}
    \sum_{\beta, c} \sum_{bd}& \delta^{a-1,b} C^{(xx)}_{bd,\alpha \beta} \big[\vv_{0,k\geq1}\big]^c_{\beta}\\
    &= \sum_{\beta,c}\sum_{d}\bigg(\sum_{i}x_{i,\alpha}^{a-1}x_{i,\beta}^{d}+\delta^{a-1,d}\bigg)\frac{1}{\sqrt{P}}\frac{1}{L}\sum_{a^\prime}\sum_{j}e^{i\frac{2\pi}{V}kj}x_{j,\beta}^{a^\prime}\\
    &= \frac{1}{\sqrt{P}}\sum_{\beta}\bigg(\sum_{d}\sum_{i}x_{i,\alpha}^{a-1}x_{i,\beta}^{d}\sum_{a^\prime}\sum_{j}e^{i\frac{2\pi}{V}kj}x_{j,\beta}^{a^\prime}+\sum_{a^\prime}\sum_{j}e^{i\frac{2\pi}{V}kj}x_{j,\beta}^{a^\prime}\bigg)\\
    &= \frac{1}{\sqrt{P}}\sum_{\beta}\bigg(\sum_{d a^\prime}\sum_{ij}e^{i\frac{2\pi}{V}kj}x_{i,\alpha}^{a-1}\big(\delta^{d a^\prime}x_{i,\beta}^{a^\prime}x_{j,\beta}^{a^\prime}+\big(1-\delta^{da^\prime}\big)x_{i,\beta}^{d}x_{j,\beta}^{a^\prime}\big)\nonumber\\
    &\qquad\qquad\qquad+\sum_{a^\prime}\sum_{j}e^{i\frac{2\pi}{V}kj}x_{j,\beta}^{a^\prime}\bigg)\\
    &\approx \frac{P}{\sqrt{P}} \bigg(\sum_{da^\prime}\sum_{ij}e^{i\frac{2\pi}{V}kj}x_{i,\alpha}^{a-1}\bigg(\delta^{da^\prime}\frac{\delta_{ij}}{V}+\Big(1-\delta^{da^\prime}\Big)\frac{1}{V^{2}}\bigg)+\sum_{a^\prime}\sum_{j}e^{i\frac{2\pi}{V}kj}\frac{1}{V}\bigg)\\
    &= \sqrt{P} \frac{L}{V}\sum_{j}e^{i\frac{2\pi}{V}kj}x_{j,\alpha}^{a-1}.
\end{align}
Now we decompose using~\eqref{eq:decomp}, inserting $x_{j,\alpha}^{a-1} = (x_{j,\alpha}^{a-1}-\frac{1}{L}\sum_{a^\prime}x_{j,\alpha}^{a^\prime})+\frac{1}{L}\sum_{a^\prime}x_{j,\alpha}^{a^\prime}$:
\begin{equation}
    = \frac{PL}{V}\sqrt{L(1-L^{-1})} \big[\vv_{1,k\geq1}\big]^a_{\alpha} +\frac{PL}{V} \big[\vv_{0,k\geq1}\big]^a_{\alpha}.
\end{equation}

For $\vv_{1,k\geq1}$, we have
\begin{align}
    \sum_{\beta, c} \sum_{bd}& \delta^{a-1,b} C^{(xx)}_{bd,\alpha \beta} \big[\vv_{1,k\geq1}\big]^c_{\beta}\\
    &= \sum_{\beta,c}\sum_{d}\bigg(\sum_{i}x_{i,\alpha}^{a-1}x_{i,\beta}^{d}+\delta^{a-1,d}\bigg)\frac{1}{\sqrt{PL(1-L^{-1})}}\sum_{j}e^{i\frac{2\pi}{V}kj}\bigg(x_{j,\beta}^{c-1}-\frac{1}{L}\sum_{a^\prime}x_{j,\beta}^{a^\prime}\bigg)\\
    &=\frac{1}{\sqrt{PL(1-L^{-1})}}\sum_{\beta,c}\sum_{j}e^{i\frac{2\pi}{V}kj}\bigg(\sum_{i}\bigg(\sum_{d}x_{i,\alpha}^{a-1}x_{i,\beta}^{d}x_{j,\beta}^{c-1}-\frac{1}{L}\sum_{da^{\prime}}x_{i,\alpha}^{a-1}x_{i,\beta}^{d}x_{j,\beta}^{a^{\prime}}\bigg)\nonumber\\
    &\qquad\qquad\qquad+\bigg(x_{j,\beta}^{c-1}-\frac{1}{L}\sum_{a^{\prime}}x_{j,\beta}^{a^{\prime}}\bigg)\bigg)\\
    &\approx \frac{P}{\sqrt{PL(1-L^{-1})}}\sum_{c}\sum_{j}e^{i\frac{2\pi}{V}kj}\bigg(\sum_{i}\bigg(x_{i,\alpha}^{a-1}\bigg(\frac{\delta_{ij}}{V}+\bigg(L-1\bigg)\frac{1}{V^{2}}\bigg)\nonumber\\
    &\qquad\qquad\qquad-\frac{1}{L}\sum_{a^{\prime}}x_{i,\alpha}^{a-1}\left(\frac{\delta_{ij}}{V}+\bigg(L-1\right)\frac{1}{V^{2}}\bigg)\bigg)+\bigg(\frac{1}{V}-\frac{1}{L}\sum_{a^{\prime}}\frac{1}{V}\bigg)\bigg)\\
    &=0,
\end{align}
since the two terms in the parentheses cancel identically after summing over $c$.

Finally, for $(\mG^*,\mG^*)$ we determine how $C_{a-1,c-1,\alpha\beta}$ acts onto the Fourier basis.

For $\vv_{0,0}$, we get
\begin{align}
    \sum_{\beta, c} \sum_{bd} \delta^{a-1,b} C^{(xx)}_{bd,\alpha \beta} \delta^{c-1,d}  \big[\vv_{0,0}\big]^c_{\beta} &= \sum_{\beta,c}\bigg(\sum_{i}x_{i,\alpha}^{a-1}x_{i,\beta}^{c-1}+\delta^{a-1,c-1}\bigg)\frac{1}{\sqrt{PL}}\\
    &\approx \frac{P}{\sqrt{PL}}\sum_{c}\bigg(\sum_{i}x_{i,\alpha}^{a-1}\frac{1}{V}+\delta^{a-1,c-1}\bigg)\\
    &= \frac{P}{\sqrt{PL}}\Big(\frac{L}{V}+1\Big) = P\Big(\frac{L}{V}+1\Big) \big[\vv_{0,0}\big]^a_{\alpha}.
\end{align}

For $\vv_{0,k\geq1}$, we write
\begin{align}
    \sum_{\beta, c} \sum_{bd}& \delta^{a-1,b} C^{(xx)}_{bd,\alpha \beta} \delta^{c-1,d} \big[\vv_{0,k\geq1}\big]^c_{\beta}\\
    &= \sum_{\beta,c}\bigg(\sum_{i}x_{i,\alpha}^{a-1}x_{i,\beta}^{c-1}+\delta^{a-1,c-1}\bigg)\frac{1}{\sqrt{PL^2}}\sum_{a^\prime}\sum_{j}e^{i\frac{2\pi}{V}kj}x_{j,\beta}^{a^\prime}\\
    &= \frac{1}{\sqrt{PL^2}}\sum_{\beta}\bigg(\sum_c\sum_{i}x_{i,\alpha}^{a-1}x_{i,\beta}^{c-1}\sum_{a^\prime}\sum_{j}e^{i\frac{2\pi}{V}kj}x_{j,\beta}^{a^\prime}+\sum_{a^\prime}\sum_{j}e^{i\frac{2\pi}{V}kj}x_{j,\beta}^{a^\prime}\bigg)\\
    &= \frac{1}{\sqrt{PL^2}}\sum_{\beta}\bigg(\sum_{c a^\prime}\sum_{ij}e^{i\frac{2\pi}{V}kj}x_{i,\alpha}^{a-1}\big(\delta^{c-1, a^\prime}x_{i,\beta}^{a^\prime}x_{j,\beta}^{a^\prime}\nonumber\\
    &\qquad\qquad\qquad+\big(1-\delta^{c-1, a^\prime}\big)x_{i,\beta}^{c-1}x_{j,\beta}^{a^\prime}\big)+\sum_{a^\prime}\sum_{j}e^{i\frac{2\pi}{V}kj}x_{j,\beta}^{a^\prime}\bigg)\\
    &\approx \frac{P}{\sqrt{P L^2}} \bigg(\sum_{c a^\prime}\sum_{ij}e^{i\frac{2\pi}{V}kj}x_{i,\alpha}^{a-1}\bigg(\delta^{c-1, a^\prime}\frac{\delta_{ij}}{V}+\Big(1-\delta^{c-1, a^\prime}\Big)\frac{1}{V^{2}}\bigg)+\sum_{a^\prime}\sum_{j}e^{i\frac{2\pi}{V}kj}\frac{1}{V}\bigg)\\
    &= \frac{P}{\sqrt{P L^2}} \frac{L}{V} \sum_{j}e^{i\frac{2\pi}{V}kj}x_{j,\alpha}^{a-1} = \frac{\sqrt{P}}{V}\sum_{j}e^{i\frac{2\pi}{V}kj}x_{j,\alpha}^{a-1}.
\end{align}
Decomposing via~\eqref{eq:decomp}:
\begin{equation}
    = \frac{P}{V}\sqrt{L(1-L^{-1})} \big[\vv_{1,k\geq1}\big]^a_{\alpha} +\frac{P}{V} \big[\vv_{0,k\geq1}\big]^a_{\alpha}.
\end{equation}

For $\vv_{1,k\geq1}$, it is
\begin{align}
    \sum_{\beta, c} \sum_{bd}& \delta^{a-1,b} C^{(xx)}_{bd,\alpha \beta} \delta^{c-1,d} \big[\vv_{1,k\geq1}\big]^c_{\beta}\\
    &= \sum_{\beta,c}\bigg(\sum_{i}x_{i,\alpha}^{a-1}x_{i,\beta}^{c-1}+\delta^{a-1,c-1}\bigg)\frac{1}{\sqrt{PL(1-L^{-1})}}\sum_{j}e^{i\frac{2\pi}{V}kj}\bigg(x_{j,\beta}^{c-1}-\frac{1}{L}\sum_{a^\prime}x_{j,\beta}^{a^\prime}\bigg)\\
    &=\frac{1}{\sqrt{PL(1-L^{-1})}}\sum_{\beta,c}\sum_{j}e^{i\frac{2\pi}{V}kj}\nonumber\\
    &\qquad\qquad\bigg(\sum_{i}\bigg(x_{i,\alpha}^{a-1}x_{i,\beta}^{c-1}x_{j,\beta}^{c-1}-\frac{1}{L}\sum_{a^{\prime}}x_{i,\alpha}^{a-1}x_{i,\beta}^{c-1}x_{j,\beta}^{a^{\prime}}\bigg)+\bigg(x_{j,\beta}^{c-1}-\frac{1}{L}\sum_{a^{\prime}}x_{j,\beta}^{a^{\prime}}\bigg)\bigg).
\end{align}
Applying the equivalent kernel approximation:
\begin{align}
    &\approx \frac{P}{\sqrt{PL(1-L^{-1})}}\sum_{c}\sum_{j}e^{i\frac{2\pi}{V}kj} \bigg(\sum_{i}\bigg(x_{i,\alpha}^{a-1}\frac{\delta_{ij}}{V}
    -\frac{1}{L}\sum_{a^{\prime}}x_{i,\alpha}^{a-1}\left(\delta^{c-1,a^{\prime}}\frac{\delta_{ij}}{V}+\bigg(1-\delta^{c-1,a^{\prime}}\right)\frac{1}{V^{2}}\bigg)\bigg)\nonumber\\
    &\qquad\qquad\qquad+\bigg(\frac{1}{V}-\frac{1}{L}\sum_{a^{\prime}}\frac{1}{V}\bigg)\bigg)\\
    &=\frac{P}{\sqrt{PL(1-L^{-1})}}\sum_{c}\sum_{ij}e^{i\frac{2\pi}{V}kj}\bigg(x_{i,\alpha}^{a-1}\frac{\delta_{ij}}{V} -\frac{1}{L}x_{i,\alpha}^{a-1}\bigg(\frac{\delta_{ij}}{V}+\left(L-1\right)\frac{1}{V^{2}}\bigg)\bigg)\\
    &=\frac{P}{\sqrt{PL(1-\frac{1}{L})}}\frac{L}{V}\bigg(1-\frac{1}{L}\bigg)\sum_{j}e^{i\frac{2\pi}{V}kj}x_{j,\alpha}^{a-1}.
\end{align}
Now decomposing $x_{j,\alpha}^{a-1}$ via~\eqref{eq:decomp}:
\begin{align}
    &= \frac{P}{\sqrt{PL(1-\frac{1}{L})}}\frac{L}{V}\bigg(1-\frac{1}{L}\bigg)\bigg(\sqrt{PL(1-L^{-1})}\;[\vv_{1,k}]^a_\alpha + \sqrt{P}\;[\vv_{0,k}]^a_\alpha\bigg)\\
    &= \frac{P(L-1)}{V}\big[\vv_{1,k\geq1}\big]^a_{\alpha} + \frac{P\sqrt{L-1}}{V}\big[\vv_{0,k\geq1}\big]^a_{\alpha}.
\end{align}

\subsection*{Explicit expression for the data term}

Collecting the results from all previous derivations, we have:
\begin{center}
\renewcommand{\arraystretch}{1.4}
\begin{tabular}{c|c|c|c}
Term & $\vv_{0,0}$ & $\vv_{0,k\geq1}$ & $\vv_{1,k\geq1}$ \\\hline
$(\bm{1}_L,\bm{1}_L)$ & $PL^2(\frac{L}{V}+1)\;\vv_{0,0}$ & $\frac{PL^2}{V}\;\vv_{0,k}$ & $0$ \\[3pt]
$(\bm{1}_L,\mG^*)$ & $PL(\frac{L}{V}+1)\;\vv_{0,0}$ & $\frac{PL}{V}\;\vv_{0,k}$ & $\frac{PL\sqrt{L-1}}{V}\;\vv_{0,k}$ \\[3pt]
$(\mG^*,\bm{1}_L)$ & $PL(\frac{L}{V}+1)\;\vv_{0,0}$ & $\frac{PL}{V}\;\vv_{0,k} + \frac{PL\sqrt{L-1}}{V}\;\vv_{1,k}$ & $0$ \\[3pt]
$(\mG^*,\mG^*)$ & $P(\frac{L}{V}+1)\;\vv_{0,0}$ & $\frac{P}{V}\;\vv_{0,k} + \frac{P\sqrt{L-1}}{V}\;\vv_{1,k}$ & $\frac{P\sqrt{L-1}}{V}\;\vv_{0,k} + \frac{P(L-1)}{V}\;\vv_{1,k}$
\end{tabular}
\end{center}

We now assemble the full kernel $g_O\sum_{bd}G^{ab}G^{cd}C^{(xx)}_{bd,\alpha\beta}$ in the Fourier basis. Since $G = c_1\bm{1}_L + c_G\mG^{\ast}$, we have
\begin{equation}
    G^{ab}G^{cd} = c_1^2 + c_1c_G\delta^{c-1,d} + c_Gc_1\delta^{a-1,b} + c_G^2\delta^{a-1,b}\delta^{c-1,d},
\end{equation}
so the kernel is a weighted sum of the four terms computed in the previous section.

For $k=0$, only $\vv_{0,0}$ contributes since $\vv_{1,0}=0$. Reading off the $\vv_{0,0}\to\vv_{0,0}$ eigenvalue from each term in the summary table, we get
\begin{align}
    \mathcal{K}_0 &= g_O\bigg[c_1^2 \cdot PL^2\Big(\frac{L}{V}+1\Big) + c_1c_G \cdot PL\Big(\frac{L}{V}+1\Big) + c_Gc_1 \cdot PL\Big(\frac{L}{V}+1\Big) + c_G^2 \cdot P\Big(\frac{L}{V}+1\Big)\bigg]\\
    &= g_O P\Big(\frac{L}{V}+1\Big)\big(c_1L + c_G\big)^2.
\end{align}
Switching to normalized operators with $\hat{c}_1=\frac{c_1}{L}$ and $\hat{c}_G=\frac{c_G}{\sqrt{L}}$, we get
\begin{equation}
    \mathcal{K}_0 = g_O P\Big(\frac{L}{V}+1\Big)\Big(\hat{c}_1 + \frac{\hat{c}_G}{\sqrt{L}}\Big)^2.
\end{equation}

For $k\geq 1$, the kernel acts on the two-dimensional space $(\vv_{0,k},\;\vv_{1,k})$. We can read off the $2\times 2$ matrix representation of each term from the summary table.
\paragraph{$(\bm{1}_L,\bm{1}_L)$:} $\vv_{0,k}\to \frac{PL^2}{V}\vv_{0,k}$, $\vv_{1,k}\to 0$, so
\begin{equation}
    \frac{P}{V}\begin{pmatrix} L^2 & 0 \\ 0 & 0\end{pmatrix}.
\end{equation}

\paragraph{$(\bm{1}_L,\mG^{\ast})$:} $\vv_{0,k}\to \frac{PL}{V}\vv_{0,k}$, $\vv_{1,k}\to \frac{PL\sqrt{L-1}}{V}\vv_{0,k}$, so
\begin{equation}
    \frac{P}{V}\begin{pmatrix} L & L\sqrt{L-1} \\ 0 & 0\end{pmatrix}.
\end{equation}

\paragraph{$(\mG^{\ast},\bm{1}_L)$:} $\vv_{0,k}\to \frac{PL}{V}\vv_{0,k} + \frac{PL\sqrt{L-1}}{V}\vv_{1,k}$, $\vv_{1,k}\to 0$, so
\begin{equation}
    \frac{P}{V}\begin{pmatrix} L & 0 \\ L\sqrt{L-1} & 0\end{pmatrix}.
\end{equation}

\paragraph{$(\mG^{\ast},\mG^{\ast})$:} $\vv_{0,k}\to \frac{P}{V}\vv_{0,k} + \frac{P\sqrt{L-1}}{V}\vv_{1,k}$, $\vv_{1,k}\to \frac{P\sqrt{L-1}}{V}\vv_{0,k} + \frac{P(L-1)}{V}\vv_{1,k}$, so
\begin{equation}
    \frac{P}{V}\begin{pmatrix} 1 & \sqrt{L-1} \\ \sqrt{L-1} & L-1\end{pmatrix}.
\end{equation}

The full kernel matrix for $k\geq 1$ is
\begin{align}
    \mathcal{K}_k &= g_O\bigg[c_1^2\frac{P}{V}\begin{pmatrix} L^2 & 0 \\ 0 & 0\end{pmatrix} + c_1c_G\frac{P}{V}\begin{pmatrix} L & L\sqrt{L-1} \\ 0 & 0\end{pmatrix}\nonumber\\
    &\qquad\qquad + c_Gc_1\frac{P}{V}\begin{pmatrix} L & 0 \\ L\sqrt{L-1} & 0\end{pmatrix} + c_G^2\frac{P}{V}\begin{pmatrix} 1 & \sqrt{L-1} \\ \sqrt{L-1} & L-1\end{pmatrix}\bigg]\\
    &= \frac{g_OP}{V}\begin{pmatrix} c_1^2L^2 + 2c_1c_GL + c_G^2 & c_1c_GL\sqrt{L-1} + c_G^2\sqrt{L-1} \\ c_Gc_1L\sqrt{L-1} + c_G^2\sqrt{L-1} & c_G^2(L-1)\end{pmatrix}.
\end{align}
Altogether, we obtain
\begin{equation}
    \mathcal{K}_k = \frac{g_OP}{V}\begin{pmatrix} (c_1L+c_G)^2 & (c_1L+c_G)c_G\sqrt{L-1} \\ (c_1L+c_G)c_G\sqrt{L-1} & c_G^2(L-1)\end{pmatrix} = \frac{g_OP}{V}\;\bm{u}\,\bm{u}^\T,
\end{equation}
where
\begin{equation}
    \bm{u} = \begin{pmatrix} c_1L + c_G \\ c_G\sqrt{L-1}\end{pmatrix} = \begin{pmatrix} \hat{c}_1 + \hat{c}_G/\sqrt{L} \\ \hat{c}_G\sqrt{(L-1)/L}\end{pmatrix}.
\end{equation}
We again replaced the bare operators by normalized operators $c_1 \bm{1}_L\mapsto \hat{c}_1 \frac{1}{L}\bm{1}_L$ and $c_G \mG^{\ast}\mapsto \hat{c}_G \frac{1}{\sqrt{L}}\mG^{\ast}$.
The final result is a rank-1 matrix, which will allow closed-form inversion via Sherman--Morrison.

\subsection*{Inversion of kernel using Sherman-Morrison}

We want to express
\begin{equation}
    S_{\mathrm{data}} = \frac{1}{2}\sum_{ac,\alpha\beta,i} y^a_{i,\alpha}\big[\mathcal{K} + \sigma^2\mI\big]^{-1}y^c_{i,\beta},
\end{equation}
which in the Fourier basis becomes a sum over modes.

First, we look at the contributions of modes with $k=0$. Since $\vv_{1,0}=0$, only the scalar mode $\vv_{0,0}$ contributes with
\begin{align}
    S_{\mathrm{data},0} &= \frac{1}{2}\sum_i \frac{|\vy_i^\T\vv_{0,0}|^2}{\mathcal{K}_0+\sigma^2} = \frac{1}{2}\cdot\frac{V\cdot PL/V^2}{\mathcal{K}_0+\sigma^2} = \frac{1}{2}\cdot\frac{PL/V}{\mathcal{K}_0+\sigma^2},
\end{align}
where we used $|\vy_i^\T\vv_{0,0}|^2 = PL/V^2$ (independent of $i$) and $\sum_i 1 = V$. Substituting $\mathcal{K}_0$:
\begin{equation}
    S_{\mathrm{data},0} = \frac{PL/V}{2\Big[g_OP\big(\hat{c}_1 + \hat{c}_G/\sqrt{L}\big)^2(L/V+1)+\sigma^2\Big]}.
\end{equation}

For modes with $k\geq 1$, the matrix to invert is $\mathcal{K}_k + \sigma^2\mI = \frac{g_OP}{V}\bm{u}\bm{u}^\T + \sigma^2\mI$. Since this is a rank-1 update of $\sigma^2\mI$, the Sherman--Morrison formula gives
\begin{equation}
    \big(\sigma^2\mI + A\bm{u}\bm{u}^\T\big)^{-1} = \frac{1}{\sigma^2}\mI - \frac{A}{\sigma^2(\sigma^2 + A|\bm{u}|^2)}\;\bm{u}\bm{u}^\T,
\end{equation}
where $A = g_OP/V$. We now compute $|\bm{u}|^2$:
\begin{align}
    |\bm{u}|^2 &= (c_1L+c_G)^2 + c_G^2(L-1)\\
    &= \Big(\hat{c}_1 + \frac{\hat{c}_G}{\sqrt{L}}\Big)^2 + \hat{c}_G^2\frac{L-1}{L}.
\end{align}

The quadratic form $\bm{y}_{i,k}^\dagger(\mathcal{K}_k+\sigma^2\mI)^{-1}\bm{y}_{i,k}$ becomes
\begin{align}
    \bm{y}_{i,k}^\dagger(\mathcal{K}_k+\sigma^2\mI)^{-1}\bm{y}_{i,k} &= \frac{1}{\sigma^2}|\bm{y}_{i,k}|^2 - \frac{A}{\sigma^2(\sigma^2+A|\bm{u}|^2)}|\bm{u}^\T\bm{y}_{i,k}|^2.
\end{align}
The first term gives
\begin{equation}
    |\bm{y}_{i,k}|^2 = \frac{P}{V^2}(1 + L-1) = \frac{PL}{V^2}.
\end{equation}
Next, the inner product $\bm{u}^\T\bm{y}_{i,k}$ yields
\begin{align}
    \bm{u}^\T\bm{y}_{i,k} &= \frac{\sqrt{P}}{V}e^{i\frac{2\pi}{V}ki}\Big[(c_1L+c_G)\cdot 1 + c_G\sqrt{L-1}\cdot\sqrt{L-1}\Big]\\
    &= \frac{\sqrt{P}}{V}e^{i\frac{2\pi}{V}ki}\big[c_1L + c_G + c_G(L-1)\big]\\
    &= \frac{\sqrt{P}}{V}e^{i\frac{2\pi}{V}ki}\cdot L(c_1 + c_G).
\end{align}
In normalized coordinates, $L(c_1+c_G) = L(\hat{c}_1/L + \hat{c}_G/\sqrt{L}) = \hat{c}_1 + \sqrt{L}\,\hat{c}_G$, so
\begin{equation}
    |\bm{u}^\T\bm{y}_{i,k}|^2 = \frac{P}{V^2}(\hat{c}_1 + \sqrt{L}\,\hat{c}_G)^2.
\end{equation}

Substituting into the quadratic form yields
\begin{align}
    \bm{y}_{i,k}^\dagger(\mathcal{K}_k+\sigma^2\mI)^{-1}\bm{y}_{i,k} &= \frac{PL}{V^2\sigma^2} - \frac{(g_OP/V)\cdot P(\hat{c}_1+\sqrt{L}\,\hat{c}_G)^2}{V^2\sigma^2(\sigma^2+(g_OP/V)|\bm{u}|^2)}.
\end{align}
This is independent of $i$ since $|e^{i2\pi ki/V}|^2=1$. Summing over $i=0,\ldots,V-1$ gives a factor of $V$, and summing over $k=1,\ldots,V-1$ gives $V-1$ identical terms (using that $\mathcal{K}_k$ is independent of $k$ for $k\geq 1$):
\begin{align}
    S_{\mathrm{data},k\geq 1} &= \frac{1}{2}\sum_{k=1}^{V-1}\sum_{i=0}^{V-1}\bm{y}_{i,k}^\dagger(\mathcal{K}_k+\sigma^2\mI)^{-1}\bm{y}_{i,k}\\
    &= \frac{(V-1)V}{2}\bigg(\frac{PL}{V^2\sigma^2} - \frac{g_OP^2(\hat{c}_1+\sqrt{L}\,\hat{c}_G)^2}{V^3\sigma^2(\sigma^2+\frac{g_OP}{V}|\bm{u}|^2)}\bigg)\\
    &= \frac{V-1}{2}\bigg(\frac{PL}{V\sigma^2} - \frac{g_OP^2(\hat{c}_1+\sqrt{L}\,\hat{c}_G)^2}{V^2\sigma^2(\sigma^2+\frac{g_OP}{V}|\bm{u}|^2)}\bigg).
\end{align}

Combining all terms, we get
\begin{equation}
S_{\mathrm{data}}(\hat{c}_1,\hat{c}_G) = \frac{PL/V}{2\big[g_OP(\hat{c}_1 + \hat{c}_G/\sqrt{L})^2(L/V+1)+\sigma^2\big]} + \frac{V-1}{2}\bigg(\frac{PL}{V\sigma^2} - \frac{g_OP^2(\hat{c}_1+\sqrt{L}\,\hat{c}_G)^2}{V^2\sigma^2\big(\sigma^2+\frac{g_OP}{V}|\bm{u}|^2\big)}\bigg),
\end{equation}
where $|\bm{u}|^2 = (\hat{c}_1 + \hat{c}_G/\sqrt{L})^2 + \hat{c}_G^2(L-1)/L$.

\section{Fluctuation Term}
\label{app:fluct_term}
In this section, we derive an expression for
\begin{equation}
    \ln\det \Big[g_O \sum_{b, d} G^{ab} G^{cd} \, C^{(xx)}_{b d, \alpha\beta} + \sigma^2 \mI \Big],
\end{equation}
in the basis $\mG=c_1 \bm{1}_L + c_G \mG^{\ast}$ and then transform to the normalized basis $(\hat{c}_1, \hat{c}_G)$ in the end. We denote the matrix inside the determinant as $\mathcal{K}+\sigma^2\mI$.

\subsection*{Identifying all eigenspaces and eigenvalue structure}

The Fourier basis $\{\vv_{0,0},\;\vv_{0,k\geq1},\;\vv_{1,k\geq1}\}$ used in the previous section spans a $(2V-1)$-dimensional subspace of $\mathbb{R}^{LP}$. For the fluctuation term, we must carefully check what happens on the orthogonal complement, because the $\delta^{bd}$ term in $C^{(xx)}_{bd,\alpha\beta} = \sum_i x^b_{i,\alpha}x^d_{i,\beta} + \delta^{bd}$ can produce non-trivial eigenvalues there.

Consider vectors of the form
\begin{equation}
    [w_m]^c_\beta = \frac{1}{\sqrt{P}}f_m(c), \qquad m=1,\ldots,L-1,
\end{equation}
that are constant in the data index $\beta$ while being traceless in the sequences index $\sum_c f_m(c) = 0$. These give $L-1$ independent directions and they are orthogonal to all Fourier basis vectors:
\begin{itemize}
    \item $\vv_{0,0}$: $\sum_{c,\beta} [w_m]^c_\beta [\vv_{0,0}]^c_\beta \propto \sum_c f_m(c) = 0$.
    \item $\vv_{0,k\geq1}$: $\sum_{c,\beta} [w_m]^c_\beta [\vv_{0,k}]^c_\beta \propto \sum_c f_m(c) \sum_\beta \sum_j e^{i\frac{2\pi}{V}kj}x^{a}_{j,\beta} \propto 0 \sum_\beta \sum_j e^{i\frac{2\pi}{V}kj}x^{a}_{j,\beta} = 0$.
    \item $\vv_{1,k\geq1}$: $\sum_{c,\beta} [w_m]^c_\beta [\vv_{1,k}]^c_\beta \propto \sum_\beta \sum_j e^{i\frac{2\pi}{V}kj}\sum_c f_m(c)(x^{c-1}_{j,\beta}-\frac{1}{L}\sum_a x^a_{j,\beta})$. Under the equivalent kernel, $\sum_\beta x^c_{j,\beta}\approx P/V$, so $\sum_\beta(x^{c-1}_{j,\beta}-\frac{1}{L}\sum_a x^a_{j,\beta})\approx 0$.
\end{itemize}

We now compute $[\mathcal{K}\,w_m]^a_\alpha = g_O\sum_{c,\beta}\sum_{bd}G^{ab}G^{cd}C^{(xx)}_{bd,\alpha\beta}[w_m]^c_\beta$, separating the two parts of $C^{(xx)}$.

For the data term $\sum_i x^b_{i,\alpha}x^d_{i,\beta}$, we get
\begin{align}
    &g_O\sum_{c,\beta}\sum_{bd}G^{ab}G^{cd}\bigg(\sum_i x^b_{i,\alpha}x^d_{i,\beta}\bigg)\frac{f_m(c)}{\sqrt{P}}\\
    &\qquad = \frac{g_O}{\sqrt{P}}\sum_{bd}G^{ab}\bigg(\sum_i x^b_{i,\alpha}\sum_\beta x^d_{i,\beta}\bigg)\sum_c G^{cd}f_m(c).
\end{align}
Under the equivalent kernel, $\sum_\beta x^d_{i,\beta}\approx P/V$, so the factor $\sum_i x^b_{i,\alpha}\cdot P/V = P x^b_{\cdot,\alpha}/V$ is independent of $d$. Then
\begin{equation}
    \sum_d\sum_c G^{cd}f_m(c) = \sum_c f_m(c)\sum_d G^{cd} = \sum_c f_m(c)\big(c_1 L + c_G\big) = (c_1L+c_G)\sum_c f_m(c) = 0,
\end{equation}
since $\sum_d G^{cd} = c_1L + c_G$ is independent of $c$ and $\sum_c f_m(c)=0$. So the data part vanishes.

For the identity term $\delta^{bd}$ part, we get
\begin{align}
    &g_O\sum_{c,\beta}\sum_{bd}G^{ab}G^{cd}\delta^{bd}\frac{f_m(c)}{\sqrt{P}} = \frac{g_O}{\sqrt{P}}\underbrace{\sum_\beta 1}_{=P}\sum_b G^{ab}\sum_c G^{cb}f_m(c).
\end{align}
We compute $\sum_b G^{ab}G^{cb}$ explicitly. Since $G^{ab} = c_1 + c_G\delta^{a-1,b}$:
\begin{align}
    \sum_b G^{ab}G^{cb} &= \sum_b (c_1 + c_G\delta^{a-1,b})(c_1 + c_G\delta^{c-1,b})\\
    &= c_1^2 L + c_1c_G + c_Gc_1 + c_G^2\delta^{a-1,c-1}\\
    &= c_1^2L + 2c_1c_G + c_G^2\delta_{ac}.
\end{align}
The first two terms are independent of $c$, so contracting with $f_m(c)$ with $\sum_c f_m(c)=0$ kills them:
\begin{equation}
    \sum_c(c_1^2L + 2c_1c_G + c_G^2\delta_{ac})f_m(c) = c_G^2 f_m(a).
\end{equation}
Therefore
\begin{equation}
    [\mathcal{K}\,w_m]^a_\alpha = g_O P\,c_G^2\cdot\frac{f_m(a)}{\sqrt{P}} = g_O P\,c_G^2\,[w_m]^a_\alpha,
\end{equation}
with eigenvalue $\lambda_{w} = g_O P c_G^2 = \frac{g_OP\hat{c}_G^2}{L}$.

\subsection*{Deriving the full log-determinant}

Taking together the results from the previous section, we have that the $LP$-dimensional space decomposes as:
\begin{center}
\renewcommand{\arraystretch}{1.4}
\begin{tabular}{l|c|c}
Subspace & Multiplicity & Eigenvalue of $\mathcal{K}+\sigma^2\mI$ \\\hline
$\vv_{0,0}$ & 1 & $\mathcal{K}_0 + \sigma^2$ \\[3pt]
$(\vv_{0,k},\vv_{1,k})$, non-trivial direction & $V-1$ & $\sigma^2 + \frac{g_OP}{V}|\bm{u}|^2$ \\[3pt]
$(\vv_{0,k},\vv_{1,k})$, trivial direction & $V-1$ & $\sigma^2$ \\[3pt]
$\vw_m$ & $L-1$ & $g_OP c_G^2 + \sigma^2$ \\[3pt]
Null space & $LP - L - 2V + 2$ & $\sigma^2$
\end{tabular}
\end{center}
We now switch to the normalized coordinates $c_1 \mapsto \frac{\hat{c}_1}{L}$ and $c_G \mapsto \frac{\hat{c}_G}{\sqrt{L}}$.

For $\vv_{0,0}$, we get a scalar:
\begin{equation}
    \ln(\mathcal{K}_0+\sigma^2) = \ln\bigg[g_OP\Big(\hat{c}_1+\frac{\hat{c}_G}{\sqrt{L}}\Big)^2\Big(\frac{L}{V}+1\Big)+\sigma^2\bigg].
\end{equation}
Next, we determine the $\vv_{i,k\geq 1}$ component. Each $2\times 2$ block has the form $\sigma^2\mI_2 + A\bm{u}\bm{u}^\T$ where $A = g_OP/V$. By the matrix determinant lemma:
\begin{equation}
    \det(\sigma^2\mI_2 + A\bm{u}\bm{u}^\T) = \det(\sigma^2\mI_2)\cdot\big(1 + A\,\bm{u}^\T(\sigma^2\mI_2)^{-1}\bm{u}\big) = \sigma^4\Big(1+\frac{A}{\sigma^2}|\bm{u}|^2\Big).
\end{equation}
Taking the logarithm and summing over $k=1,\ldots,V-1$ (the kernel is independent of $k$), we obtain
\begin{equation}
    \sum_{k=1}^{V-1}\ln\det\Big[\frac{g_OP}{V}\bm{u}\bm{u}^\T + \sigma^2\mI_2\Big] = (V-1)\bigg[2\ln\sigma^2 + \ln\Big(1+\frac{g_OP}{V\sigma^2}|\bm{u}|^2\Big)\bigg].
\end{equation}

For $\vw_m$, each of the $(L-1$ directions contributes the same term, yielding:
\begin{equation}
    (L-1)\ln\Big(\frac{g_OP\hat{c}_G^2}{L}+\sigma^2\Big).
\end{equation}
From the remaining $LP-L-2V+2$ directions of the null-space, each contributes $\ln\sigma^2$.
Combining all terms, we get
\begin{equation}
    \begin{aligned}
    \ln\det[\mathcal{K}+\sigma^2\mI] &= (LP-L)\ln\sigma^2 + \ln\bigg[g_OP\Big(\hat{c}_1+\frac{\hat{c}_G}{\sqrt{L}}\Big)^2\Big(\frac{L}{V}+1\Big)+\sigma^2\bigg]\\
    &+ (V-1)\ln\bigg(\sigma^2+\frac{g_OP}{V}|\bm{u}|^2\bigg) + (L-1)\ln\bigg(\frac{g_OP\hat{c}_G^2}{L}+\sigma^2\bigg),
    \end{aligned}
\end{equation}
where $|\bm{u}|^2 = (\hat{c}_1 + \hat{c}_G/\sqrt{L})^2 + \hat{c}_G^2(L-1)/L$.

The term $(LP-L)\ln\sigma^2$ is independent of $(\hat{c}_1,\hat{c}_G)$ and can be dropped. The term $(L-1)\ln(g_OP\hat{c}_G^2/L+\sigma^2)$ arises from the $\delta^{bd}$ part of $C^{(xx)}$ acting on directions that are constant in the data index $\alpha$ but traceless in the copy index $a$.

\section{Mapping the Posterior Through the Softmax}
\label{app:softmax_map}

We derive here the one-dimensional posterior for softmax attention theory, with $\sm$ explicitly defined as
\begin{equation}
    \sm^a(\vx) = \frac{e^{x^a}}{\sum_{a=1}^L e^{x^a}}.
\end{equation}
The starting point is the posterior over the pre-softmax attention scores from linear attention restricted to the two-dimensional order-parameter ansatz
\begin{equation}
    \mG
    =
    \hat c_1\,\frac{\bm 1_L}{L}
    +
    \hat c_G\,\frac{\mG^*}{\sqrt L},
    \qquad
    G^*_{ab}=\delta^{a-1,b}.
\end{equation}
For softmax attention, the actual attention matrix is the row-wise softmax of the score matrix. With the normalization used in the main text, this means
\begin{equation}
    \mA=\sm(L\mG).
\end{equation}
The posterior over $\mA$ is therefore the pushforward of the posterior over $\mG$,
\begin{align}
    p(\mA)
    &=
    \int d\hat c_1\,d\hat c_G\,
    \delta\!\left(\mA-\sm(L\mG(\hat c_1,\hat c_G))\right)
    p(\hat c_1,\hat c_G)
    \label{eq:softmax_pushforward_full}
\end{align}

\paragraph{Eliminating the uniform mode.}
The softmax is invariant under adding a constant to every entry in a row. In the ansatz above, the $\hat c_1$ direction shifts all entries in each row by the same amount, and therefore drops out of the softmax. Explicitly,
\begin{equation}
    \sm(LG)
    =
    \frac{e^{\sqrt L\hat c_G+\hat c_1}}
    {(L-1)e^{\hat c_1}+e^{\sqrt L\hat c_G+\hat c_1}},
\end{equation}
so that the common factor $e^{\hat c_1}$ cancels between numerator and denominator. Thus the softmax attention pattern depends only on $\hat c_G$. Thus, each row can be taken to have one copy entry, where $G_{ab}=\hat c_G/\sqrt{L}$, and $L-1$ non-copy entries, where $G_{ab}=0$, hence
\begin{equation}
    A_G(\hat c_G)
    =
    \frac{e^{\sqrt L\hat c_G}}
    {L-1+e^{\sqrt L\hat c_G}},
    \qquad
    A_1(\hat c_G)
    =
    \frac{1}{L-1+e^{\sqrt L\hat c_G}},
    \label{eq:softmax_entries}
\end{equation}
where $A_G$ denotes the attention weight on the copy position and $A_1$ the weight on each of the other $L-1$ positions. These variables obey the row-normalization constraint
\begin{equation}
    A_G+(L-1)A_1=1.
    \label{eq:AG_A1_constraint}
\end{equation}
Consequently, the pushforward posterior is supported on a one-dimensional manifold in the space of row-normalized attention matrices.

Let
\begin{equation}
    p(\hat c_G)
    =
    \int d\hat c_1\,p(\hat c_1,\hat c_G)
\end{equation}
be the marginal posterior over the softmax-relevant score direction. Then the induced distribution over the copy entry is
\begin{equation}
    p(A_G)
    =
    \int d\hat c_G\,
    \delta\!\left(
        A_G-
        \frac{e^{\sqrt L\hat c_G}}
        {L-1+e^{\sqrt L\hat c_G}}
    \right)
    p(\hat c_G)
    \label{eq:p_AG_delta}
\end{equation}
and,
\begin{equation}
    p(A_G,A_1)
    =
    \delta\!\left(A_1-\frac{1-A_G}{L-1}\right)
    p(A_G).
\end{equation}

\paragraph{Inverse map.}
The map in Eq.~\eqref{eq:softmax_entries} is monotone in $\hat c_G$, so we can invert it explicitly. Starting from
\begin{equation}
    A_G
    =
    \frac{e^{\sqrt L\hat c_G}}
    {L-1+e^{\sqrt L\hat c_G}},
\end{equation}
we obtain
\begin{equation}
    A_G(L-1)
    =
    e^{\sqrt L\hat c_G}(1-A_G),
\end{equation}
and therefore
\begin{equation}
    \hat c_G(A_G)
    =
    \frac{1}{\sqrt L}
    \log\!\left[
        \frac{(L-1)A_G}{1-A_G}
    \right].
    \label{eq:cG_of_AG}
\end{equation}
The corresponding Jacobian is
\begin{equation}
    \frac{d\hat c_G}{dA_G}
    = \frac{1}{
    \sqrt L\,A_G(1-A_G)},
\end{equation}
so the one-dimensional pushforward density can be written as
\begin{equation}
    p(A_G)
    =
    \frac{
        p(\hat c_G(A_G))
    }{
        \sqrt L\,A_G(1-A_G)
    }.
    \label{eq:p_AG_jacobian}
\end{equation}

\paragraph{Softmax order parameter.}
In the main text we express the post-softmax attention matrix in the same structural basis as the linear-attention ansatz
\begin{equation}
    \mA
    =
    \hat a_1\,\frac{\bm 1_L}{L}
    +
    \hat a_G\,\frac{\mG^*}{\sqrt L}.
    \label{eq:A_ansatz_softmax}
\end{equation}
Since $\mA$ is row-normalized, the two coefficients are not independent:
\begin{equation}
    \hat a_1+\frac{\hat a_G}{\sqrt L}=1,
    \qquad
    \hat a_1=1-\frac{\hat a_G}{\sqrt L}.
    \label{eq:a1_ag_normalization_app}
\end{equation}
The entries of the attention matrix are therefore
\begin{equation}
    A_G
    =
    \frac{\hat a_1}{L}+\frac{\hat a_G}{\sqrt L}
    =
    \frac{1+\frac{L-1}{\sqrt L}\hat a_G}{L},
    \qquad
    A_1
    =
    \frac{\hat a_1}{L}
    =
    \frac{1-\frac{1}{\sqrt L}\hat a_G}{L}.
    \label{eq:AG_A1_from_aG}
\end{equation}
Substituting Eq.~\eqref{eq:AG_A1_from_aG} into Eq.~\eqref{eq:cG_of_AG} gives the softmax inverse map used in the main text:
\begin{equation}
    \hat c_G(\hat a_G)
    =
    \frac{1}{\sqrt L}
    \log\!\left[
        \frac{
            1+\frac{L-1}{\sqrt L}\hat a_G
        }{
            1-\frac{1}{\sqrt L}\hat a_G
        }
    \right].
    \label{eq:cG_to_aG_app}
\end{equation}
The allowed range is
\begin{equation}
    -\frac{\sqrt L}{L-1}<\hat a_G<\sqrt L,
\end{equation}
which is exactly the condition that both $A_G$ and $A_1$ are positive.

For completeness, the Jacobian in the $\hat a_G$ coordinate is
\begin{equation}
    \frac{d\hat c_G}{d\hat a_G}
    =
    \frac{1}{
    \left(1+\frac{L-1}{\sqrt L}\hat a_G\right)
    \left(1-\frac{1}{\sqrt L}\hat a_G\right)
    }.
    \label{eq:jacobian_cG_aG}
\end{equation}
Thus the exact one-dimensional pushforward density contains a Jacobian contribution
\begin{align}
    p(\hat a_G)
    &\propto
    \exp\!\left[
        -\mathcal S_{\rm softmax}(\hat a_G)
    \right],\\
    \qquad
    \mathcal S_{\rm softmax}(\hat a_G)
    &=
    \mathcal S_{\rm prior}(\hat c_G(\hat a_G))
    +
    \mathcal S_{\rm data}(\hat a_1(\hat a_G),\hat a_G)
    -
    \log\left|
        \frac{d\hat c_G}{d\hat a_G}
    \right|.
    \label{eq:softmax_action_with_jacobian}
\end{align}
The Jacobian term is subleading in the large-$\chi L$ saddle-point limit because the prior and data terms scale as $\chi L$, while the Jacobian contributes only an additive logarithmic correction. We therefore drop it in the saddle-point analysis in the main text.

\paragraph{Softmax action.}
We now substitute the map above into the reduced action. The prior term is obtained from Eq.~\eqref{eq:NNGP_result} by setting $\hat c_1=0$, since $\hat c_1$ is a uniform mode that gets absorbed by the softmax and is fixed to zero by the prior at the saddle point. Using Eq.~\eqref{eq:cG_to_aG_app}, this gives
\begin{equation}
    \mathcal S_{\rm prior}^{\rm softmax}(\hat a_G)
    =
    \frac{\chi L}{2}
    \frac{(L+V)^2-L-2V}{g_W(L+V)^2}
    \log^2\!\left[
        \frac{
            1+\frac{L-1}{\sqrt L}\hat a_G
        }{
            1-\frac{1}{\sqrt L}\hat a_G
        }
    \right].
    \label{eq:softmax_prior_exact}
\end{equation}
The data term depends only on the post-softmax attention matrix itself. Therefore, in Eq.~\eqref{eq:Sdata} we substitute
\begin{equation}
    \hat c_1\mapsto \hat a_1=1-\frac{\hat a_G}{\sqrt L},
    \qquad
    \hat c_G\mapsto \hat a_G.
\end{equation}
Under this substitution,
\begin{equation}
    \hat a_1+\frac{\hat a_G}{\sqrt L}=1,
\end{equation}
so the uniform-mode contribution is independent of $\hat a_G$ and can be dropped when determining the saddle point. The nontrivial copy-sensitive part gives
\begin{equation}
    \mathcal S_{\rm data}^{\rm softmax}(\hat a_G)
    =
    -\frac{\chi L}{2}
    \frac{P}{\sigma^2}
    \frac{V-1}{V}
    \frac{
        \left(
            1+\sqrt L\,\hat a_G-\frac{\hat a_G}{\sqrt L}
        \right)^2 g_O
    }{
        \left(
            1+\frac{L-1}{L}\hat a_G^2
        \right)g_O
        +
        \frac{\sigma^2}{P}V
    },
    \label{eq:softmax_data_exact}
\end{equation}
up to terms independent of $\hat a_G$.

Combining Eqs.~\eqref{eq:softmax_prior_exact} and \eqref{eq:softmax_data_exact}, and dropping constants and the subleading Jacobian term, yields
\begin{equation}
\begin{aligned}
    \mathcal S_{\rm softmax}(\hat a_G)
    =
    \frac{\chi L}{2}
    \Bigg[
    &
    \frac{(L+V)^2-L-2V}{g_W(L+V)^2}
    \log^2\!\left(
        \frac{
            1+\frac{L-1}{\sqrt L}\hat a_G
        }{
            1-\frac{1}{\sqrt L}\hat a_G
        }
    \right)
    \\
    &
    -
    \frac{P}{\sigma^2}
    \frac{V-1}{V}
    \frac{
        \left(
            1+\sqrt L\,\hat a_G-\frac{\hat a_G}{\sqrt L}
        \right)^2 g_O
    }{
        \left(
            1+\frac{L-1}{L}\hat a_G^2
        \right)g_O
        +
        \frac{\sigma^2}{P}V
    }
    \Bigg].
\end{aligned}
\label{eq:softmax_action_exact}
\end{equation}
Finally, writing $\sigma^2=\bar\sigma^2 L$ and keeping the leading large-$L$ terms gives the expression in the main text:
\begin{equation}
    \mathcal S_{\rm softmax}(\hat a_G)
    =
    \frac{\chi L}{2}
    \left[
    \frac{1}{g_W}
    \log^2\!\left(
        \frac{1+\sqrt L\,\hat a_G}
        {1-\hat a_G/\sqrt L}
    \right)
    -
    \frac{P}{\bar\sigma^2 L}
    \frac{V-1}{V}
    \frac{
        \left(\hat a_G+\frac{1}{\sqrt L}\right)^2 g_O
    }{
        (\hat a_G^2+1)g_O
        +
        \frac{\bar\sigma^2 L}{P}V
    }
    +
    O\!\left(\frac{1}{L}\right)
    \right].
    \label{eq:softmax_action_large_L}
\end{equation}

\section{From the Reduced Action to the Training Loss}
\label{app:action_to_loss}

We start from the marginal likelihood of the labels before integrating out the
network outputs:
\begin{equation}
Z(\sigma^2)
\;\coloneqq\;
p(\{\vy_\alpha\}\mid \{\vx_\alpha\}_\alpha)
\propto
\int \D \mG \int \D \vf\;
p(\mG\mid \mX)\,p(\mF\mid \mG)\,
\exp\!\Bigl(
-\frac{1}{2\sigma^2}\sum_{\alpha=1}^{P}\|\vy_\alpha-\vf_\alpha\|_2^2
\Bigr).
\label{eq:app_loss_Z_full}
\end{equation}
Differentiating \eqref{eq:app_loss_Z_full} with respect to $\sigma^2$ gives
\begin{align}
\frac{\partial Z}{\partial \sigma^2}
&=
\int \D \mG \int \D \vf\;
p(\mG\mid \mX)\,p(\mF\mid \mG)\,
\frac{\partial}{\partial \sigma^2}
\exp\!\Bigl(
-\frac{1}{2\sigma^2}\sum_{\alpha=1}^{P}\|\vy_\alpha-\vf_\alpha\|_2^2
\Bigr)
\nonumber\\
&=
\frac{1}{2\sigma^4}
\int \D \mG \int \D \vf\;
\Bigl(\sum_{\alpha=1}^{P}\|\vy_\alpha-\vf_\alpha\|_2^2\Bigr)\,
p(\mG\mid \mX)\,p(\mF\mid \mG)\,
\exp\!\Bigl(
-\frac{1}{2\sigma^2}\sum_{\alpha=1}^{P}\|\vy_\alpha-\vf_\alpha\|_2^2
\Bigr).
\end{align}
Dividing by $Z(\sigma^2)$, we obtain
\begin{equation}
\sigma^4 \frac{\partial}{\partial \sigma^2}\ln Z(\sigma^2)
=
\frac12
\left\langle
\sum_{\alpha=1}^{P}\|\vy_\alpha-\vf_\alpha\|_2^2
\right\rangle,
\label{eq:app_loss_master}
\end{equation}
where $\langle \cdots \rangle$ denotes the posterior average with respect to the
measure in \eqref{eq:app_loss_Z_full}. Thus, differentiating the log-marginal
likelihood with respect to $\sigma^2$ brings down the quadratic data-fit term.

\paragraph{Relation to the reduced action.}
After integrating out $\mF$ and projecting onto the two-dimensional ansatz
\begin{equation}
\mG
=
\hat c_1\,\frac{\bm{1}_L}{L}
+
\hat c_G\,\frac{\mG^*}{\sqrt L},
\end{equation}
the same marginal likelihood takes the form
\begin{equation}
Z(\sigma^2)
\propto
\int d\hat c_1\, d\hat c_G\;
e^{-\mathcal S(\hat c_1,\hat c_G;\sigma^2)}.
\label{eq:app_loss_Z_reduced}
\end{equation}
Differentiating \eqref{eq:app_loss_Z_reduced} yields
\begin{equation}
\frac{\partial}{\partial \sigma^2}\ln Z(\sigma^2)
=
\frac{
\int d\hat c_1\, d\hat c_G\;
\bigl(\partial_{\sigma^2}\mathcal S(\hat c_1,\hat c_G;\sigma^2)\bigr)
e^{-\mathcal S(\hat c_1,\hat c_G;\sigma^2)}
}{
\int d\hat c_1\, d\hat c_G\;
e^{-\mathcal S(\hat c_1,\hat c_G;\sigma^2)}
}
=
\Bigl\langle \partial_{\sigma^2}\mathcal S \Bigr\rangle_{\hat c_1,\hat c_G},
\label{eq:app_loss_average}
\end{equation}
where the average is now taken with respect to the reduced posterior over
$(\hat c_1,\hat c_G)$.

In the large-$\chi$ limit used in the main text, the reduced posterior is sharply
peaked and the integral \eqref{eq:app_loss_Z_reduced} is dominated by its saddle
point. Writing
\begin{equation}
(\hat c_1^\star,\hat c_G^\star)
=
\arg\min_{\hat c_1,\hat c_G}\mathcal S(\hat c_1,\hat c_G;\sigma^2),
\end{equation}
we have, to leading order in the saddle-point approximation,
\begin{equation}
\ln Z(\sigma^2)
\simeq
\mathcal S(\hat c_1^\star,\hat c_G^\star;\sigma^2),
\qquad
\frac{\partial}{\partial \sigma^2}\ln Z(\sigma^2)
\simeq
\frac{d}{d\sigma^2}\mathcal S(\hat c_1^\star,\hat c_G^\star;\sigma^2).
\label{eq:app_loss_saddle}
\end{equation}
Using the chain rule,
\begin{equation}
\frac{d}{d\sigma^2}\mathcal S(\hat c_1^\star,\hat c_G^\star;\sigma^2)
=
\frac{\partial \mathcal S}{\partial \sigma^2}
+
\frac{\partial \mathcal S}{\partial \hat c_1}\frac{d\hat c_1^\star}{d\sigma^2}
+
\frac{\partial \mathcal S}{\partial \hat c_G}\frac{d\hat c_G^\star}{d\sigma^2}.
\label{eq:app_loss_chain_rule}
\end{equation}
At the saddle point, however,
\begin{equation}
\left.\frac{\partial \mathcal S}{\partial \hat c_1}\right|_{(\hat c_1^\star,\hat c_G^\star)}=0,
\qquad
\left.\frac{\partial \mathcal S}{\partial \hat c_G}\right|_{(\hat c_1^\star,\hat c_G^\star)}=0,
\end{equation}
so the implicit derivatives of the order parameters drop out and
\begin{equation}
\frac{d}{d\sigma^2}\mathcal S(\hat c_1^\star,\hat c_G^\star;\sigma^2)
=
\left.
\frac{\partial \mathcal S(\hat c_1,\hat c_G;\sigma^2)}{\partial \sigma^2}
\right|_{(\hat c_1,\hat c_G)=(\hat c_1^\star,\hat c_G^\star)}.
\label{eq:app_loss_no_inner_derivatives}
\end{equation}

Combining \eqref{eq:app_loss_master}, \eqref{eq:app_loss_saddle}, and
\eqref{eq:app_loss_no_inner_derivatives}, we obtain the relation
\begin{equation}
\frac12
\left\langle
\sum_{\alpha=1}^{P}\|\vy_\alpha-\vf_\alpha\|_2^2
\right\rangle
\simeq
\sigma^4
\left.
\frac{\partial \mathcal S(\hat c_1,\hat c_G;\sigma^2)}{\partial \sigma^2}
\right|_{(\hat c_1,\hat c_G)=(\hat c_1^\star,\hat c_G^\star)}.
\label{eq:app_loss_final_half}
\end{equation}
Under the equivalent-kernel (EK) approximation ($\sum_\alpha\circ\approx P\,\mathbb{E}_x[\circ]$), which is valid when $\sigma^2 \gg \sum_a (\vy^a-\vf^a(\vx))^2$, the generalization gap is small~\cite{silverman1984_ek, Ringel25_review} and we find
\begin{equation}
\frac12
\left\langle
\mathbb{E}_{\vx} \|\vy(\vx)-\vf(\vx)\|_2^2
\right\rangle
\simeq
\frac{\sigma^4}{P}
\left.
\frac{\partial \mathcal S(\hat c_1,\hat c_G;\sigma^2)}{\partial \sigma^2}
\right|_{(\hat c_1,\hat c_G)=(\hat c_1^\star,\hat c_G^\star)}.
\label{eq:app_loss_final_half}
\end{equation}

\section{Probability Distribution of Gaussian Matrix Product $W^{Q\T} W^K$}
\label{app:dist_matrix_product}
The probability distribution of the matrix product $\mW^{\mG}=\mW^{Q\T}\mW^{K}$ in the attention layer is given by
\begin{align}
    p(\mW^{\mG})&=\int \D \mW^Q \int \D \mW^K \, \prod_{nm} \delta \big(W^{\mG}_{nm}-\sum_l W_{ln}^{Q}W_{lm}^{K}\big) \, p(\mW^K) \, p(\mW^Q)\\
    &=\int \D \tilde{\mW}^{\mG} \int \D \mW^Q \int \D \mW^K \exp \big(\sum_{nm}\tilde{W}^{\mG}_{nm} \big(W^{\mG}_{nm}-\sum_l W_{ln}^{Q}W_{lm}^{K}\big)\big) \, p(\mW^K) \, p(\mW^Q)\\
    &=\int \D \tilde{\mW}^{\mG}  \exp \Big(\sum_{nm}\tilde{W}^{\mG}_{nm} W^{\mG}_{nm}-\ln\Big\langle\exp\Big(\sum_{nml} \tilde{W}^{\mG}_{nm} W_{ln}^{Q} W_{lm}^{K}\Big)\Big\rangle_{\mW^Q, \mW^K}\Big),
\end{align}
where we rewrite the $\delta$-function using its Fourier representation as in previous sections.

We identify the second term as the cumulant-generating function of $\mW^{\mG}$ and calculate it to obtain its statistics
\begin{align}
    \mathcal{W}(\tilde{\mW}^{\mG}) &= \ln\Big\langle\exp\Big(\sum_{nml} \tilde{W}^{\mG}_{nm} W_{ln}^{Q} W_{lm}^{K}\Big)\Big\rangle_{\mW^Q, \mW^K}\\
    &= \ln\prod_l\Big\langle\Big\langle\exp\Big(\sum_{nm} \tilde{W}^{\mG}_{nm} W_{ln}^{Q} W_{lm}^{K}\Big)\Big\rangle_{\mW^K}\Big\rangle_{\mW^Q}\\
    &= \ln\prod_l\Big\langle\exp\Big(\frac{1}{2}\frac{g_K}{\dmod}\sum_{nmk} \tilde{W}^{\mG}_{nm} \tilde{W}^{\mG}_{km} W_{ln}^{Q} W_{lk}^{Q}\Big)\Big\rangle_{\mW^Q}\\
    &= d_k \ln\Big\langle\exp\Big(\frac{1}{2}\frac{g_K}{\dmod}\sum_{nmk} \tilde{W}^{\mG}_{nm} \tilde{W}^{\mG}_{km} W_{n}^{Q} W_{k}^{Q}\Big)\Big\rangle_{\mW^Q}.
\end{align}
In the last line, we used the independence in $l$ to pull out a factor $d_k$. Explicitly inserting the prior of $\mW^Q$, we get
\begin{align}
    \mathcal{W}(\tilde{\mW}^{\mG}) &= d_k \ln\exp\bigg(-\frac{1}{2}\sum_{nm} W_{n}^{Q} \bigg[-\frac{g_K}{\dmod}\sum_m\tilde{W}^{\mG}_{nm} \tilde{W}^{\mG}_{nm} + \frac{\dmod}{g_Q} \delta_{nm} \bigg] W_{m}^{Q}\bigg) \times \det \bigg[ \bigg( \frac{\dmod}{g_Q} \mI\bigg)^{1/2}\bigg] \\
    &= d_k \ln \bigg(\det \bigg[\sum_{nm}\frac{\dmod}{g_Q} \delta_{nm} -\frac{g_K}{\dmod}\sum_{nm}\tilde{W}^{\mG}_{nm} \tilde{W}^{\mG}_{nm}\bigg]^{-1/2} \, \det \bigg[\bigg( \frac{\dmod}{g_Q} \mI\bigg)^{1/2}\bigg]\bigg)\\
    &= d_k \ln \bigg(\det \bigg[\sum_{nm}\delta_{nm} -\frac{g_K g_Q}{\dmod^2}\sum_{nm}\tilde{W}^{\mG}_{nm} \tilde{W}^{\mG}_{nm}\bigg]^{-1/2}\bigg)\\
    &= -\frac{d_k}{2} \ln \det \bigg[\mI -\frac{g_K g_Q}{\dmod^2}\tilde{\mW}^{\mG} \tilde{\mW}^{\mG\T}\bigg]
\end{align}
Thus, up to normalization constants and prefactors,
\begin{align}
    I(\mW^{\mG})
    &:=
    -\ln p(\mW^{\mG})
    \\
    &=
    \sup_{\tilde{\mW}^{\mG}}
    \bigg\{
        \tr\big[\tilde{\mW}^{\mG\T}\mW^{\mG}\big]
        -
        \mathcal{W}(\tilde{\mW}^{\mG})
    \bigg\}\\
    &=
    \sup_{\tilde{\mW}^{\mG}}
    \bigg\{
        \tr\big[\tilde{\mW}^{\mG\T}\mW^{\mG}\big]
        +
        \frac{d_k}{2}
        \ln \det \bigg[\mI -\frac{g_K g_Q}{\dmod^2}\tilde{\mW}^{\mG} \tilde{\mW}^{\mG\T}\bigg]
    \bigg\}.
\end{align}

Since the second term depends only on the singular values of $\tilde{\mW}^{\mG}$, the saddle point aligns the singular vectors of $\tilde{\mW}^{\mG}$ with those of $\mW^{\mG}$. Writing
\begin{equation}
    \mW^{\mG}=\mU\diag(\sigma_i)\mV^\T,
    \qquad
    \tilde{\mW}^{\mG}=\mU\diag(\tilde{\sigma}_i)\mV^\T,
\end{equation}
we obtain
\begin{equation}
    I(\mW^{\mG})
    =
    \sup_{\tilde{\sigma}_i}
    \sum_i
    \left[
        \tilde{\sigma}_i\sigma_i
        +
        \frac{d_k}{2}
        \ln\left(
            1
            -
            \frac{g_K g_Q}{\dmod^2}
            \tilde{\sigma}_i^2
        \right)
    \right].
\end{equation}
The saddle point satisfies
\begin{align}
    0
    &=
    \frac{\partial}{\partial\tilde{\sigma}_{j}}
    \left(
        \sum_{i}\tilde{\sigma}_{i}\sigma_{i}
        +
        \frac{d_{k}}{2}
        \sum_{i}
        \ln\left[
            1-\frac{g_{K}g_{Q}}{\dmod^{2}}\tilde{\sigma}_{i}^{2}
        \right]
    \right)\\
    &=
    \sigma_{j}
    -
    d_{k}
    \frac{g_{K}g_{Q}}{\dmod^{2}}
    \frac{\tilde{\sigma}_{j}}
    {1-\frac{g_{K}g_{Q}}{\dmod^{2}}\tilde{\sigma}_{j}^{2}}.
\end{align}
Equivalently,
\begin{equation}
    \sigma_j \tilde{\sigma}_j^2
    +
    d_k\tilde{\sigma}_j
    -
    \sigma_j\frac{\dmod^2}{g_Kg_Q}
    =
    0.
\end{equation}
The maximizing branch is
\begin{equation}
    \tilde{\sigma}_{j}^{\star}
    =
    \frac{
        -d_{k}
        +
        \sqrt{
            d_{k}^{2}
            +
            4\sigma_{j}^{2}\frac{\dmod^{2}}{g_{K}g_{Q}}
        }
    }
    {2\sigma_{j}}
    =
    \sqrt{
        \frac{d_{k}^{2}}{4\sigma_{j}^{2}}
        +
        \frac{\dmod^{2}}{g_{K}g_{Q}}
    }
    -
    \frac{d_{k}}{2\sigma_{j}}.
\end{equation}
Substituting this saddle point back into the Legendre transform gives
\begin{equation}
    I(\mW^{\mG})
    =
    \frac{d_k}{2}
    \sum_i
    \left[
        \sqrt{
            1
            +
            \frac{4\dmod^2\sigma_i^2}{d_k^2 g_K g_Q}
        }
        -1
        +
        \ln\left(
            \frac{2}
            {1+\sqrt{
                1
                +
                \frac{4\dmod^2\sigma_i^2}{d_k^2 g_K g_Q}
            }}
        \right)
    \right].
\label{eq:product_exact_rate}
\end{equation}

At small singular values, the exact rate function expands as
\begin{equation}
    I(\mW^{\mG})
    =
    \frac{\dmod^2}{2d_k g_K g_Q}
    \sum_i \sigma_i^2
    +
    O\left(\sum_i \sigma_i^4\right)
    \approx
    \frac{\dmod^2}{2d_k g_K g_Q}
    \|\mW^{\mG}\|_F^2
    \label{eq:approx_prod_gaus}
\end{equation}
Thus, near the origin, the effective regularization induced by the Gaussian product prior is standard $\ell_2$ weight decay / Gaussian prior on the combined matrix $\mW^{\mG}$.

At large singular values, the leading term is linear in the singular values,
\begin{equation}
    I(\mW^{\mG})
    =
    \frac{\dmod}{\sqrt{g_K g_Q}}
    \sum_i \vert \sigma_i \vert
    +
    O(\ln \sigma_i)
    \approx
    \frac{\dmod}{\sqrt{g_K g_Q}}
    \|\mW^{\mG}\|_*.
\end{equation}
Therefore, for general $d_k$, the exact product prior interpolates between Frobenius-norm regularization at small singular values and nuclear-norm regularization in the tails. 

The Gaussian approximation is valid when the typical singular values of $\mW^{\mG}$ 
are small compared to the crossover scale $\sigma_* = d_k\sqrt{g_K g_Q}/(2\dmod)$ 
at which correction to Eq.~\eqref{eq:approx_prod_gaus} become large. Since the entries of $\mW^{\mG}$ have 
variance $d_k g_K g_Q/\dmod^2$ and $\mW^{\mG}$ is a $\dmod\times\dmod$ matrix, 
its typical singular values scale as 
$\sigma_{\mathrm{typ}}\sim\sqrt{d_k g_K g_Q/\dmod}$. 
The ratio to the crossover scale is
\begin{equation}
    \frac{\sigma_{\mathrm{typ}}}{\sigma_*}
    \sim 2\sqrt{\frac{\dmod}{d_k}},
\end{equation}
which is independent of $g_K g_Q$ -- rescaling the variances shifts both scales 
identically, so they cancel. The Gaussian approximation is therefore 
self-consistently valid only when $d_k\gg\dmod$, and breaks down when $d_k \approx \dmod$, where typical singular values sit right at the 
crossover between the $\ell^2$ and nuclear-norm regimes.

This nuclear-norm behavior aligns with recent work showing that factorizing the key and query matrices changes the implicit regularization of attention. For one-layer softmax attention, \cite{sheenImplicitRegularizationGradient2024} showed that gradient flow with separately trained key and query matrices implicitly minimizes the nuclear norm of their product. Similarly, \cite{tarzanaghTransformersSupportVector2024} showed that the factorized $(\mW^K,\mW^Q)$ parametrization induces a nuclear-norm SVM over the combined matrix $\mW=\mW^Q \mW^{K \top}$, rather than the Frobenius-norm objective obtained by optimizing $\mW$ directly. Relatedly, \cite{Zhang25TrainingDynamicsLinearAttention} analyzed the in-context regression setting and compared the saddle-point structure of the fused and factorized dynamics.

Nevertheless, many theoretical works replace the key-query product by a single fused matrix~\cite{biettiBirthTransformerMemory2023,edelmanEvolutionStatisticalInduction2024,nichaniHowTransformersLearn2024,musatEmergenceInductionHeads2026}. Our results above clarify that this replacement is benign when $d_k\gg \dmod$ and pinpoint when the effective regularization crosses over from $\ell_2$ weight decay to nuclear norm. Remarkably, at the level of attention patterns instead of on the level of the weights this approximation hold when $\dmod,d_k\gg L$ as we show in Appendix~\ref{app:attention_prior_KQ}.

\section{Extension to multiple attention layers}
\label{app:multi-layer}

\global\long\def\N{\mathcal{N}}%
\global\long\def\dmodel{d}%
\global\long\def\dmodelm{d^{-1}}%
\global\long\def\dmodelmm{d^{-2}}%
\global\long\def\th{\tilde{h}}%
\global\long\def\tG{\tilde{G}}%
\global\long\def\tC{\tilde{C}}%
\global\long\def\tf{\tilde{f}}%
\global\long\def\tr{\mathrm{tr}}%
\global\long\def\T{\mathrm{T}}%
\global\long\def\I{\mathrm{I}}%

To extend the theory to linear transformers with $M$ layers, each
with a single attention head, we generalize the single layer case
as

\begin{align}
h_{i}^{(\ell+1)a}(x) & =G^{(\ell)ab}\,B_{ij}^{(\ell)}h_{j}^{(\ell)b}\,,\quad1\le\ell\le M-1\,,\label{eq:simplified_linear_transformer_multi}\\
G^{(\ell)ab} & =\frac{1}{L\sqrt{d}}\,h_{i}^{(\ell)a}\,W_{ij}^{(\ell)}\,h_{j}^{(\ell)b}\,,\label{eq:attention_multi}
\end{align}
together with the convention $h^{(0)}=x$ and $h^{(M+1)}=f$. We here
again combine the key and the query matrix into a single matrix $W$
with prior $W_{ij}^{(\ell)}\stackrel{\text{i.i.d.}}{\sim}\N\big(0,g_{W}\,d^{-1}\big)$
and denote the output matrix $W^{O}$ of each layer as $B_{ij}^{(\ell)}\stackrel{\text{i.i.d.}}{\sim}\N(0,g_{B}\,\dmodelm$)
here. We also use the shorthand $\dmodel=d_{\mathrm{model}}$ in the
following.

As the priors on $W^{(\ell)}$ and $B^{(\ell)}$ are independent across
different layers $\ell$, we may consider one layer at a time. Performing
the analogous steps as in App. \ref{app:network_prior}, we get by enforcing the definition
of the attention layer (\ref{eq:attention_multi})
\begin{align*}
 & \int d\tG^{(\ell)}\,\Big\langle\exp\big(-\tr\tG^{(\ell)\T}G^{(\ell)}+\tG_{\alpha}^{(\ell)ab}\frac{1}{L\sqrt{\dmodel}}\,h_{\alpha i}^{(\ell)a}\,W_{ij}^{(\ell)}\,h_{\alpha j}^{(\ell)b}\big)\Big\rangle_{W_{ij}^{(\ell)}\stackrel{\text{i.i.d.}}{\sim}\N(0,g_{W}\,1/\dmodel)}\\
= & \int d\tG^{(\ell)}\,\exp\big(-\tr\tG^{(\ell)\T}G^{(\ell)}+\tG_{\alpha}^{(\ell)ab}\tG_{\beta}^{(\ell)a^{\prime}b^{\prime}}\,\frac{g_{W}}{2L^{2}}\,C_{\alpha\beta}^{(\ell)aa^{\prime}}\,C_{\alpha\beta}^{(\ell)bb^{\prime}}\big)\,,
\end{align*}
where we defined
\begin{align*}
C_{\alpha\beta}^{(\ell)ab} & =\dmodelm\,h_{\alpha i}^{(\ell)a}h_{\beta i}^{(\ell)b}\,.
\end{align*}
Enforcing the equation (\ref{eq:simplified_linear_transformer_multi})
which projects the activity from one layer to the next, we likewise
have

\begin{align*}
 & \int d\th^{(\ell+1)}\Big\langle\exp\big(-\tr\th^{(\ell+1)\T}h^{(\ell+1)}+\th_{\alpha i}^{(\ell+1)a}G_{\alpha}^{(\ell)ab}\,B_{ij}^{(\ell)}h_{\alpha j}^{(\ell)b}\big)\Big\rangle_{B_{ij}^{(\ell)}\stackrel{\text{i.i.d.}}{\sim}\N(0,g_{B}\,\dmodelm)}\\
= & \int d\th^{(\ell+1)}\exp\big(-\tr\th^{(\ell+1)\T}h^{(\ell+1)}+\frac{1}{2}g_{B}\,\th_{\alpha i}^{(\ell+1)a}\th_{\beta i}^{(\ell+1)a^{\prime}}G_{\alpha}^{(\ell)ab}G_{\beta}^{(\ell)a^{\prime}b^{\prime}}\,C_{\alpha\beta}^{(\ell)bb^{\prime}}\big)\,,
\end{align*}
where again the attention-weighted input kernel appears in each layer
$\ell$
\begin{align*}
\mathcal{K}_{\alpha\beta}^{(\ell)aa^{\prime}} & :=G_{\alpha}^{(\ell)ab}G_{\beta}^{(\ell)a^{\prime}b^{\prime}}\,C_{\alpha\beta}^{(\ell)bb^{\prime}}\,.
\end{align*}
The appearance of $\sum_{i}\th_{\alpha i}^{(\ell+1)a}\th_{\beta i}^{(\ell+1)a^{\prime}}$
shows that, given $G^{(\ell)}$ and $C^{(\ell)}$, the distribution
of the preactivations $h_{\alpha i}$ factorizes over $i$, so that
the kernel $C$ is expected to concentrate also at intermediate layers.
Its definition is enforced by
\begin{align*}
 & C_{\alpha\beta}^{(0)ab}=\dmodelm x_{\alpha i}^{a}x_{\beta i}^{b}+g_{p}\,,\\
\\ & p(C^{(\ell)}|C^{(\ell-1)},G^{(\ell-1)})\\
= & \int d\tC^{(\ell)}\Big\langle\exp\big(-\tr\tC^{(\ell)\T}C^{(\ell)}+\tC_{\alpha\beta}^{(\ell)ab}\,\dmodelm\,h_{\alpha i}^{(\ell)a}h_{\beta i}^{(\ell)b}\big)\Big\rangle_{h_{\alpha i}^{(\ell)a}\stackrel{\text{i.i.d. in }i}{\sim}\N\big(0,g_{B}\mathcal{K}_{\alpha\beta}^{(\ell-1)ab}\big)}\,\\
= & \int d\tC^{(\ell)}\exp\Big(-\tr\tC^{(\ell)\T}C^{(\ell)}+\dmodel\,\ln\Big\langle\exp\big(\tC_{\alpha\beta}^{(\ell)ab}\,\dmodelm\,h_{\alpha}^{(\ell)a}h_{\beta}^{(\ell)b}\big)\Big\rangle_{h_{\alpha}^{(\ell)a}\sim\N\big(0,g_{B}\mathcal{K}_{\alpha\beta}^{(\ell-1)ab}\big)}\,\Big)\,,
\end{align*}
where we exploited the independence over $i$ in the last step. The
appearance of the scaling form $\dmodel\,\ln\langle\exp\big(\dmodelm\,\tC\ldots\big)\rangle$
of the cumulant-generating function shows that $C$ concentrates. In
the limit of large $\dmodel$, we may thus approximate its distribution
with the rate function

\begin{align}
-\ln p(C^{(\ell)}|C^{(\ell-1)})/\dmodel & \simeq\sup_{\tC^{(\ell)}}\tr\tC^{(\ell)\T}C^{(\ell)}-\dmodel\,W_{C}\big(\tC^{(\ell)}|\mathcal{K}^{(\ell-1)}\big)\,,\label{eq:p_C}\\
W_{C}(\tC) & :=\ln\Big\langle\exp\big(\tC_{\alpha\beta}^{(\ell)ab}\,\dmodelm\,h_{\alpha}^{(\ell)a}h_{\beta}^{(\ell)b}\big)\Big\rangle_{h_{\alpha}^{(\ell)a}\sim\N\big(0,g_{B}\mathcal{K}_{\alpha\beta}^{(\ell-1)ab}\big)}\,.\label{eq:def_W_C}
\end{align}
Evaluating the supremum condition yields
\begin{align*}
0 & =C_{\alpha\beta}^{(\ell)ab}-\langle h_{\alpha}^{(\ell)a}h_{\beta}^{(\ell)b}\rangle\,,
\end{align*}
which shows that the second order statistics of $h$ is fixed by $C^{(\ell)}$,
the argument of the left hand side of (\ref{eq:p_C}). This implies
that the quadratic form $A$ in the exponent $-\frac{1}{2}h^{\T}Ah$
appearing in (\ref{eq:def_W_C}) satisfies
\begin{align}
\big[C^{(\ell)}\big]^{-1}=A= & -2\,\dmodelm\,\tC_{\alpha\beta}^{(\ell)ab}+\big[g_{B} \mathcal{K}^{(\ell-1)}\big]^{-1}\,,\label{eq:C_l_as_tilde}
\end{align}
so that
\begin{align*}
\tC_{\alpha\beta}^{(\ell)ab} & =\frac{\dmodel}{2}\,\big(\big[g_{B} \mathcal{K}^{(\ell-1)}\big]^{-1}-\big[C^{(\ell)}\big]^{-1}\big)\,.
\end{align*}
Likewise it implies that the Gaussian integral over $h$ in $W_{C}$
(\ref{eq:def_W_C}) yields $\frac{1}{2}\ln\det\big(C^{(\ell)}\big)$
and has the normalization constant $-\frac{1}{2}\ln\det\big(g_{B} \mathcal{K}^{(\ell-1)}\big)$,
so that together we get
\begin{align}
-\ln p(C^{(\ell)}|C^{(\ell-1)})/\dmodel & \simeq\frac{1}{2}\tr\,\big[g_{B} \mathcal{K}^{(\ell-1)}\big]^{-1}C^{(\ell)}-\frac{1}{2}\ln\det\big(C^{(\ell)}\big)+\frac{1}{2}\ln\det\big(g_{B} \mathcal{K}^{(\ell-1)}\big)+\mathrm{const.}\,,\label{eq:KL}\\
 & \equiv\mathrm{KL}\big(\N(0,C^{(\ell)})\|\N(0,g_{B} \mathcal{K}^{(\ell-1)})\big)\,,
\end{align}
where we dropped the constant term $-\frac{1}{2}\tr\big[C^{(\ell)^{-1}}C^{(\ell)}\big]=-\frac{PL}{2}$.
The expression (\ref{eq:KL}) is that of a KL divergence between two
Gaussian distributions with covariance matrices $g_{B} \mathcal{K}^{(\ell-1)}$
and $C^{(\ell)}$. Since the KL divergence is minimal for $C^{(\ell)}=g_{B} \mathcal{K}^{(\ell-1)}$,
this shows that this term tries to keep these two kernels close; the
point where the distance vanishes is the neural network Gaussian process
solution. So together with the likelihood of the labels as $\N(y|f,\sigma^{2})$,
we have

\begin{align}
-\ln\,p(y|C^{(0)}) & +\mathrm{const.}\simeq\inf_{G}\,\inf_{C}\label{eq:min_max_action-1}\\
 & \sum_{\ell=1}^{M-1}\frac{1}{2}C^{(\ell)}\,\big[g_{B} \mathcal{K}^{(\ell-1)}\big]^{-1}-\frac{1}{2}\ln\det\big(C^{(\ell)}\big)+\frac{1}{2}\ln\det\big(g_{B} \mathcal{K}^{(\ell-1)}\big)\nonumber \\
+ & \sup_{\tG}\,\sum_{\ell=1}^{M-1}\tr\tG^{(\ell)\T}G^{(\ell)}-\tG_{\alpha}^{(\ell)ab}\tG_{\beta}^{(\ell)a^{\prime}b^{\prime}}\,\frac{g_{W}}{2L^{2}}\,C_{\alpha\beta}^{(\ell)aa^{\prime}}\,C_{\alpha\beta}^{(\ell)bb^{\prime}}\,\\
+ & \sup_{\tf}\Big\{\tf^{\T}y-\frac{\sigma^{2}}{2}\sum_{i}\tf_{\alpha i}^{a}\tf_{\alpha i}^{a}+\frac{g_{B}}{2}\,\sum_{i}\tf_{\alpha i}^{a}\tf_{\beta i}^{b}\, \mathcal{K}_{\alpha\beta}^{(M)\,ab}\Big\}\,
\end{align}
A more compact form is achieved by performing the Gaussian integrals
over $\tG $ and $\tf$ instead of taking their supremum; this yields
a slightly better approximation, since a normalization determinant
is contained and yields
\begin{align}
-\ln\,p(y|C^{(0)}) & =\inf_{G}\,\inf_{C}\label{eq:log_likelihood_multi_final}\\
 & \sum_{\ell=1}^{M-1}\mathrm{KL}\big(\N(0,C^{(\ell)})\|\N(0,g_{B} \mathcal{K}^{(\ell-1)})\big)\nonumber \\
- & \sum_{\ell=0}^{M-1}\,\ln\N(G^{(\ell)}|0,g_{W}L^{-2}\,C^{(\ell)}C^{(\ell)})\nonumber \\
- &\ln\N(y|0, g_B\mathcal{K}^{(M)}+\sigma^{2}\I)\nonumber \\
\nonumber \\\big[C^{(\ell)}C^{(\ell)}\big]_{(\alpha ab),(\beta a^{\prime}b^{\prime})} & :=C_{\alpha\beta}^{(\ell)aa^{\prime}}\,C_{\alpha\beta}^{(\ell)bb^{\prime}}\,,\nonumber \\
\mathcal{K}_{\alpha\beta}^{(\ell)aa^{\prime}} & :=G_{\alpha}^{(\ell)ab}G_{\beta}^{(\ell)a^{\prime}b^{\prime}}\,C_{\alpha\beta}^{(\ell)bb^{\prime}}\,.\nonumber 
\end{align}
This expression shows the different tendencies of the network:
\begin{itemize}
\item The first line aims to minimize the KL-divergence between the attention-weighted
Gaussian kernel matrix $g_{B} \mathcal{K}^{(\ell-1)}$ of the previous layer
and the Gaussian kernel matrix of the next layer $C^{(\ell)}$. The
KL-divergence measures the information loss, when approximating the
Gaussian with $C^{(\ell)}$ by the Gaussian with $g_{B}K^{(\ell-1)}$.
\item The second line is minimal, if the attention matrix $G$ is most likely
under the Gaussian with covariance $g_{W}L^{-2}C^{(\ell)}C^{(\ell)}$;
the structure of the indices shows that the attention patterns $G_{\alpha}^{ab}$
and $G_{\beta}^{a^{\prime}b^{\prime}}$ tend to be correlated if the
sequences $\alpha$ and $\beta$ of the previous layer are similar
at positions $a$ and $a^{\prime}$ and at positions $b$ and $b^{\prime}$.
The sign of the resulting correlation depends on the product of the
signs of the correlations in each factor $C^{(\ell)}$. This term
shows the tendency of the attention patterns to focus attention on
positions with similar entries.
\item The third line is minimal, if the target $y$ is most likely under
the Gaussian with covariance $\mathcal{K}^{(M)}+\sigma^{2}\I$.
\end{itemize}
The state after learning is a compromise between these three forces.

\section{Stochastic Gradient Langevin Dynamics\label{app:langevin}}

To numerically validate theoretical predictions against real networks,
we draw samples from the posterior over networks conditioned on training
data $X=(x_{\alpha})_{\alpha=1,\ldots,P},Y=(y_{\alpha})_{\alpha=1,\ldots,P}$. We assume Gaussian priors on the network weights at initialization as $W_{ij}^{\mathrm{emb}}\stackrel{\text{i.i.d.}}{\sim}\mathcal{N}(0,g_{\mathrm{emb}})$, $p_{i}^{a} \stackrel{\text{i.i.d.} }{\sim}\mathcal{N}(0,g_p/\dmod)$, $W_{ij}^{Q}\stackrel{\text{i.i.d.}}{\sim}\mathcal{N}(0,g_{Q}/\dmod)$, $W_{ij}^{K}\stackrel{\text{i.i.d.}}{\sim}\mathcal{N}(0,g_{K}/\dmod)$, $W_{ij}^{O}\stackrel{\text{i.i.d.}}{\sim}\mathcal{N}(0,g_O/\dmod)$ with $\dmod$ being the model dimension. We set $g_{\mathrm{emb}} = g_P = 1$. Further, we assume Gaussian readout noise $\vy_{\alpha}=\vf_{\alpha}+\bm{\xi}$ with $\bm{\xi}\sim\N(0,\sigma^2\mI)$.

Sampling from the Bayesian posteriors then corresponds to training the network via stochastic gradient Langevin dynamics (SGLD)~\cite{wellingBayesianLearningStochastic2011}. Following~\cite{Naveh21_064301}, the parameters $\Theta$ are evolved according to the stochastic differential equation
\begin{align}
\partial_{t}\Theta(t) & =-\gamma\Theta(t)-\nabla_{\Theta}\mathcal{L}(\Theta(t);Y)+\sqrt{2T}\zeta(t),\label{eq:Appendix_SingleParameter_SGLD}\\
\big\langle\zeta_{i}(t)\zeta_{j}(s)\big\rangle & =\delta_{ij}\delta(t-s),\nonumber 
\end{align}
where the squared error loss is $\mathcal{L}(\Theta;Y)=\sum_{\alpha=1}^{P}(f_{\alpha}(\Theta)-y_{\alpha})^{2}$
and $f_{\alpha}(\Theta)$ denotes the network output for input $\alpha$.
At large times $t$, this dynamics converges to the equilibrium distribution over $\Theta$,
given by
\begin{equation}
\lim_{t\rightarrow\infty}p\left(\Theta(t)\right)\sim\exp\left(-\frac{\gamma}{2T}\|\Theta\|^{2}-\frac{1}{T}\mathcal{L}(\Theta;Y)\right).
\end{equation}
This stationary distribution follows from the Fokker-Planck equation~\cite{Risken1996} governing the evolution of the density of $\Theta$. The corresponding
marginal density over network outputs takes the form
\begin{align}
p(Y|X)\propto & \int d\Theta\,\exp\big(-\frac{\gamma}{2T}\,\|\Theta\|^{2}-\frac{1}{T}\,\|f-Y\|^{2}\big)\\
\propto & \Big\langle\exp\big(-\frac{1}{T}\,\|f-Y\|^{2}\big)\Big\rangle_{\Theta_{k}\stackrel{\text{i.i.d.}}{\sim}\N(0,T/\gamma)}\nonumber \\
\propto & \N(Y|f,T/2)\,\langle\delta\big[f-f(\Theta)\big]\rangle_{\Theta_{k}\stackrel{\text{i.i.d.}}{\sim}\N(0,T/\gamma)},\nonumber 
\end{align}
which, upon identifying $p(f|X)\equiv\langle\delta\big[f-f(\Theta)\big]\rangle_{\Theta_{k}\stackrel{\text{i.i.d.}}{\sim}\N(0,T/\gamma)}$,
coincides with the network prior
under the identifications $\sigma^2=T/2$
for the regularization noise and $T/\ensuremath{\gamma}=g/d$ for
the variance of parameter $\Theta_{k}$. In practice, realizing distinct
variances per parameter requires assigning a separate weight decay $\gamma$
to each parameter.
The time-discretized form of \eqref{eq:Appendix_SingleParameter_SGLD}
reads
\begin{align}
\Theta_{t} & =\Theta_{t-1}-\eta\left(\gamma\Theta_{t-1}+\nabla_{\Theta}\mathcal{L}(\Theta_{t-1};Y)\right)+\sqrt{2T\eta}\,\zeta_{t},\\
\langle\zeta_{t}\zeta_{s}\rangle & =\delta_{ts},\nonumber 
\end{align}
with $\zeta_{t}$ drawn from a standard normal distribution and $\eta$ a finite step size serving
as the learning rate. Faithful discretization of the continuous-time dynamics in \eqref{eq:Appendix_SingleParameter_SGLD}
requires $\eta$ to remain small. In this form, SGLD is equivalent to full-batch gradient descent
augmented with i.i.d. standard normal noise and weight decay regularization \cite{Krogh91}.
The parameter $\sigma^2$, which enters the main text as the regularization term on the diagonal of the output kernel, controls the relative
weight of the prior against the likelihood. Large $\sigma^2$ implies
large $T=2\sigma^2$, amplifying the stochastic noise in SGLD and
thereby upweighting the parameter prior. In contrast, small $\sigma^2$ suppresses the noise and directs the optimization towards minimizing
the loss of training.

The rescaling by the factor $\chi$ used in the main text in Section \ref{sec:posterior}
for large $\chi$
corresponds to training with gradient flow. To see this, note that the
prior variance of the parameters $\theta$ is given by $g=T/\gamma$.
Rescaling the regularization noise $\sigma^2=T/2$ as $T/\chi$ and simultaneously
rescaling $g/\chi$ therefore downscales the noise in the gradient update
\eqref{eq:Appendix_SingleParameter_SGLD} but maintains the same weight decay $\gamma$.
In the limit of large $\chi$ the update equation \eqref{eq:Appendix_SingleParameter_SGLD}
thus tends to gradient flow without noise but with regularization by weight decay.

A sublety arises when taking the limit of Langevin noise to zero, which concerns the
stationary distribution, because the limit $\lim_{\chi \to \infty}$ and the limit
$\lim_{t \to \infty}$ do not necessarily commute: The stationary state reached with
gradient flow corresponds to $\lim_{t\to\infty} \lim_{\chi \to \infty}$, while considering
the (stationary) Bayesian posterior distribution in the limit $\chi\to\infty$,  corresponds to
$\lim_{\chi\to\infty} \lim_{t\to\infty}$. The two limits may not commute in cases in which
ergodicity is broken in gradient flow; that is, if the training dynamics gets stuck within a
local minimum. In the Bayesian setting this cannot happen as long as $T>0$, or correspondingly, 
$\chi < \infty$. The reason why training with Adam and our theory obtained from the limit
$\lim_{\chi\to\infty \lim_{t\to\infty}}$ still agree relies on our training procedure, which
filters training runs that got stuck at sub-optimal minima, as described in App. \ref{appendix:traj_filtering}.
Thus we avoid this form of ergodicity breaking in the noiseless training procedure, which is
equivalent to having an infinitesimal noise in the Bayesian posterior.

\section{Field-Theoretic Toy Model of the Avoided Second Phase Transition}
\label{app:field_theory_toy}

It is useful to formulate a minimal Landau theory that captures the same qualitative phenomenon as in the main text: one genuine symmetry-breaking transition, followed by a second lower scale at which the subleading structure changes rapidly without an additional thermodynamic singularity.

Consider two real scalar fields $\phi_1,\phi_2$ with Euclidean action
\begin{equation}
    S_{\mathrm{toy}}[\phi_1,\phi_2]
    =
        \frac{r_1(T)}{2}\phi_1^2
        + \frac{r_2(T)}{2}\phi_2^2
        - J\,\phi_1\phi_2
        + \frac{u_1}{4}\phi_1^4
        + \frac{u_2}{4}\phi_2^4,
    \label{eq:toy_action}
\end{equation}
with $u_1,u_2>0$ for stability. The bilinear coupling $J$ breaks the independent sign-flip symmetry
$\mathbb{Z}_2\times \mathbb{Z}_2$ down to the diagonal $\mathbb{Z}_2$ symmetry
$(\phi_1,\phi_2)\mapsto(-\phi_1,-\phi_2)$. This symmetry already suggests that for $J\neq 0$ there is generically only \emph{one} true symmetry-breaking transition.

To make contact with a control parameter such as temperature, we take
\begin{equation}
    r_i(T)=a_i(T-T_i^0),\qquad a_i>0.
    \label{eq:toy_ri}
\end{equation}
If the fields were decoupled, each would become critical at its own bare scale $T_i^0$. For the analogy to the main text we take $T_1^0>T_2^0$, so that $\phi_1$ is the mode that would order first upon cooling, while $\phi_2$ becomes soft only at a lower scale.

\paragraph{One true phase transition.}
Around the symmetric saddle $\phi_1=\phi_2=0$, the quadratic form is
\begin{equation}
    \mathcal{M}(T)
    =
    \begin{pmatrix}
        r_1(T) & -J \\
        -J & r_2(T)
    \end{pmatrix},
\end{equation}
with eigenvalues
\begin{equation}
    m_{\pm}^2(T)
    =
    \frac{r_1(T)+r_2(T)}{2}
    \pm
    \frac{1}{2}\sqrt{(r_1(T)-r_2(T))^2+4J^2}.
    \label{eq:toy_eigs}
\end{equation}
The disordered phase loses stability when the lower eigenvalue $m_-^2$ vanishes, equivalently when
\begin{equation}
    r_1(T_c)\,r_2(T_c)=J^2.
    \label{eq:toy_Tc}
\end{equation}
Thus the transition is controlled by the mixed normal mode rather than by either field separately. For weak mixing and well-separated bare scales, one finds
\begin{equation}
    T_c
    \simeq
    T_1^0
    +
    \frac{J^2}{a_1\,r_2(T_1^0)},
\end{equation}
so the true transition occurs close to the higher bare scale $T_1^0$.

The mean-field equations in the ordered phase are
\begin{align}
    r_1(T)\,\bar{\phi}_1 - J\,\bar{\phi}_2 + u_1\bar{\phi}_1^3 &= 0,
    \label{eq:toy_mf1}\\
    r_2(T)\,\bar{\phi}_2 - J\,\bar{\phi}_1 + u_2\bar{\phi}_2^3 &= 0.
    \label{eq:toy_mf2}
\end{align}
Because $J\neq 0$, once one component condenses the other is immediately induced. Near the transition, where $\bar{\phi}_2$ is still small, Eq.~\eqref{eq:toy_mf2} gives
\begin{equation}
    \bar{\phi}_2
    \simeq
    \frac{J\,\bar{\phi}_1}{r_2(T)}.
    \label{eq:toy_phi2_induced}
\end{equation}
So the lower-scale field is not waiting for an independent symmetry-breaking transition; it is sourced explicitly by the condensate of the first field.

\paragraph{A second lower scale without a second singularity.}
Even though there is only one exact transition, there is generally still a second physically important scale. As the control parameter is varied deeper into the ordered phase, the effective mass $r_2(T)$ in~\eqref{eq:toy_phi2_induced} can become small near the lower bare scale $T_2^0$. When that happens, $\bar{\phi}_2$ grows rapidly and the composition of the ordered state reorganizes sharply.

The fluctuations around the ordered state are controlled by the Hessian
\begin{equation}
    H(\bar{\phi}_1,\bar{\phi}_2)
    =
    \begin{pmatrix}
        r_1 + 3u_1\bar{\phi}_1^2 &
        -J  \\
        -J &
        r_2 + 3u_2\bar{\phi}_2^2
    \end{pmatrix}.
    \label{eq:toy_hessian}
\end{equation}
At the true critical point, the smaller eigenvalue of $H$ vanishes. The other eigenvalue, however, stays strictly positive for generic $J\neq 0$. This means that the second mode can become \emph{soft} but not truly critical. Correspondingly, the associated susceptibility and correlation length,
\begin{equation}
    \chi_{\mathrm{soft}}\sim \lambda_{\mathrm{soft}}^{-1},
    \qquad
    \xi_{\mathrm{soft}}\sim \lambda_{\mathrm{soft}}^{-1/2},
\end{equation}
can become large but remain finite when the second eigenvalue $\lambda_{\mathrm{soft}}$ develops a pronounced minimum.

This is the standard avoided-critical-point scenario: for $J=0$, the lower-scale field could undergo its own independent transition, but any nonzero $J$ rounds this into a crossover. Thermodynamically there is only one nonanalyticity, yet dynamically and phenomenologically the second scale can look highly transition-like, with large fluctuations, long relaxation times, and a rapid change of the order-parameter composition.

\paragraph{Relation to the main text.}
This toy model provides a useful analogy for the behavior of the two attention order parameters $(\hat{c}_1,\hat{c}_G)$ in the main text. The first instability corresponds to a genuine phase transition into a state with nontrivial attention structure. The later sharp growth of the copy component is analogous to the lower-scale field $\phi_2$: it is strongly enhanced when its effective mass becomes small, but because it is already linearly coupled to the ordered phase, this enhancement is a crossover rather than a second symmetry-breaking transition. In this sense, the later reorganization of the attention pattern is naturally interpreted as an \emph{avoided} second transition.

Finally, we note that if the bilinear coupling is set to zero, $J=0$, the symmetry is enlarged back to $\mathbb{Z}_2\times \mathbb{Z}_2$, and two distinct transitions are generically possible when the bare scales $T_1^0$ and $T_2^0$ are well separated. The absence of such an additional singularity for $J\neq 0$ is precisely the point of the present toy model.

\section{Attention Pattern Coverage by Order Parameters}
\label{app:attn_coverage}
\begin{figure}[h!]
    \centering
    \includegraphics[width=0.49\linewidth]{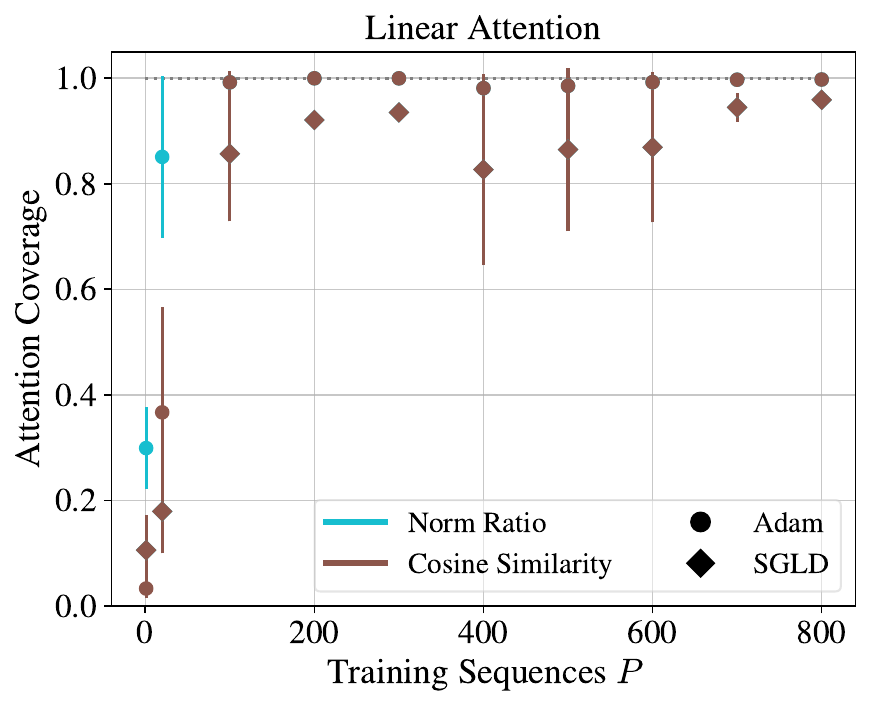}
    \includegraphics[width=0.49\linewidth]{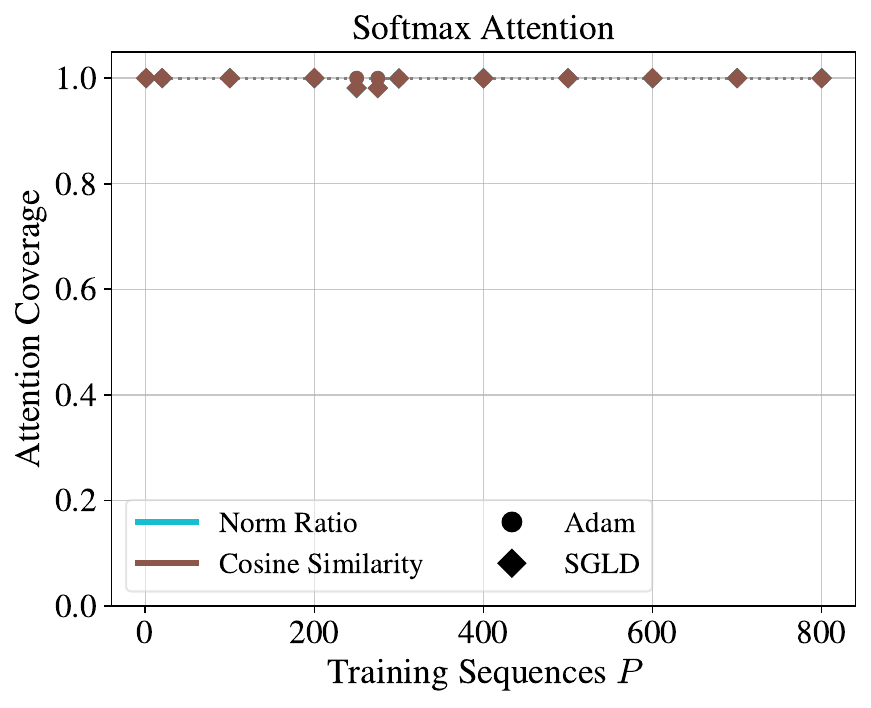}
    \caption{\textbf{Attention-pattern coverage by order-parameter projections.}
    We quantify how much of the empirical attention pattern lies in the two-dimensional subspace spanned by the pooling and copy order-parameter directions, using the projected norm ratio and cosine similarity.
    \textbf{Left:} \emph{Linear attention} is nearly fully described by this subspace for Adam across training set sizes \(P\); throughout much of the panel, Adam markers are hidden behind SGLD markers, and the norm-ratio points are hidden behind the cosine-similarity points.
    SGLD shows the only visible loss of coverage, with larger fluctuations in the crossover region \(300 \leq P \leq 600\), consistent with transient attention components outside the order-parameter plane.
    \textbf{Right:} \emph{Softmax attention} shows essentially perfect coverage for both Adam and SGLD over the full range of \(P\), with all metrics and optimizers nearly overlapping at one.}
    \label{fig:attention_coverage}
\end{figure}

Fig.\ref{fig:attention_coverage} shows the fraction of the full attention pattern that is captured by the patterns in the $(\hat{c}_1, \hat{c}_G)$ and $(\hat{a}_1, \hat{a}_G)$ subspace for linear and softmax attention. We quantified the coverage by cosine similarity and norm ratio between the full empirical attention pattern and the attention pattern projected down to the two-dimensional subspace.

\emph{For linear attention} (Fig.~\ref{fig:attention_coverage} left panel), the two-dimensional order parameter space faithfully captures the degrees of freedom used by the network during training. The agreement for Adam means that the competition between pooling and copy attention accounts for essentially all of the learned attention structure, with no significant orthogonal components. 
SGLD samples show more variance than Adam in the crossover region $300 \leq P \leq 600$, where the coverage plot reveals leakage to attention patterns outside the two-dimensional subspace but remains above $80 \%$ for all $P$ values after the attention pattern departs from zero ($P \gtrsim 20$).
We attribute this effect to the large fluctuations expected near the point where the pooling minimum destabilizes and the copy minimum emerges (Fig.~\ref{fig:mesh1}): as the saddle structure dissolves and the copy minimum deepens, the posterior broadens and the network transiently explores directions outside out 2d space. Away from this region, coverage recovers, confirming that the two-parameter reduction is accurate throughout the ordered phase.

\emph{For softmax attention} (Fig.~\ref{fig:attention_coverage} right panel), we see perfect agreement for both Adam and SGLD for the full range of $P$.
\clearpage

\section{Details of Numerical Experiments}

\paragraph{Experimental details.}
Unless stated otherwise, all numerical experiments use the single-layer,
single-head linear attention model introduced in Section~\ref{sec:model} with fused attention and readout matrices. Training data consist of
length-$L$ sequences of one-hot encoded tokens drawn uniformly from a vocabulary of size $V$, with shift-by-one copy labels. Across experiments,
we vary only the number of training sequences
\[
P \in \{1,10,20,60,100, 250, 275,200,300,400,500,600,700,800\},
\]
and linear/softmax attention while keeping all other settings fixed.

For each value of $P$, we train an ensemble of $10$ models. Optimization is
performed in two stages. We first run an Adam warmup phase for $2{,}000$
epochs with learning rate $10^{-5}$. For softmax we use learning rate warmup from $10^{-7}$ to $10^{-5}$ over $10\%$ of the Adam epochs. Early stopping is used
with patience $100$ and minimum improvement threshold $0$. We then continue training with
Langevin updates using learning rate $5\times 10^{-10}$, noise level
$\sigma^2 = 30$, and batch size
$1{,}000$, for at most $15{,}000$ epochs in total. 
Unless noted
otherwise, reported results are averaged over the ensemble at fixed $P$.

Each run of the full ensemble required roughly 5 hours on an H100, such that in total the experiments required $100$ H100 hours.

\begin{table}[h]
\centering
\small
\setlength{\tabcolsep}{10pt}
\renewcommand{\arraystretch}{1.15}
\begin{tabular}{@{}ll@{}}
\toprule
Hyperparameter & Value \\
\midrule
Sequence length $L$ & $25$ \\
Vocabulary size $V$ & $5$ \\
Training set size $P$ & $\{1,10,20,60,100,200, 250, 275, 300,400,500,600,700,800\}$ \\
Test set size & $100$ \\
Ensemble size & $10$ \\
Architecture & single-layer linear/softmax attention \\
Number of heads & $1$ \\
Model dimension $d_{\mathrm{model}}$ & $10{,}000$ \\
$\chi$ & $30$ \\
Warmup optimizer & Adam \\
Warmup epochs & $2{,}000$ \\
Warmup learning rate & $10^{-5}$ \\
Second-stage optimizer & Langevin / SGLD \\
Second-stage learning rate & $5\times 10^{-10}$ \\
Batch size & $1{,}000$ \\
Maximum number of epochs & $15{,}000$ \\
Noise level $\sigma^2$ & $30$ \\
Early stopping patience & $100$ \\
Early stopping minimum improvement & $0$ \\
\bottomrule
\end{tabular}
\caption{Hyperparameters used in the numerical experiments. The only swept
quantity is the number of training sequences $P$; all other settings are
kept fixed across runs.}
\label{tab:numerical_hyperparameters}
\end{table}

\subsection{Training Trajectories and Filtering}
\label{appendix:traj_filtering}
\begin{figure}[h]
    \centering
    \includegraphics[width=0.45\linewidth]{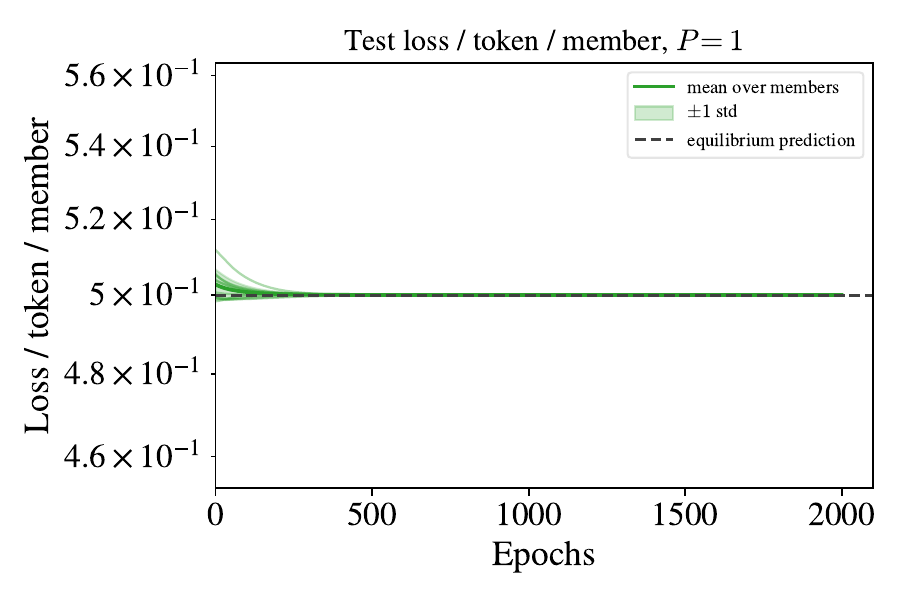}
    \includegraphics[width=0.45\linewidth]{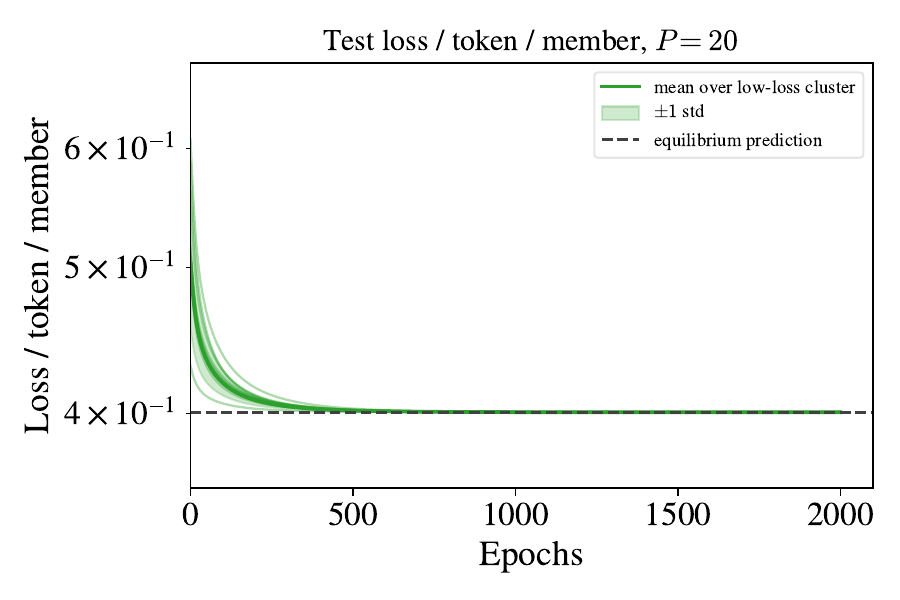}
    \includegraphics[width=0.45\linewidth]{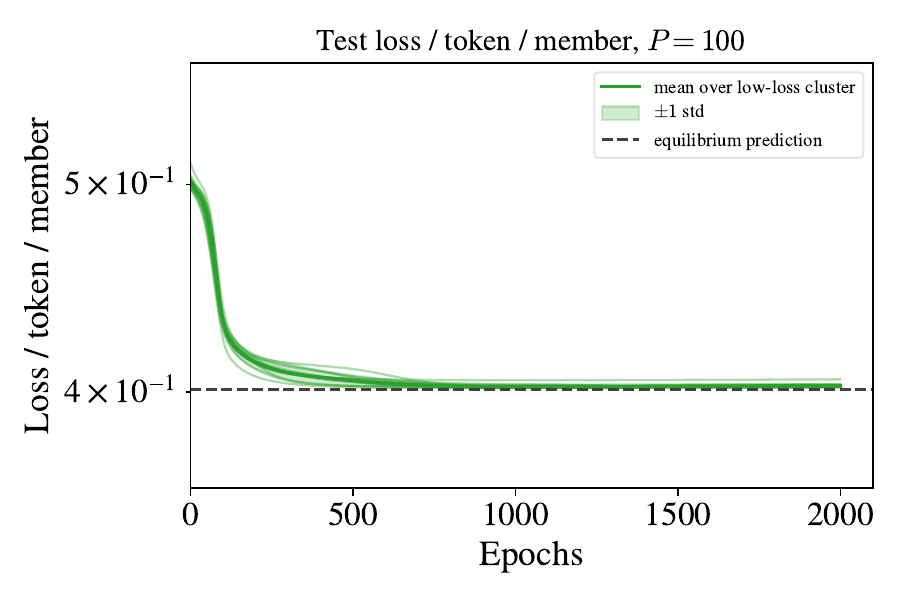}
    \includegraphics[width=0.45\linewidth]{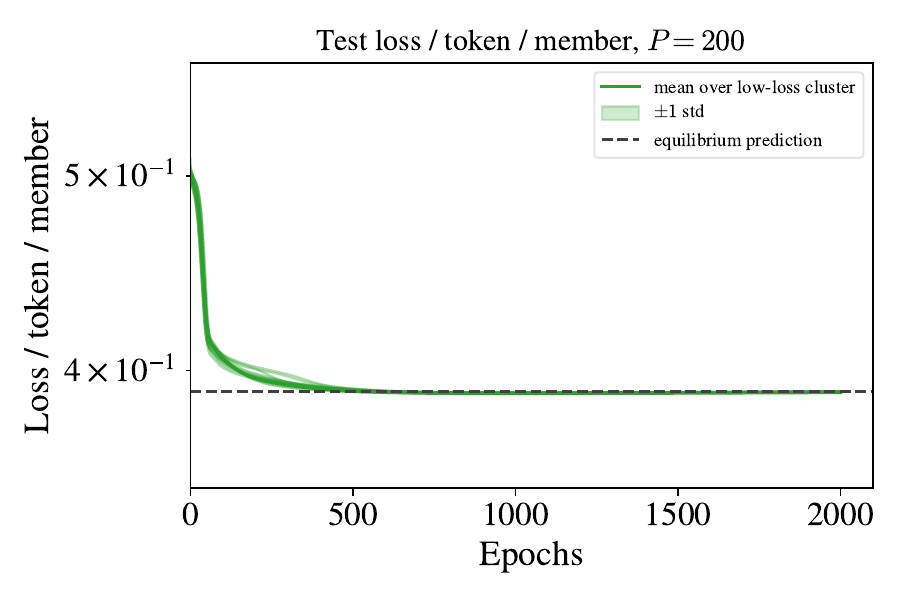}
    \includegraphics[width=0.45\linewidth]{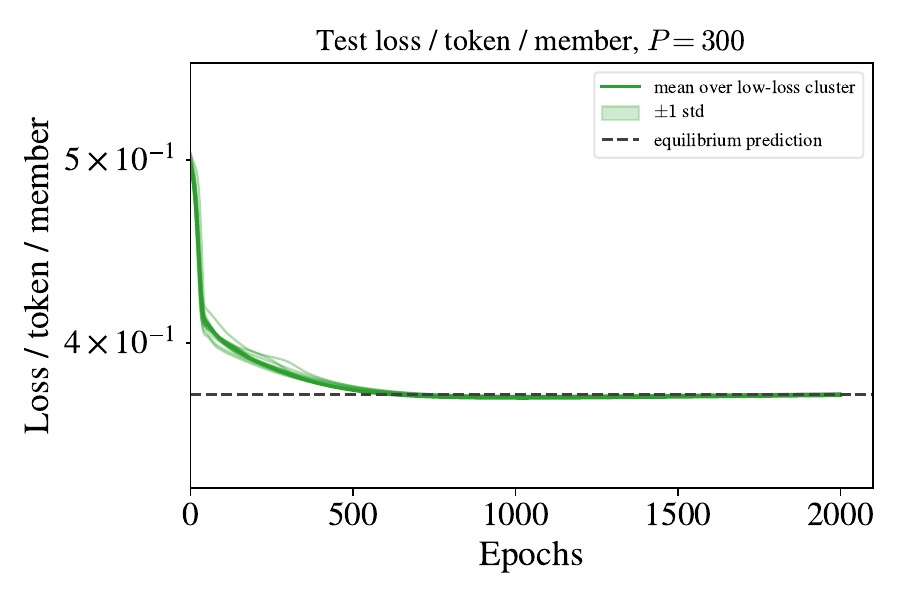}
    \includegraphics[width=0.45\linewidth]{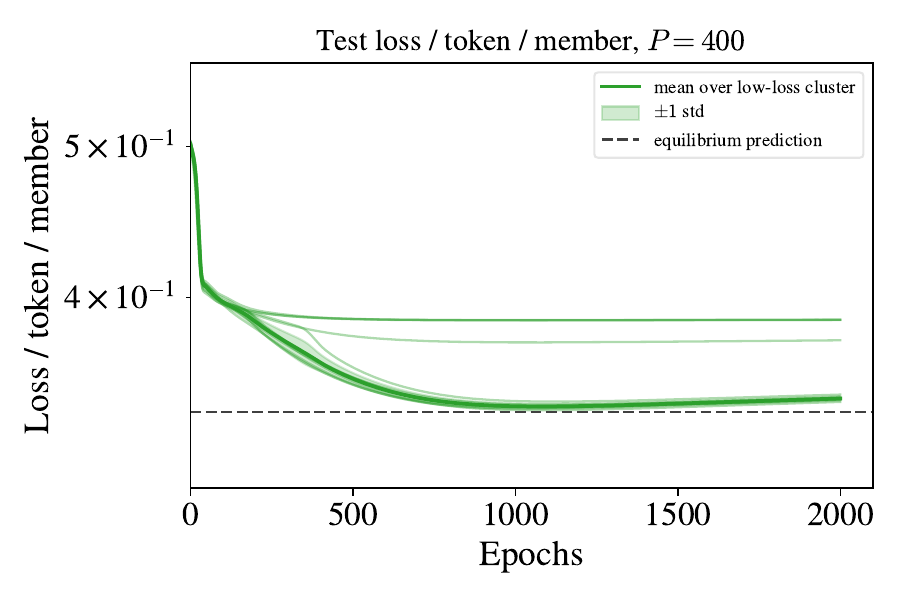}
    \includegraphics[width=0.45\linewidth]{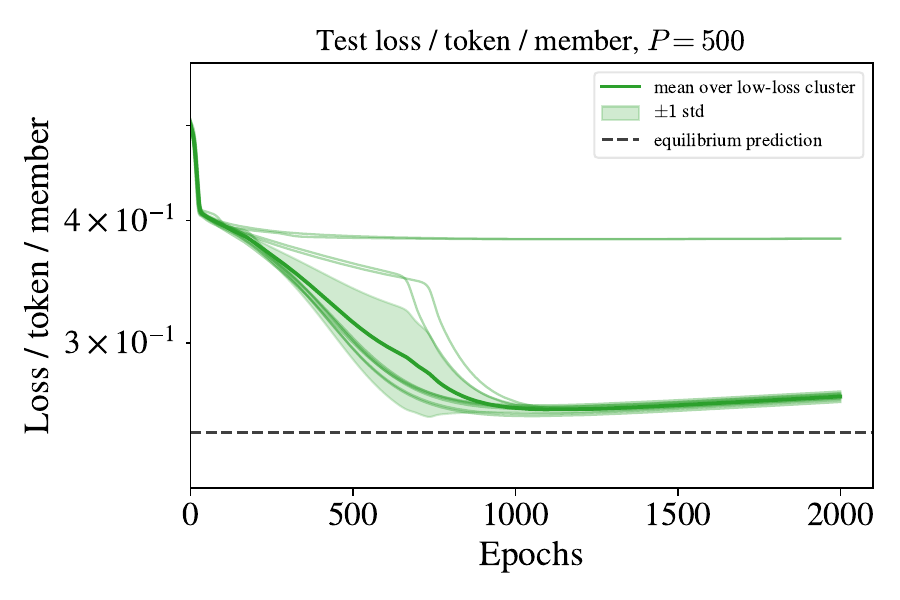}
    \includegraphics[width=0.45\linewidth]{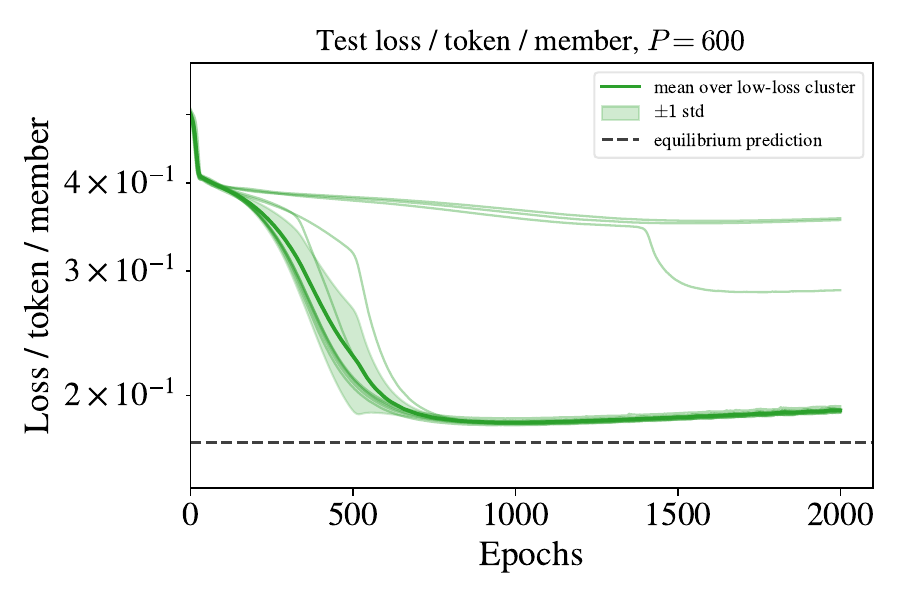}
    \includegraphics[width=0.45\linewidth]{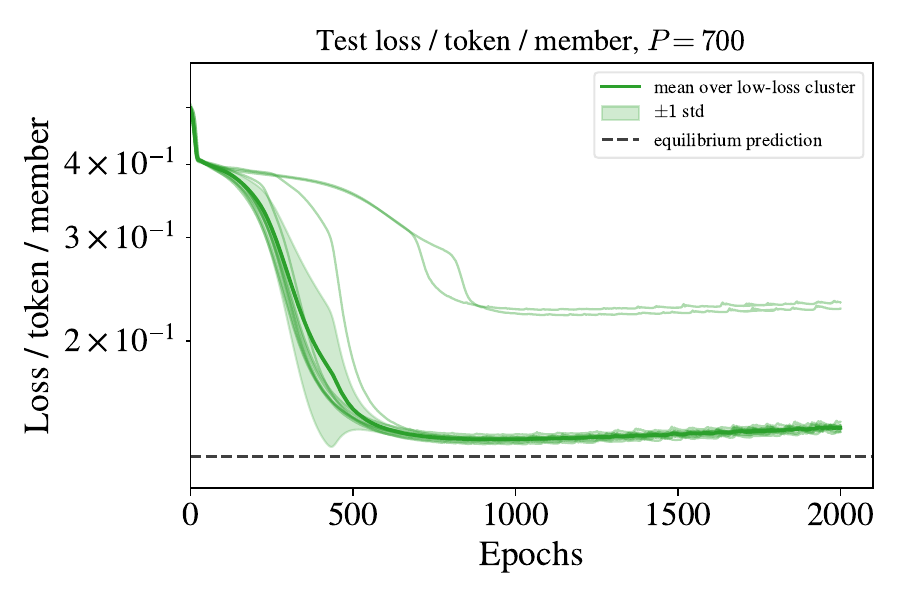}
    \includegraphics[width=0.45\linewidth]{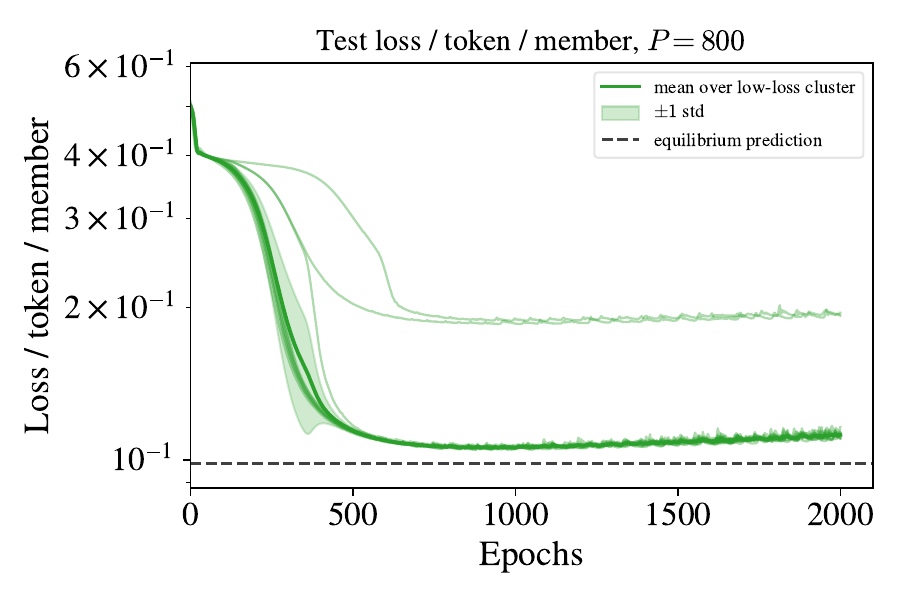}
    \caption{Test loss for \emph{linear attention} trained with Adam. We show the test loss as a function of the number of epochs for different values of $P$ on models trained with Adam. We clearly see two clusters with one reaching a lower loss value. Since the Bayesian posterior, and hence SGLD in principle, is insensitive to the initial conditions and would, in principle, reach the global minimum given sufficient time, we include only converged runs in the cluster with lower loss in the results in the main text.}
    \label{fig:loss_trajs_linear}
\end{figure}

\begin{figure}[h]
    \centering
    \includegraphics[width=0.30\linewidth]{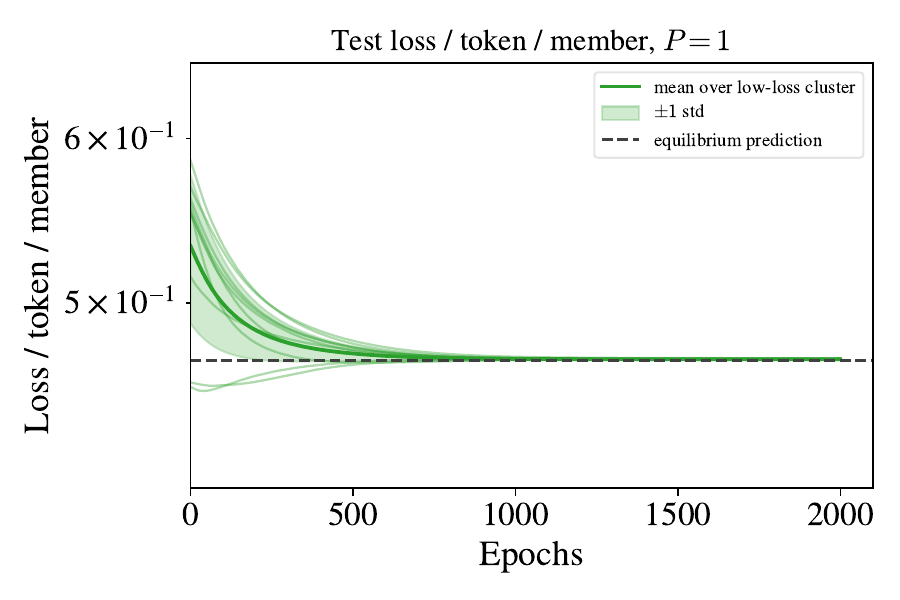}    \includegraphics[width=0.30\linewidth]{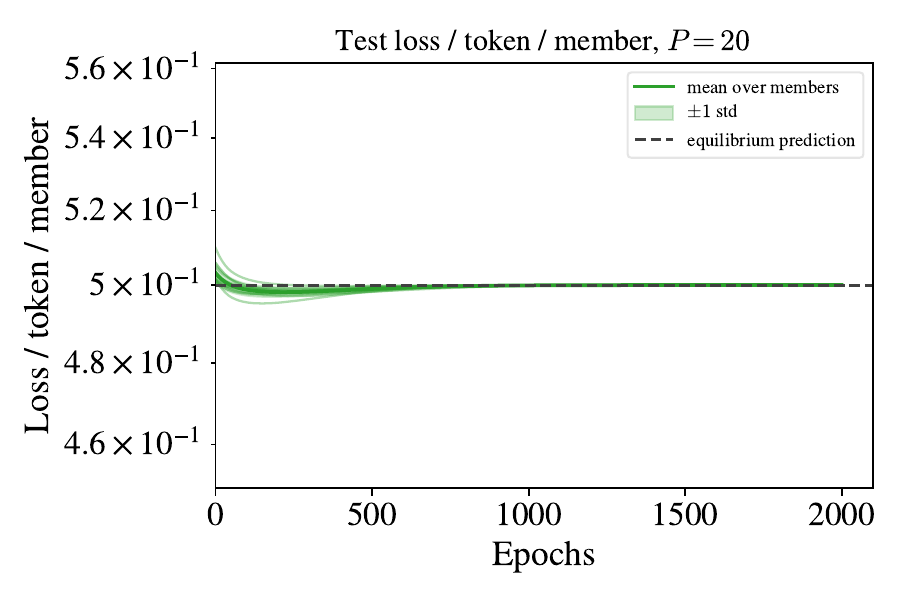}
    \includegraphics[width=0.30\linewidth]{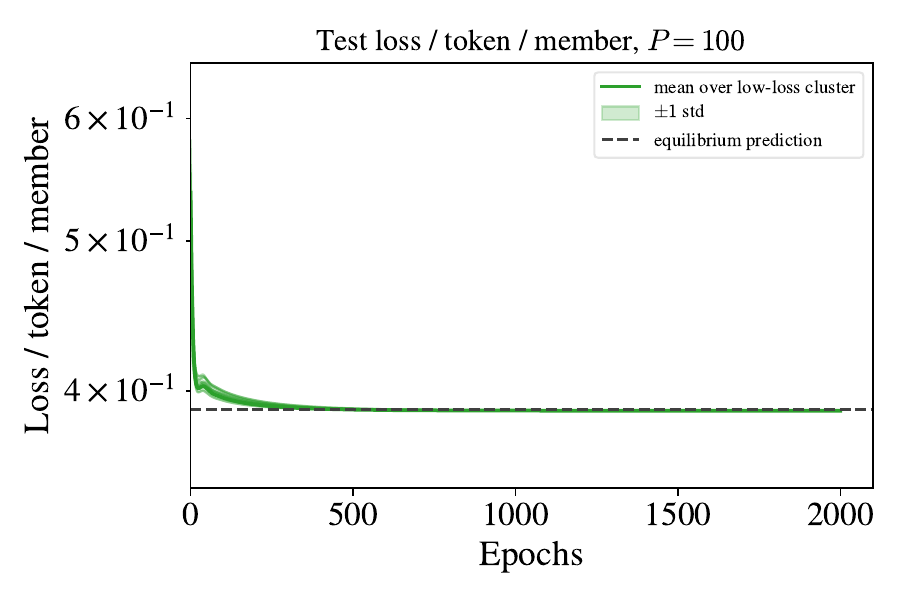}
    \includegraphics[width=0.30\linewidth]{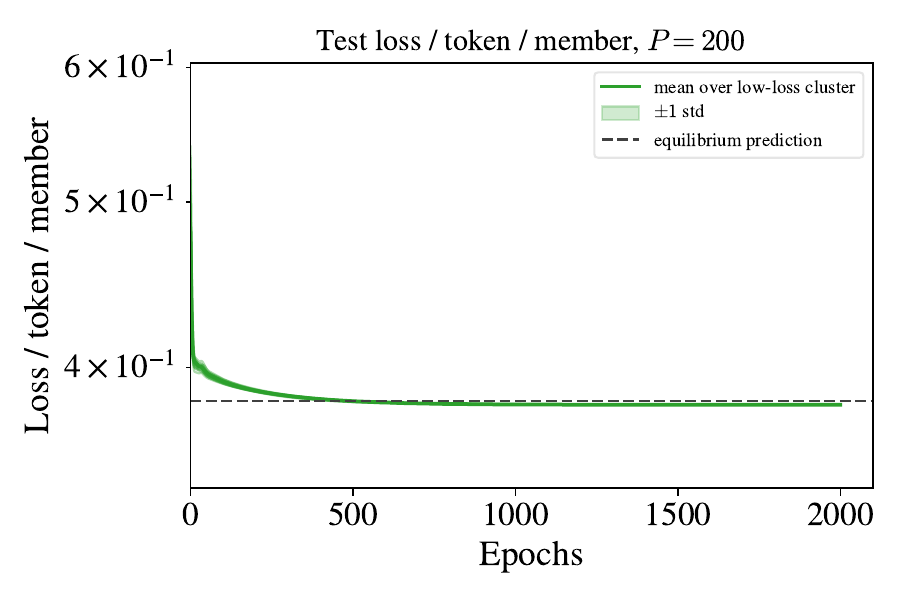}
    \includegraphics[width=0.30\linewidth]{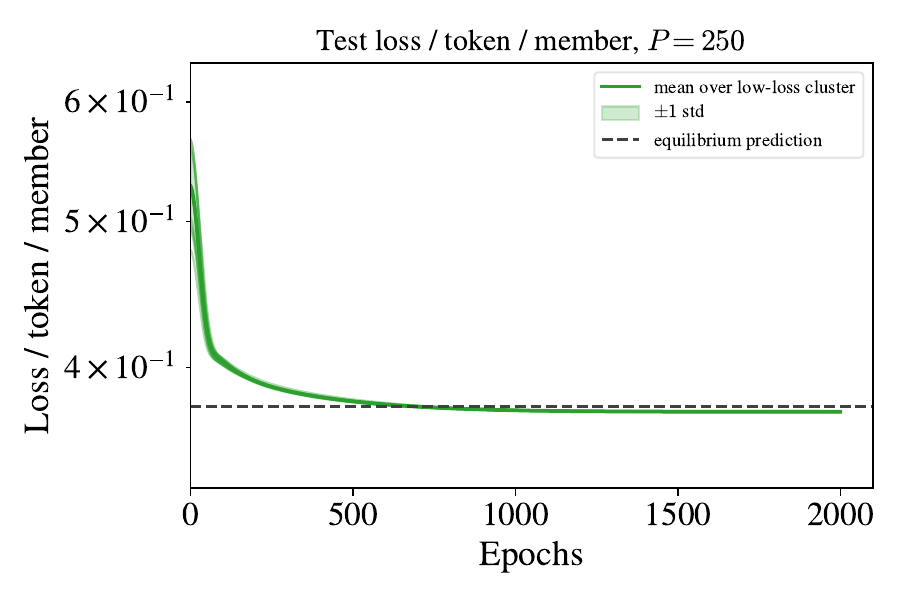}
    \includegraphics[width=0.30\linewidth]{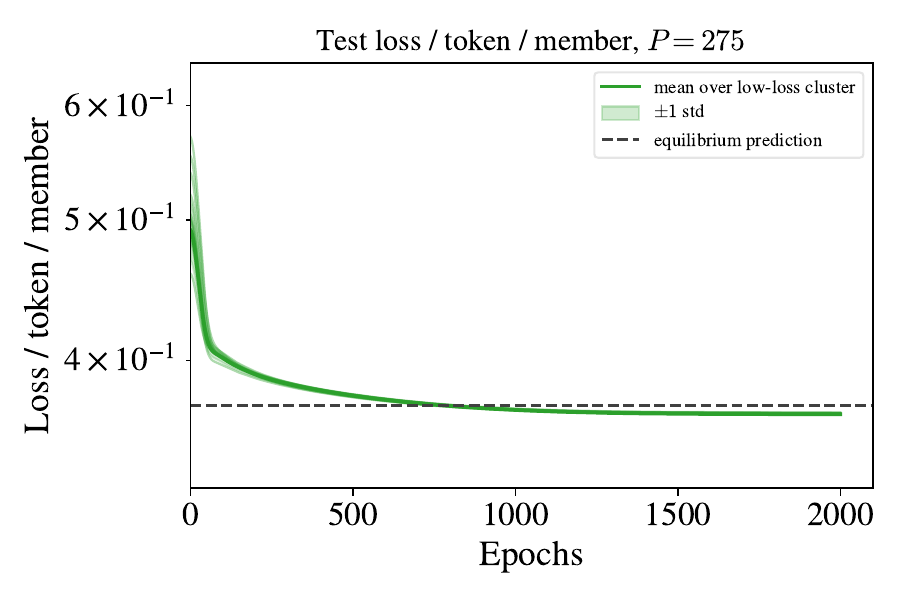}
    \includegraphics[width=0.30\linewidth]{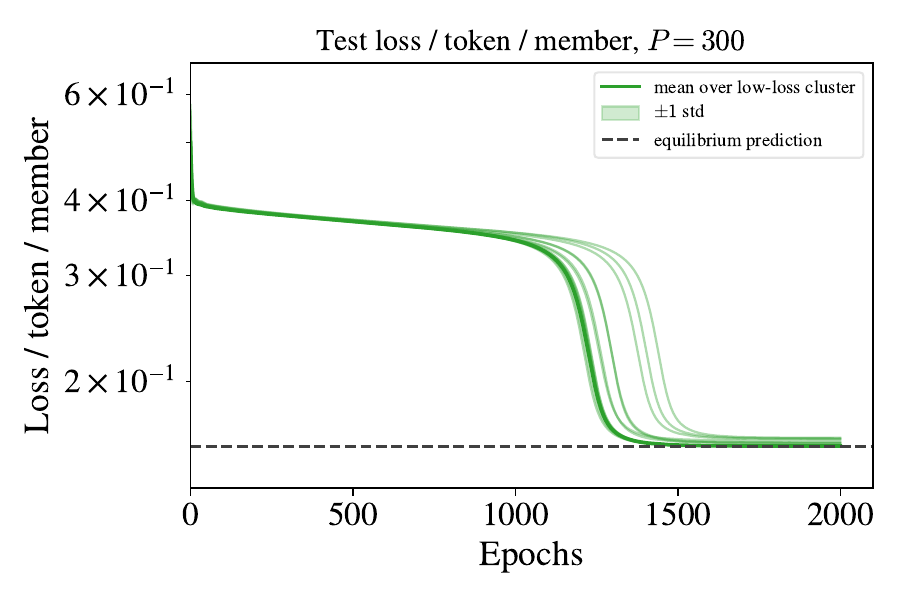}
    \includegraphics[width=0.30\linewidth]{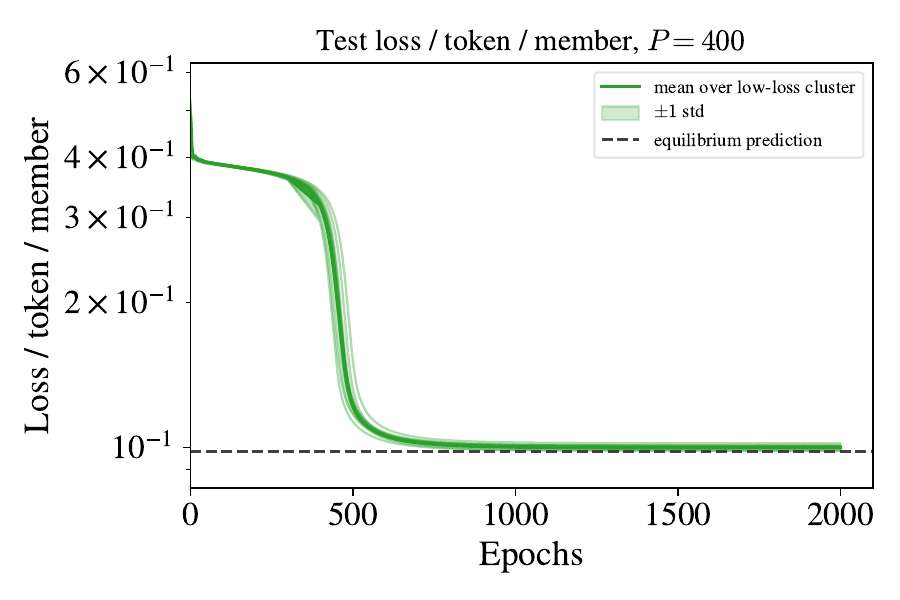}
    \includegraphics[width=0.30\linewidth]{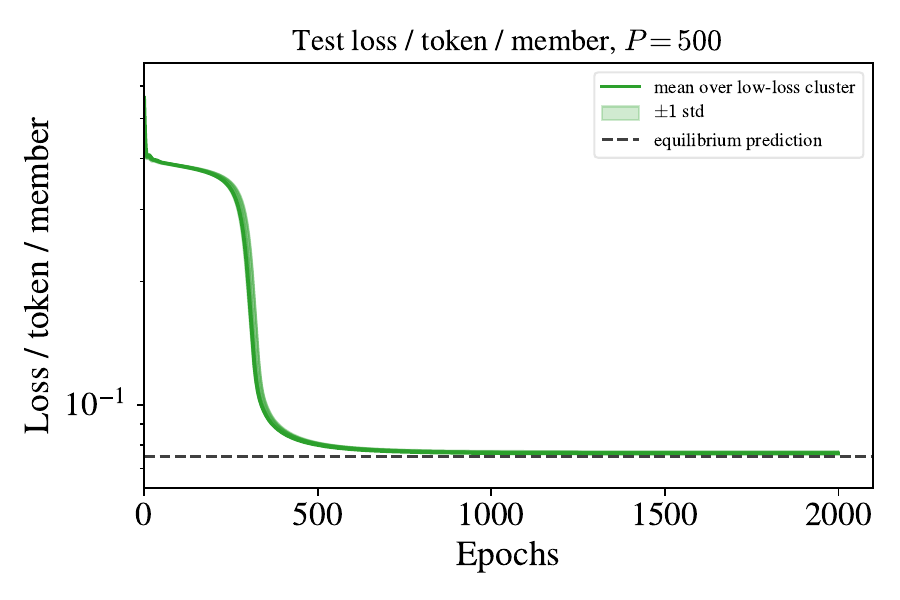}
    \includegraphics[width=0.30\linewidth]{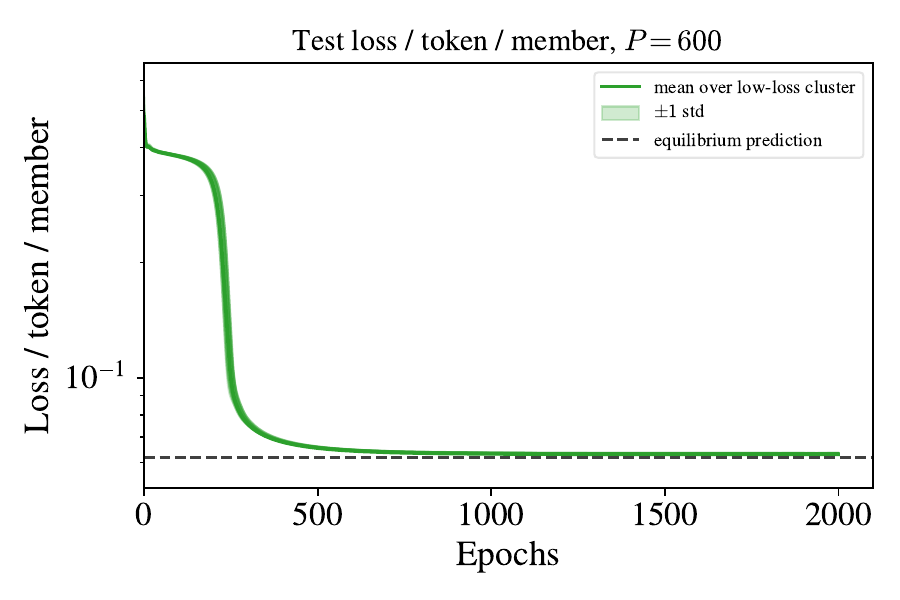}
    \includegraphics[width=0.30\linewidth]{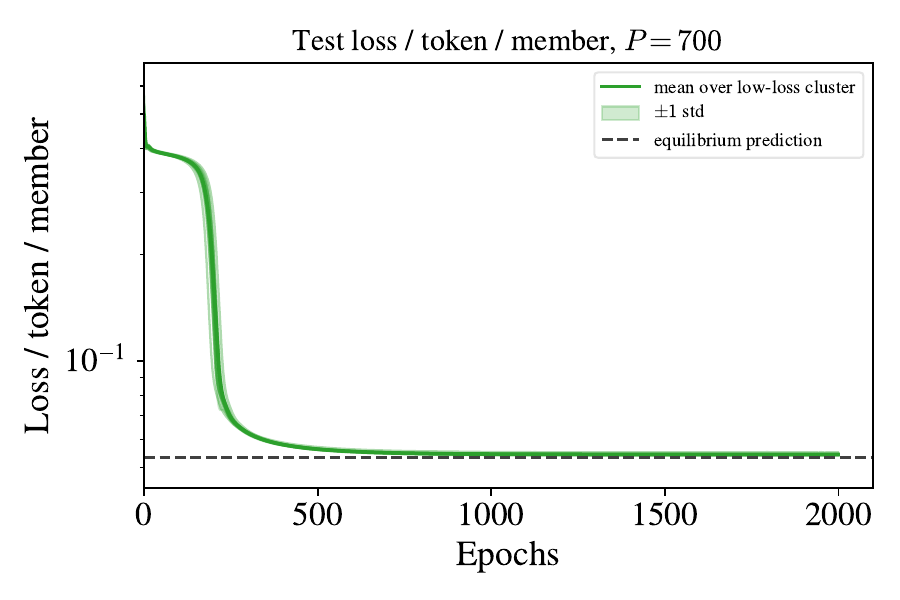}
    \includegraphics[width=0.30\linewidth]{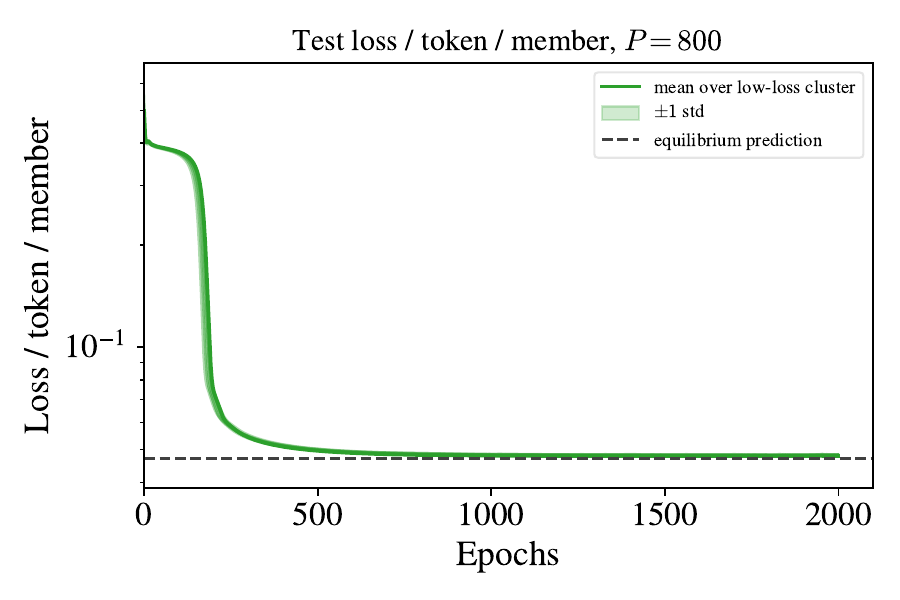}
    \caption{Test loss for \emph{softmax attention} trained with Adam. We show the test loss as a function of the number of epochs for different values of $P$ on models trained with Adam. With softmax Adam converged to the global minimum in all of our seeds.}
    \label{fig:loss_trajs_softmax}
\end{figure}

To expedite the posterior sampling, we warm start all our SGLD~\cite{wellingBayesianLearningStochastic2011} runs from the Adam early stopping minimum. We note again that the equilibrium distribution sampled by SGLD is independent of the initial conditions. Since optimizing with Adam is a more common practice, we also show the results at the end of the Adam training stage; in these cases, we filter out training runs that reached a local minimum (which is not the global minimum). 
Importantly, filtering was performed only on the basis of convergence to the global loss minimum, and not on the resulting order parameters or their agreement with theory. 
Figs.~\ref{fig:loss_trajs_linear} and ~\ref{fig:loss_trajs_softmax} show the trajectories of all runs using Adam as a function of the epoch (equiv. step, batch). For each setting, we trained an ensemble of $10$ networks.


\clearpage

\end{document}